\documentclass[sigconf]{acmart}

\usepackage{graphicx}
\usepackage{tabularx}
\usepackage{color}
\usepackage{subcaption}
\usepackage{multirow}
\usepackage{balance}
\usepackage{xcolor}
\usepackage{hyperref}

\hypersetup{
  colorlinks,
  citecolor=violet,
  linkcolor=red,
  urlcolor=blue}

\AtBeginDocument{%
  \providecommand\BibTeX{{%
    \normalfont B\kern-0.5em{\scshape i\kern-0.25em b}\kern-0.8em\TeX}}}

\copyrightyear{2022}
\acmYear{2022}
\setcopyright{rightsretained}
\acmConference[MM '22]{Proceedings of the 30th ACM International Conference on Multimedia}{October 10--14, 2022}{Lisboa, Portugal}
\acmBooktitle{Proceedings of the 30th ACM International Conference on Multimedia (MM '22), October 10--14, 2022, Lisboa, Portugal}\acmDOI{10.1145/3503161.3547838}
\acmISBN{978-1-4503-9203-7/22/10}

\acmSubmissionID{394}

\begin{document}

\title{MegaPortraits: One-shot Megapixel Neural Head Avatars} 


	



\author{Nikita Drobyshev}
\affiliation{ 
  \institution{Samsung AI Center - Moscow}
  \country{Russia} 
}

\author{Jenya Chelishev} 
\affiliation{ 
  \institution{Samsung AI Center - Moscow}
  \country{Russia} 
}

\author{Taras Khakhulin} 
\affiliation{ 
  \institution{Samsung AI Center - Moscow \and Skolkovo Institute of Science and Technology}
  \country{Russia} 
}

\author{Aleksei Ivakhnenko}
\affiliation{ 
  \institution{Samsung AI Center - Moscow}
  \country{Russia} 
}

\author{Victor Lempitsky} 
\affiliation{ 
  \institution{Yandex}
  \country{Armenia} 
}

\author{Egor Zakharov} 
\affiliation{ 
  \institution{Samsung AI Center - Moscow \and Skolkovo Institute of Science and Technology}
  \country{Russia} 
}




\renewcommand{\shortauthors}{Nikita Drobyshev et al.}
\begin{abstract}
In this work, we advance the neural head avatar technology to the megapixel resolution while focusing on the particularly challenging task of \emph{cross-driving} synthesis, i.e., when the appearance of the \emph{driving} image is substantially different from the animated \emph{source} image. We propose a set of new neural architectures and training methods that can leverage both medium-resolution video data and high-resolution image data to achieve the desired levels of rendered image quality and generalization to novel views and motion. We demonstrate that suggested architectures and methods produce convincing high-resolution neural avatars, outperforming the competitors in the cross-driving scenario. Lastly, we show how a trained high-resolution neural avatar model can be distilled into a lightweight student model which runs in real-time and locks the identities of neural avatars to several dozens of pre-defined source images. Real-time operation and identity lock are essential for many practical applications head avatar systems.
\href{https://samsunglabs.github.io/MegaPortraits/}{\color{blue}MegaPortraits website}
\vspace{-0.4cm}
\end{abstract}

\ccsdesc[500]{Computing methodologies~Image-based rendering}
\ccsdesc[300]{Computing methodologies~Neural networks}

\keywords{Neural rendering, generative models, one-shot neural avatars}

\newlength{\wid}
\newlength{\mrg}
\newlength{\mrgv}

\newcommand{\fig}[1]{Figure~\ref{fig:#1}}
\newcommand{\sect}[1]{Section~\ref{sect:#1}}
\newcommand{\tab}[1]{Table~\ref{tab:#1}}
\newcommand{\alg}[1]{Algorithm~\ref{alg:#1}}
\newcommand{\eq}[1]{(\ref{eq:#1})}

\begin{teaserfigure}
    \centering    
    \setlength{\wid}{0.19\textwidth}
    \setlength{\mrg}{-0.3cm}
    \begin{tabular}{ccccc}
        \includegraphics[width=\wid]{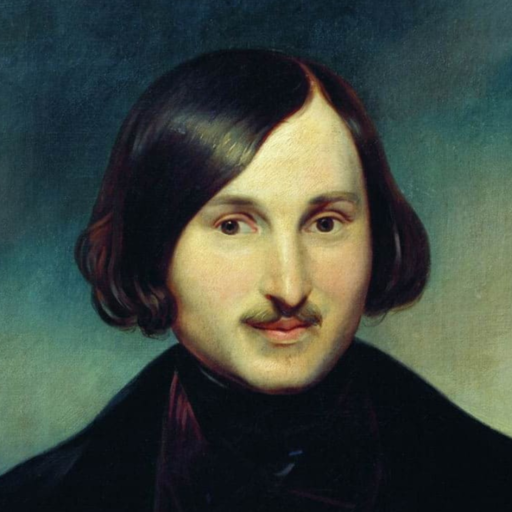} & 
        \hspace{\mrg}
        \includegraphics[width=\wid]{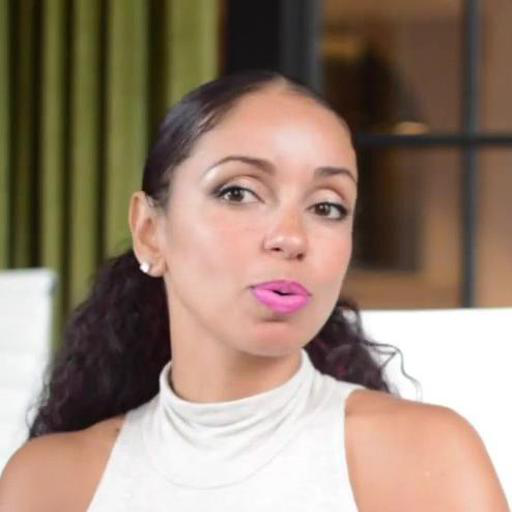} & 
        \hspace{\mrg}
        \includegraphics[width=\wid]{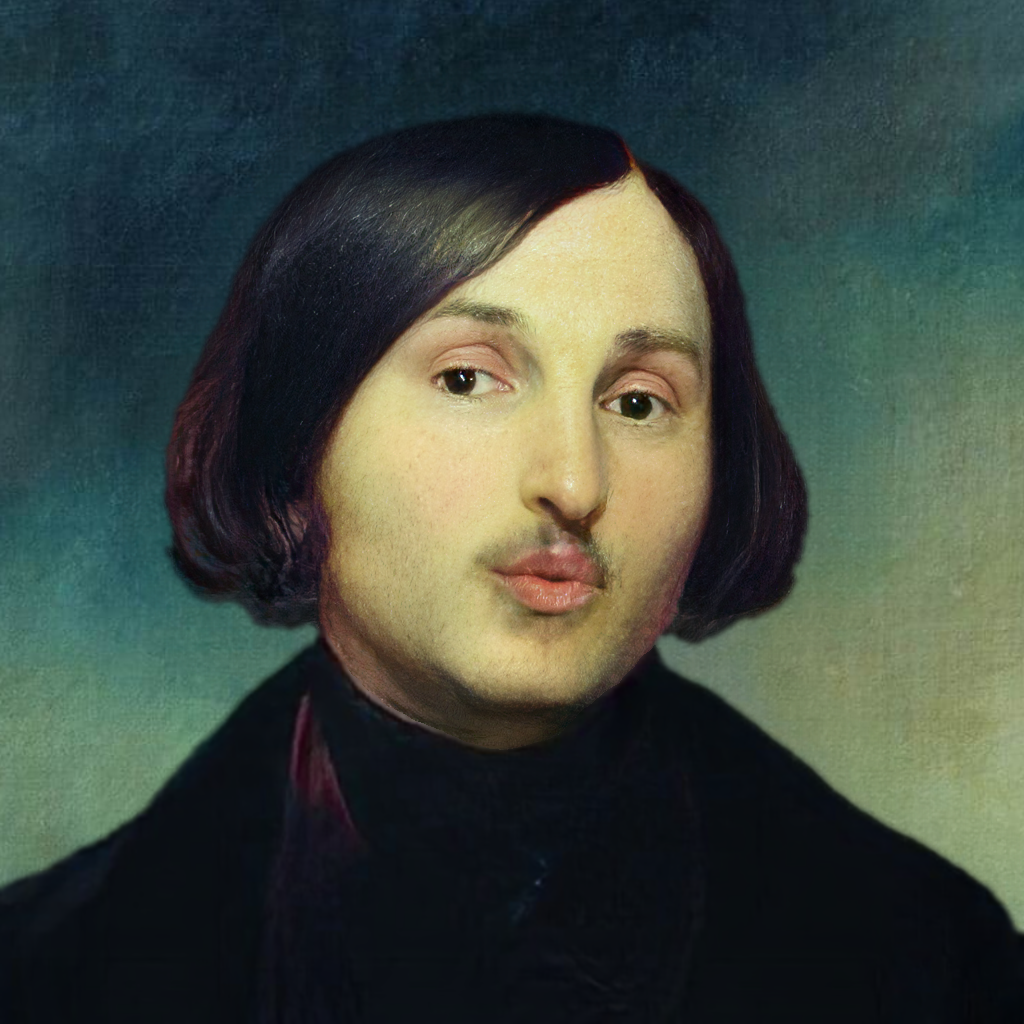} & 
        \hspace{\mrg}
        \includegraphics[width=\wid]{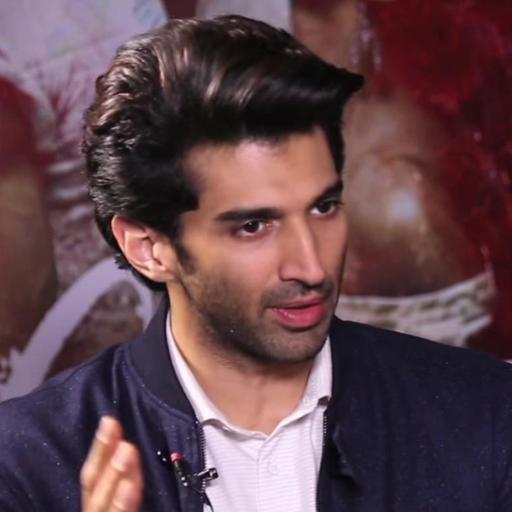} & 
        \hspace{\mrg}
        \includegraphics[width=\wid]{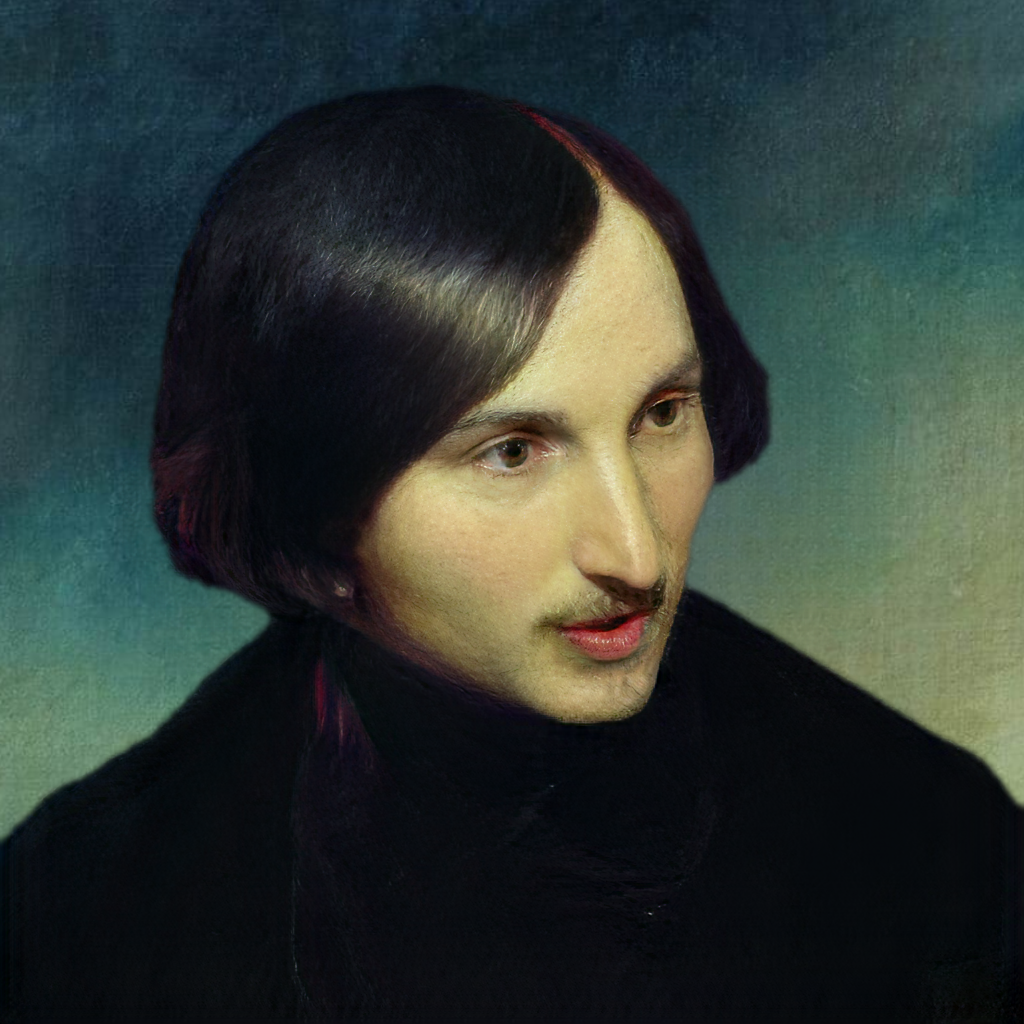}
        \\ %
        \includegraphics[width=\wid]{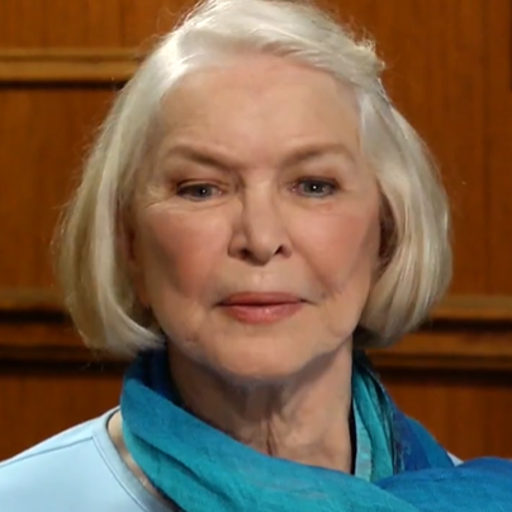} & 
        \hspace{\mrg}
        \includegraphics[width=\wid]{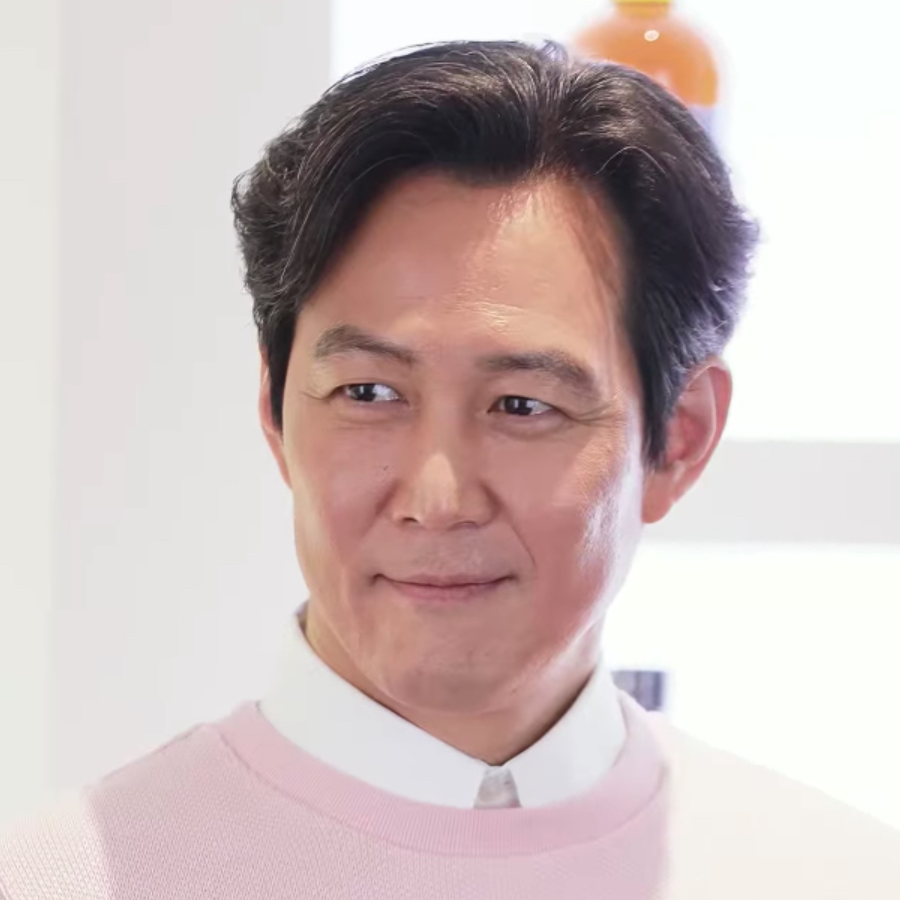} & 
        \hspace{\mrg}
        \includegraphics[width=\wid]{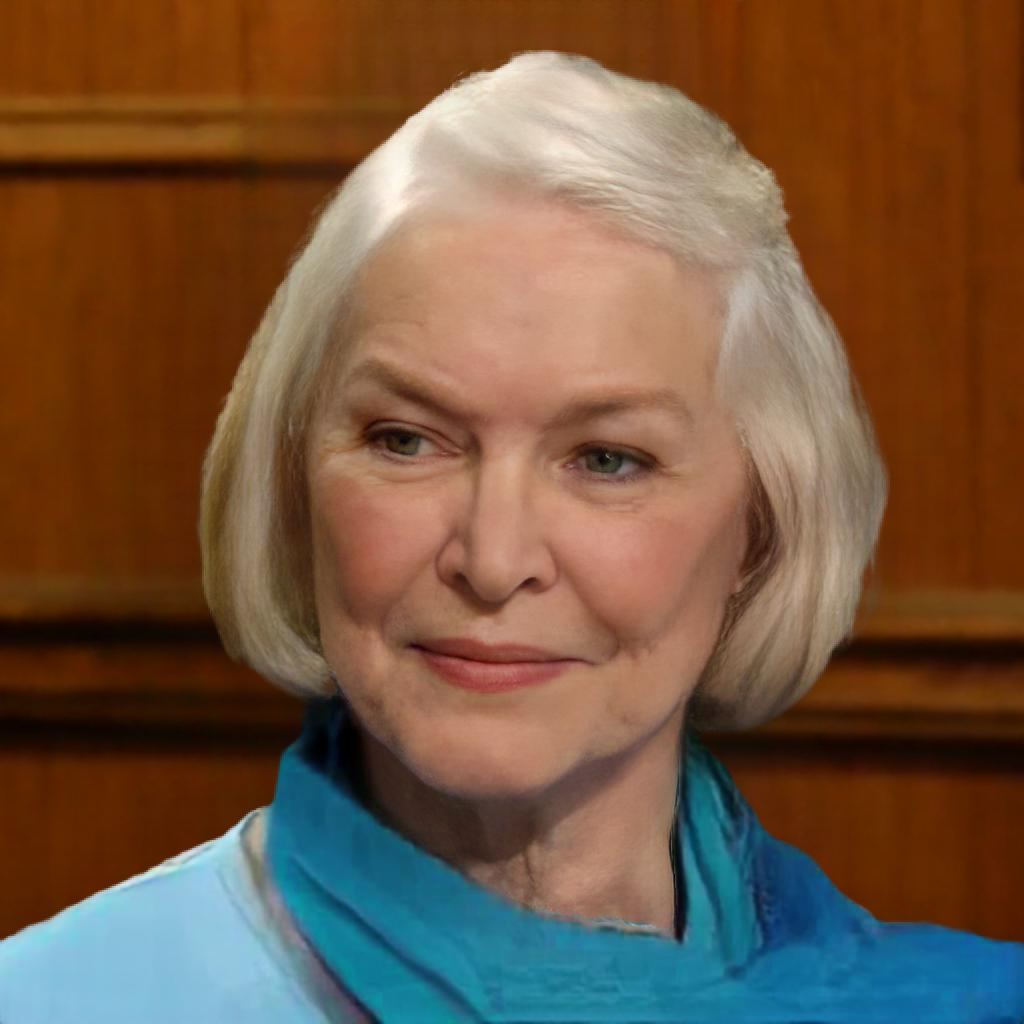} & 
        \hspace{\mrg}
        \includegraphics[width=\wid]{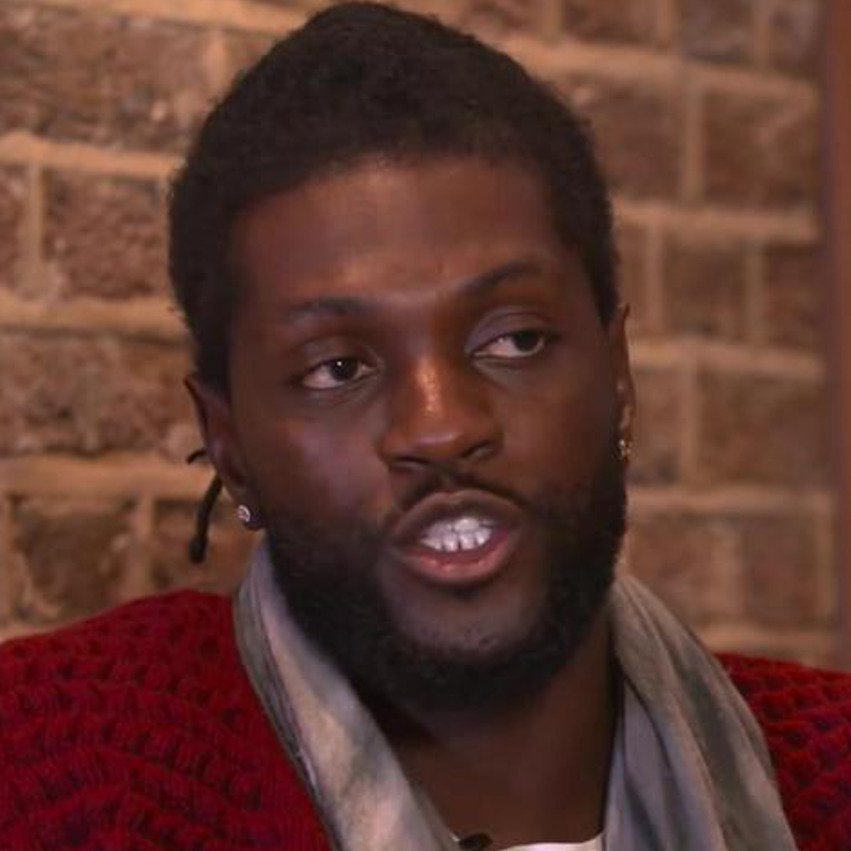} & 
        \hspace{\mrg}
        \includegraphics[width=\wid]{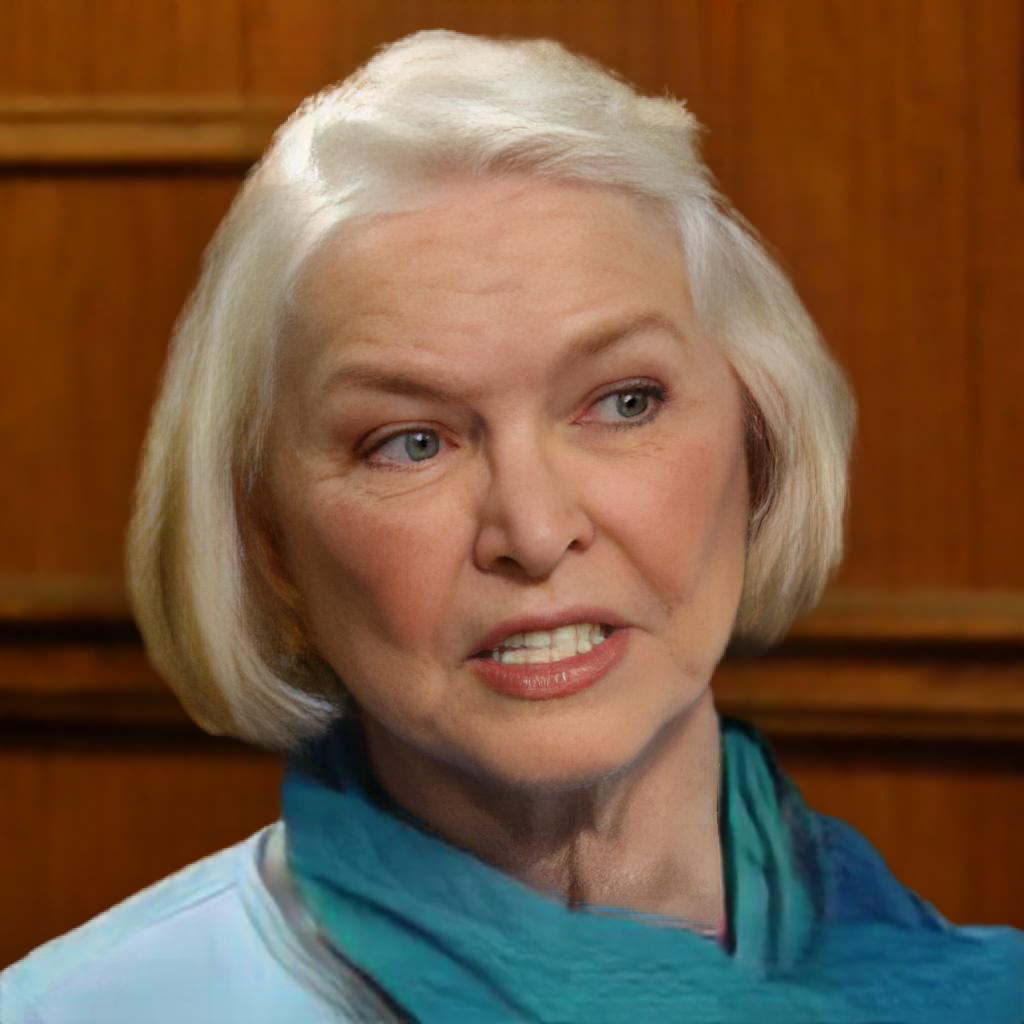}
        \\
        \textbf{Source} & 
        \hspace{\mrg} 
        \textbf{Driver} & 
        \hspace{\mrg} 
        \textbf{Ours} & 
        \hspace{\mrg} 
        \textbf{Driver} & 
        \hspace{\mrg} 
        \textbf{Ours}
    \end{tabular}
    \vspace{-0.4cm}
    \caption{We present the first system capable of creating megapixel avatars from single portrait images. Our method outperforms its competitors in the quality of the cross-driving results and manages to preserve the high-resolution appearance of the source image even for out-of-domain examples like paintings, as seen in this example.}
    \label{fig:teaser}
\end{teaserfigure}

\def\X{\mathbf{X}}
\def\x{\mathbf{x}}
\def\w{\mathbf{w}}
\def\v{\mathbf{v}}
\def\e{\mathbf{e}}
\def\E{\mathbf{E}}
\def\R{\mathbf{R}}
\def\z{\mathbf{z}}
\def\t{\mathbf{t}}
\def\G{\mathbf{G}}
\def\W{\mathbf{W}}


\maketitle

\section{Introduction}

Neural head avatars~\cite{Thies2019Face2FaceRF, Kim2018DeepVP, Siarohin2019AnimatingAO, Siarohin2019FirstOM, Zakharov2019FewShotAL, Burkov_2020_CVPR, Wang2021OneShotFN, Ha2020MarioNETteFF, Zakharov2020FastBN, Lombardi2018DeepAM, Lombardi2019NeuralV, Park2021NerfiesDN, Park2021HyperNeRFAH, Doukas2021HeadGANON} offer a new fascinating way of creating virtual head models. They bypass the complexity of realistic physics-based modeling of human avatars by learning the shape and appearance directly from the videos of talking people. Over the last several years, methods that can create realistic avatars from a single photograph (one-shot) have been developed~\cite{Wang2021OneShotFN, Zakharov2020FastBN, Siarohin2019FirstOM, Doukas2021HeadGANON}. They leverage extensive pre-training on the large datasets of videos of different people~\cite{Chung2018VoxCeleb2DS, Wang2021OneShotFN} to create the avatars in the one-shot mode using generic knowledge about human appearance.

Despite the impressive results obtained by this class of methods, their quality is severely limited by the resolution of the training datasets. This limitation cannot be easily bypassed by collecting a higher resolution dataset since it needs to be simultaneously large-scale and diverse, i.e., include thousands of humans with multiple frames per person, diverse demographics, lighting, background, face expression, and head pose. To the best of our knowledge, all public datasets~\cite{Chung2018VoxCeleb2DS, Wang2021OneShotFN} that meet these criteria are limited in resolution. As a result, even the most recent one-shot avatar systems~\cite{Wang2021OneShotFN} learn the avatars at resolutions up to $512\times{}512$.

In our work, we make three main contributions. First, we propose a new model for one-shot neural avatars that achieves state-of-the-art cross-reenactment quality in up to $512 \times 512$ resolution. In our architecture, we utilize the idea of representing the appearance of the avatars as a latent 3D volume~\cite{Wang2021OneShotFN} and propose a new way to combine it with the latent motion representations~\cite{Burkov_2020_CVPR}, which includes a novel contrastive loss that allows our system to achieve higher degrees of disentanglement between the latent motion and appearance representations. On top of that, we add a problem-specific \textit{gaze} loss that increases the realism and accuracy of eye animation.

Our second and crucial contribution is showing how a model trained on medium-resolution videos can be ``upgraded'' to the megapixel ($1024 \times 1024$) resolution using an additional dataset of high-resolution still images. As a result, our proposed method, while using the same training dataset, outperforms the baseline super-resolution approach~\cite{Yang2020HiFaceGANFR} for the task of cross-reenactment. We are thus the first to demonstrate neural head avatars in proper megapixel resolution.

Lastly, since many practical applications for human avatar creation require real-time or faster than real-time rendering, we distill our megapixel model into ten times faster student model that runs at 130 FPS on a modern GPU. This significant speedup is possible since the student is trained for specific appearances (unlike the main model that can create new avatars for previously unseen people). Furthermore, the applications based on such a student model ``locked'' to predefined identities can prevent its misuse for creating ``deep fakes'' while at the same time achieving low rendering latency.
\section{Related work}


The recent success of neural implicit scene representations~\cite{Mildenhall2020NeRFRS} for the problem of 3D reconstruction has inspired several works on the so-called 4D head avatars~\cite{Park2021NerfiesDN, Gafni2021DynamicNR, Park2021HyperNeRFAH, Yang2021BANMoBA, Lombardi2018DeepAM, Lombardi2019NeuralV}, which treat the problem of appearance and motion modeling of the avatars as a non-rigid reconstruction of the training video. These methods have different ways of handling the non-rigidity of motion and either learn it from scratch~\cite{Yang2021BANMoBA, Park2021NerfiesDN, Park2021HyperNeRFAH}, use pre-trained motion extractors~\cite{Gafni2021DynamicNR} or pre-computed coarse meshes~\cite{Lombardi2018DeepAM, Lombardi2019NeuralV}. While all these methods can achieve an impressive realism of renders and fidelity of motions, they require multi-shot training data, are trained separately for each avatar, and often fail to represent motions unseen during training. In contrast, our method can impose motion from an arbitrary video sequence on an appearance obtained from a single image while still achieving megapixel resolutions of the renders.

Direct generation of videos via convolutional neural networks, conditioned on appearance and motion descriptors, is an alternative approach to talking-head synthesis. While the early works in this area learned an avatar from the video~\cite{Thies2019Face2FaceRF, Kim2018DeepVP}, the follow-up works added few-shot and one-shot capabilities~\cite{Zakharov2019FewShotAL, Burkov_2020_CVPR, Zakharov2020FastBN, Siarohin2019FirstOM, Wang2021OneShotFN, Doukas2021HeadGANON, Siarohin2019AnimatingAO}. Most of these works use explicit representations for the motion, such as keypoints or blendshapes, while others~\cite{Burkov_2020_CVPR} have adopted latent motion parameterization. The latter achieves better expressiveness of motion if the disentanglement from the appearance is achieved during training. In our system, we chose the latter approach and proposed a new method of disentangling the motion and the appearance descriptors, which significantly improves the quality of the results.

The resolution of the talking head models is currently upper bounded by the available video datasets~\cite{Chung2018VoxCeleb2DS, Wang2021OneShotFN}, which contain videos of at most $512 \times 512$ resolution. This problem further restricts the enhancement of the output quality on the existing datasets using the standard high-quality image and video synthesis techniques~\cite{wang2018pix2pixHD, wang2018vid2vid}. Alternatively, this problem could be treated as single image super-resolution (SISR). This way, we require only the dataset of still high-resolution images for training, which is easier to obtain. However, the quality of the outputs of the one-shot talking head model varies greatly depending on the imposed motion, which results in poor performance of standard SISR methods~\cite{Yang2020HiFaceGANFR}. These classic approaches rely on supervised training procedures with an a priori known ground truth, which we cannot provide for the novel motion data since we only have one image per person. We address this problem in a novel way by combining supervised and unsupervised training and achieve considerably better performance for arbitrary motion data than the solution based on SISR.

\begin{figure*}[!ht]
    \centering
    \includegraphics[width=1\textwidth]{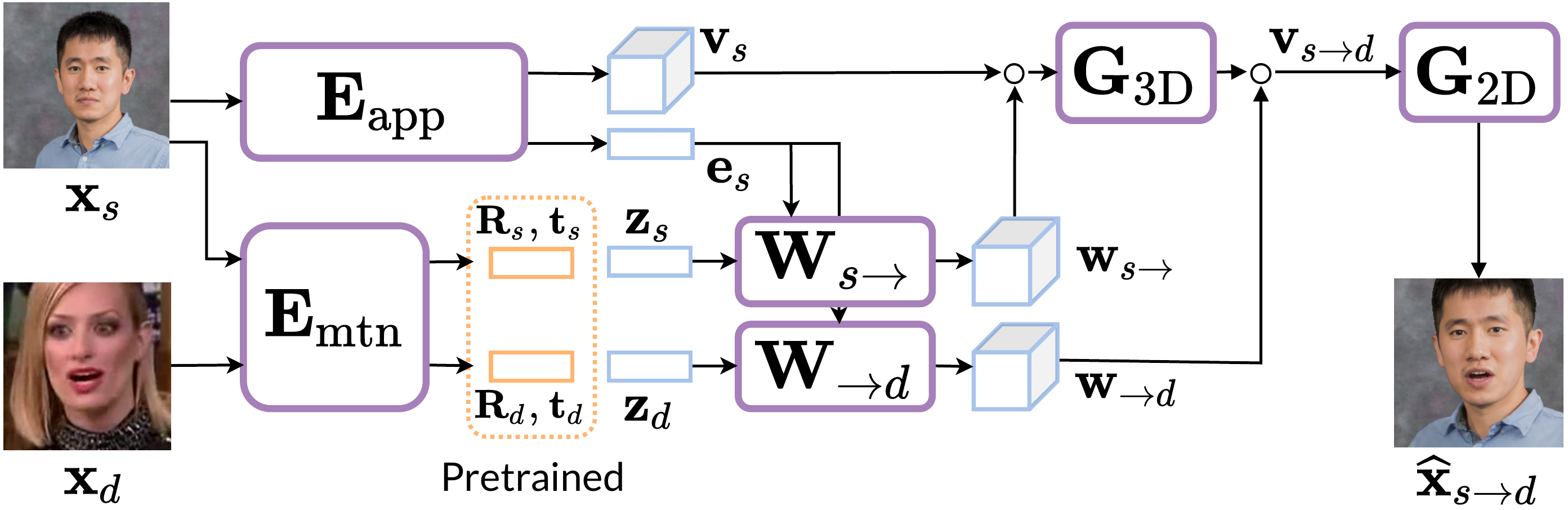}
    \vspace{-0.4cm}
    \caption{Overview of our base model. To encode the appearance of the source frame, we predict volumetric features $\v_s$ and a global descriptor $\e_s$ from the source image via an appearance encoder $\E_\text{app}$. In parallel, we predict the motion representations from both the source and driving images using a motion encoder $\E_\text{mtn}$. These representations consist of the explicit head rotations $\R_{s/d}$, translations $\t_{s/d}$, and the latent expression descriptors $\z_{s/d}$. They are used to predict the 3D warpings $\w_{s \rightarrow}$ and $\w_{\rightarrow d}$ via the separate warping generators $\W_{s \rightarrow}$ and $\W_{\rightarrow d}$. The first warping removes the source motion from the appearance features $\v_s$ by mapping them into a canonical coordinate space, and the second one imposes the driver motion. The canonical volume is processed by a 3D convolutional network $\G_\text{3D}$, and the driving volume $\v_{s \rightarrow d}$ is orthographically projected into 2D features and processed by a 2D convolutional network $\G_\text{2D}$, which predicts an output image $\hat\x_{s \rightarrow d}$.}
    \label{fig:scheme_stg1}
\end{figure*}

\section{Method}
\label{sec:method}

We propose a system for the one-shot creation of high-resolution human avatars, called \textit{megapixel portraits} or MegaPortraits for short. Our model is trained in two stages. 
Optionally, we propose an additional distillation stage for faster inference. Our training setup is relatively standard. We sample two random frames from our dataset at each step: the source frame $\x_s$ and the driver frame $\x_d$. Our model imposes the motion of the driving frame (i.e., the head pose and the facial expression) onto the appearance of the source frame to produce an image $\hat\x_{s \rightarrow d}$. The main learning signal is obtained from the training episodes where the source and the driver frames come from the same video, and hence our model's prediction is trained to match the driver frame. In this section, we will focus on the principal training regime while leaving details of the architectures to the supplementary materials. 

\subsection{Base model} 

During the first stage, we train our base model (\fig{scheme_stg1}) by sampling two frames $\x_s$ and $\x_d$ from a random training video. The driving frame acts as both an input for our system and the ground truth. The source frame $\x_s$ is passed through an \textit{appearance encoder} $\E_\text{app}$, which outputs local volumetric features $\v_s$ (a 4D tensor with the fourth dimension corresponding to channels), and the global descriptor $\e_s$. In parallel, the motion descriptors of the source and driver images are calculated by separately applying a \textit{motion encoder} $\E_\text{mtn}$ to each image. This encoder outputs head rotations $\R_{s/d}$, translations $\t_{s/d}$, and latent expression descriptors $\z_{s/d}$. 
The source tuple $(\R_s, \t_s, \z_s, \e_s)$ is then input into a warping generator $\W_{s\rightarrow}$ to produce a 3D warping field $\w_{s\rightarrow}$, which removes the motion data from the volumetric features $\v_s$ by mapping them into a canonical coordinate space. These features are then processed by a 3D convolutional network $\G_\text{3D}$. Finally, the driver tuple $(\R_d, \t_d, \z_d, \e_s)$ is fed into a separate warping generator $\W_{\rightarrow d}$, which output $\w_{\rightarrow d}$ is used to impose the driver motion. The final 4D volumetric features are therefore obtained in the following way:
\begin{equation}
    \v_{s \rightarrow d} = \w_{\rightarrow d} \circ \G_\text{3D} (\w_{s\rightarrow} \circ \v_s),
\end{equation}
where $\circ$ represents a 3D warping operation. The idea behind this approach is first to rotate the volumetric features into a frontal viewpoint, remove any face expression motion decoded from $\z_s$, process them by a 3D convolutional network, and then impose the driver head rotation and motion. We use a pre-trained network to estimate head rotation data, but the latent expression vectors $\z_{s/d}$ and the warpings to and from the canonical coordinate space are trained without direct supervision. 

The volumetric feature encoding and the explicit use of head pose are inspired by \cite{Wang2021OneShotFN}. However, a significant difference with~\citep{Wang2021OneShotFN} is that we do not use keypoints to represent expression and instead rely on the latent descriptor~\cite{Burkov_2020_CVPR}, which is decoded into the explicit 3D warping field to represent face mimics in a more person-independent way. We have also observed that the motion disentanglement scheme proposed in~\cite{Burkov_2020_CVPR} starts to fail when we increase the capacity of the avatar system to facilitate higher resolutions. This problem leads to severe appearance leakage from the driving to the predicted image. To combat that, we propose using a cycle-consistency loss, which we describe below, and improving the driving image's pre-processing pipeline. For more details, please refer to the supplementary materials.

Finally, the driver volumetric features $\v_{s \rightarrow d}$ are orthographically projected into the camera frame using the same approach as in~\cite{Wang2021OneShotFN}. We denote this operation as $\mathcal{P}$. The resulting 2D feature map is decoded into the output image by a 2D convolutional network $\G_\text{2D}$:
\begin{equation}
    \hat\x_{s \rightarrow d} = \G_\text{2D}\big( \mathcal{P}(\v_{s \rightarrow d}) \big).
\end{equation}
We refer to the combination of the networks described above as $\G_\text{base}$, so that
\begin{equation}
    \hat\x_{s \rightarrow d} = \G_\text{base}(\x_s, \x_d).
\end{equation} 

We use multiple loss functions for training, which can be split into two groups. The first group consists of the standard training objectives for image synthesis. These include perceptual~\cite{Johnson2016PerceptualLF} and GAN~\cite{wang2018pix2pixHD} losses that match the predicted image $\hat\x_{s \rightarrow d}$ to the ground-truth $\x_d$. The other objective regularizes the training and introduces disentanglement between the motion and canonical space appearance features via the cycle consistency~\cite{Zhu2017UnpairedIT} loss.

\textit{Perceptual losses} match the motion and appearance of the predicted image $\hat\x_{s \rightarrow d}$ to the ground-truth $\x_d$. We use three types of pre-trained networks for the perceptual losses: regular ILSVRC (ImageNet)~\cite{Deng2009ImageNetAL} pre-trained VGG19~\cite{Simonyan2015VeryDC} to match the general content of the images, VGGFace~\cite{Parkhi2015DeepFR} trained for face recognition to match the facial appearance, and a specialized gaze loss based on VGG16 to match the gaze direction. The latter network was trained to distill a state-of-the-art gaze detection system~\cite{Fischer2018RTGENERE}. For more details on the training and usage of gaze loss, please refer to the supplementary materials. We calculate the weighted L1 distance between the feature maps obtained for the predicted $\hat\x_{s \rightarrow d}$ and ground-truth $\x_d$ images using all these networks. The final perceptual loss is a weighted combination of individual perceptual losses:
\begin{equation}
    \mathcal{L}_\text{per} = w_\text{IN} \mathcal{L}_\text{IN} + w_\text{face} \mathcal{L}_\text{face} + w_\text{gaze} \mathcal{L}_\text{gaze}.
\end{equation}

\textit{Adversarial losses} ensure the realism of the predicted images. We calculate these losses using the same predicted and driving images. Following the previous works, we train a multi-scale patch discriminator~\cite{Zhu2017UnpairedIT} with a hinge adversarial loss alongside the generator $\G_\text{base}$. We also include a standard feature-matching loss~\cite{wang2018pix2pixHD} to improve the training stability. The GAN loss for the generator can therefore be expressed as follows:
\begin{equation}
    \mathcal{L}_\text{GAN} = w_\text{adv} \mathcal{L}_\text{adv} + w_\text{FM} \mathcal{L}_\text{FM}.
\end{equation}

\textit{Cycle consistency loss} is used to prevent the appearance leakage through the motion descriptor. During training, this task is essential since the motion descriptor is calculated using the same image as the ground truth. Without this regularizer, severe artifacts are present when the driver differs from the source in lighting, hair and beard style, or sunglasses because these features are leaked from the driver image onto the predicted image.

In order to calculate this loss, we use an additional source-driving pair $\x_{s^*}$ and $\x_{d^*}$, which is sampled from a different video and therefore has different appearance from the current $\x_s$, $\x_d$ pair. We then apply the full base model to produce the following \textit{cross-reenacted} image: $\hat\x_{s^* \rightarrow d} = \G_\text{base}(\x_{s^*}, \x_d)$, and also separately calculate a motion descriptor $\z_{d^*} = \E_\text{mtn}(\x_{d^*})$. Note that we will also use the stored motion descriptors $\z_{s^* \rightarrow d}$ and $\z_{s \rightarrow d}$ from the respective forward passes of the base network.

We then arrange the motion descriptors into \textit{positive pairs} $\mathcal{P}$ that should align with each other: $\mathcal{P} = \big\{ ( \z_{s \rightarrow d}, \z_d ), ( \z_{s^* \rightarrow d}, \z_d ) \big\} $, and the \textit{negative pairs}: $ \mathcal{N} = \big\{ ( \z_{s \rightarrow d}, \z_{d^*} ), ( \z_{s^* \rightarrow d}, \z_{d^*} ) \big\}$. These pairs are used to calculate the following cosine distance:
\begin{equation}
    d(\z_i, \z_j) = s \cdot \big( \langle \z_i, \z_j \rangle - m \big),
\end{equation}
where both $s$ and $m$ are hyperparameters. This distance is then used to calculate a large margin cosine loss (CosFace)~\cite{Wang2018CosFaceLM}:
\begin{equation}
    \mathcal{L}_\text{cos} =
    - \hspace{-0.3cm} \sum\limits_{(\z_k, \z_l) \in \mathcal{P}} \hspace{-0.3cm} \log \dfrac{ \exp \big\{ d(\z_k, \z_l) \big\} }{ \exp \big\{ d(\z_k, \z_l) \big\} + \sum\limits_{(\z_i, \z_j) \in \mathcal{N}} \exp \big\{ d(\z_i, \z_j) \big\} }.
\end{equation}

To conclude, the total loss which is used to train the base model is the sum of individual losses:
\begin{equation}
    \mathcal{L}_\text{base} = \mathcal{L}_\text{per} + \mathcal{L}_\text{GAN} + w_\text{cos} \mathcal{L}_\text{cos}.
\end{equation}

These losses are calculated using only foreground regions in both predictions and the ground truth. Hence, our model has no background generation built into it, which we found empirically to hinder its performance. Instead, we impose the background post-training via pre-trained inpainting and matting models. We obtain the background plate using a state-of-the-art inpainting system~\cite{suvorov2021resolution} and use the following systems for matting~\cite{MODNet, Gong2019GraphonomyUH}. The background is combined with the predicted image via alpha-compositing using a calculated matte. For more details, please refer to the supplementary materials.

\subsection{High-resolution model}

For the second training stage, we fix the base neural head avatar model $\G_\text{base}$, and only train an image-to-image translation network $\G_\text{enh}$ that maps the input $\hat\x$ at the resolution $512 \times 512$ to an \textit{enhanced} version $\hat\x^\text{HR}$ that has the resolution $1024 \times 1024$. We use a high-resolution dataset of photographs~\cite{Karras2019ASG} to train this model, in which we assume all images to have different identities. It implies that we cannot form source-driver pairs that only differ in their motion as we do in the first training stage.

The high-resolution model is trained using two groups of loss functions. The first group represents the standard super-resolution objectives, for which use an $L_1$ loss, denoted as $\mathcal{L}_\text{MAE}$, and a GAN loss $\mathcal{L}_\text{GAN}$. The second group of objectives works in an unsupervised way, and we use it to ensure that our model performs well for the images generated in a cross-driving scenario. To do that, for each training image $\x^\text{HR}$ we sample an additional image $\x_\text{c}^\text{HR}$, and generate its initial reconstruction $\hat\x_\text{c} = \G_\text{base}(\x^\text{LR}, \x^\text{LR}_\text{c})$, where $\x_\text{c}^\text{LR}$ is used to estimate motion, and $\x^\text{LR}$ is used to estimate appearance. Since we do not have high-resolution ground-truth for $\hat\x_\text{c}^\text{HR} = \G_\text{enh}(\hat\x_\text{c})$, we can only match its distribution to ground truth using a patch discriminator. Furthermore, we can enforce content preservation by applying the cycle-consistency loss at lower resolution:
\begin{equation}
    \mathcal{L}_\text{cyc}^\text{c} = \mathcal{L}_\text{MAE} \big( \text{DS}_{4}(\hat\x_\text{c}), \text{DS}_{8}( \hat\x_\text{c}^\text{HR} ) \big),
\end{equation}
where $\text{DS}_{k}$ denotes a $k$-times downsampling operator.

The final objective for $\G_\text{enh}$ includes the adversarial and the perceptual losses calculated for the predicted image $\hat\x^\text{HR}$ and its ground-truth $\x^\text{HR}$, as well as an adversarial loss $\mathcal{L}_\text{adv}^\text{c}$, calculated for $\hat\x_c^\text{HR}$ and $\x^\text{HR}$, and the cycle-consistency loss $\mathcal{L}_\text{cyc}^\text{c}$:
\begin{equation}
    \mathcal{L}_\text{enh} = \mathcal{L}_\text{GAN} + w_\text{MAE} \mathcal{L}_\text{MAE} + w_\text{adv}^\text{c} \mathcal{L}_\text{adv}^\text{c} + w_\text{cyc}^\text{c} \mathcal{L}_\text{cyc}^\text{c}.
\end{equation}

\subsection{Student model}

Finally, we use a small conditional image-to-image translation network $\G_\text{DT}$, which we refer to as the \textit{student}, to distill the one-shot model. We train the student to mimic the prediction of the full (teacher) model $\G_\text{HR} = \G_\text{enh} *\, \G_\text{base}$, which combines the base model and an enhancer. The student is trained only in the cross-driving mode by generating pseudo-ground truth with the teacher model. Since we train our student network for a limited number of avatars, we condition it using an index $i$, which selects an image from the set of all $N$ appearances $\{ \x_i \}_{i=1}^N$. Therefore, training proceeds as follows: we sample the driving frame $\x_d$ and the index $i$. We then match the following two images:
\begin{align*}
    & \hat\x_{i \rightarrow d}^\text{DT} = \G_\text{DT}(\x_d, i);
    & \hat\x_{i \rightarrow d}^\text{HR} = \G_\text{HR}(\x_i, \x_d).
\end{align*}
We train this network using a combination of perceptual and adversarial losses. For architectural details, please refer to the supplementary materials.

\begin{figure*}
    \centering    
    \setlength{\wid}{0.225\textwidth}
    \setlength{\mrg}{-0.3cm}
    \resizebox{1.0\linewidth}{!}{
    \begin{tabular}{cccc}
        \includegraphics[width=\wid]{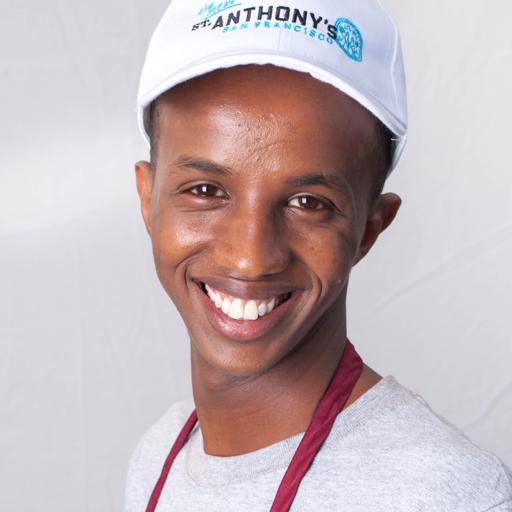} &
        \hspace{\mrg}
        \includegraphics[width=\wid]{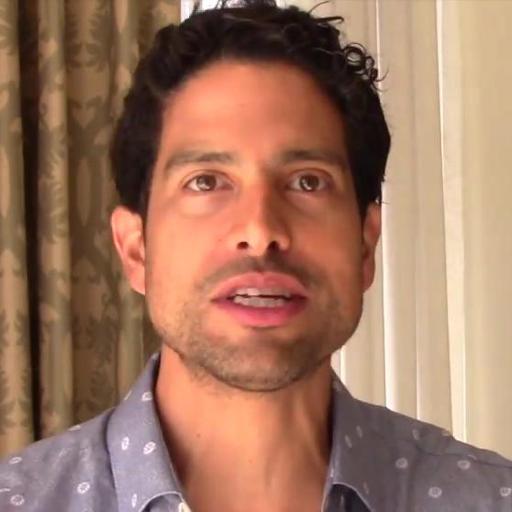} &
        \hspace{\mrg}
        \includegraphics[width=\wid]{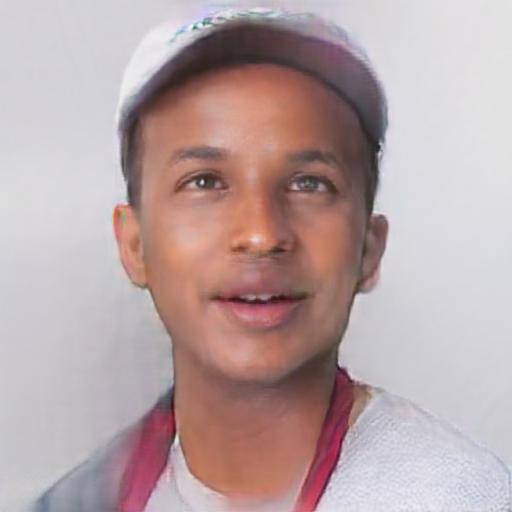} & 
        \hspace{\mrg}
        \includegraphics[width=\wid]{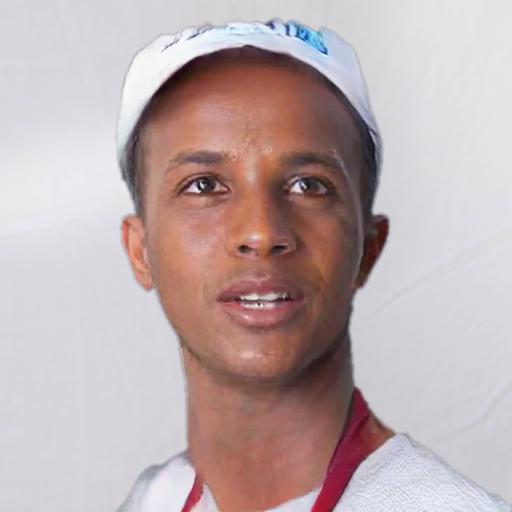}
        \\ %
        \includegraphics[width=\wid]{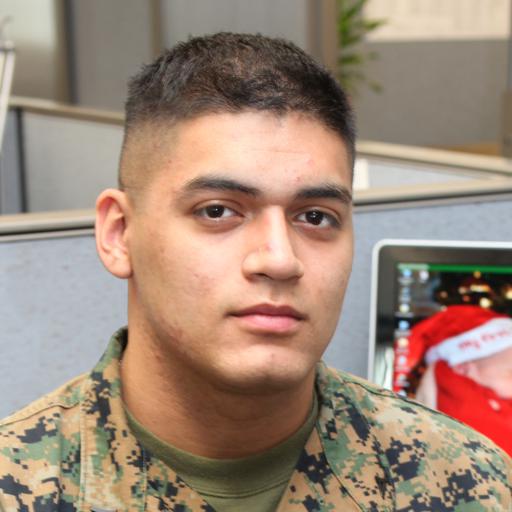} &
        \hspace{\mrg}
        \includegraphics[width=\wid]{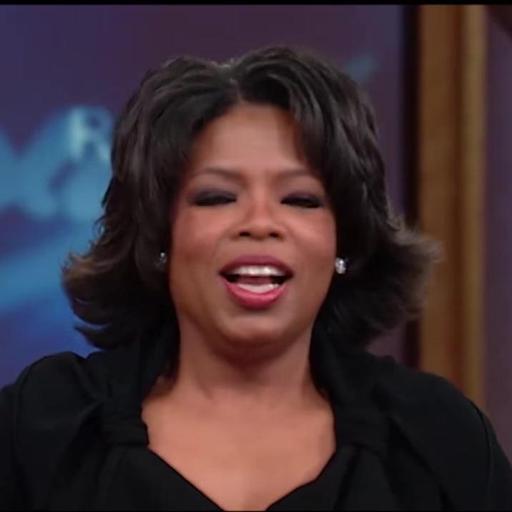} &
        \hspace{\mrg}
        \includegraphics[width=\wid]{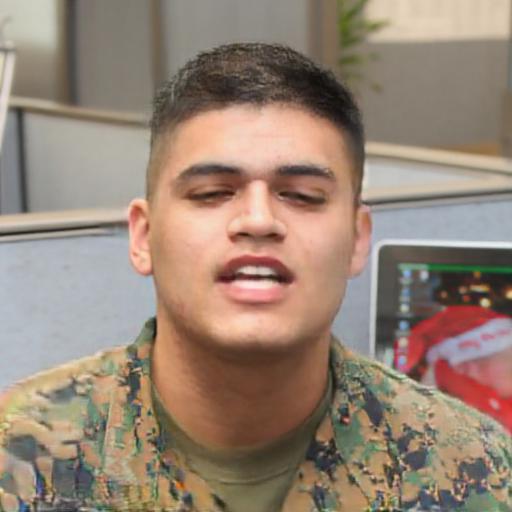} & 
        \hspace{\mrg}
        \includegraphics[width=\wid]{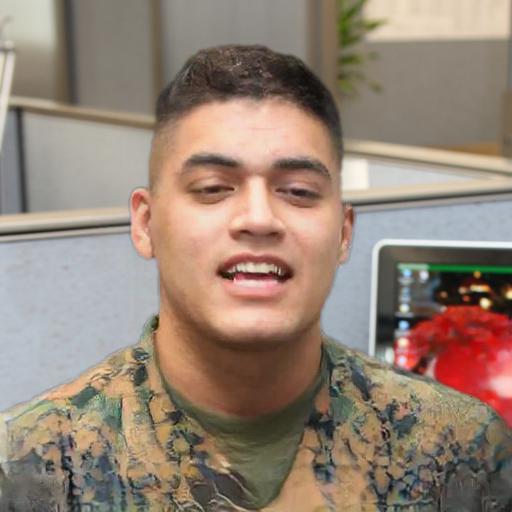}
        \\
        \includegraphics[width=\wid]{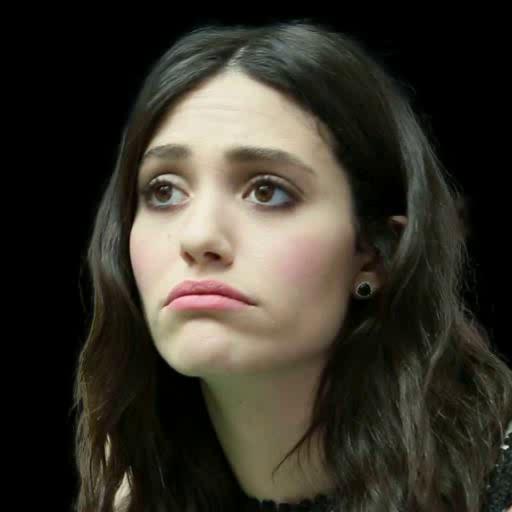} & 
        \hspace{\mrg}
        \includegraphics[width=\wid]{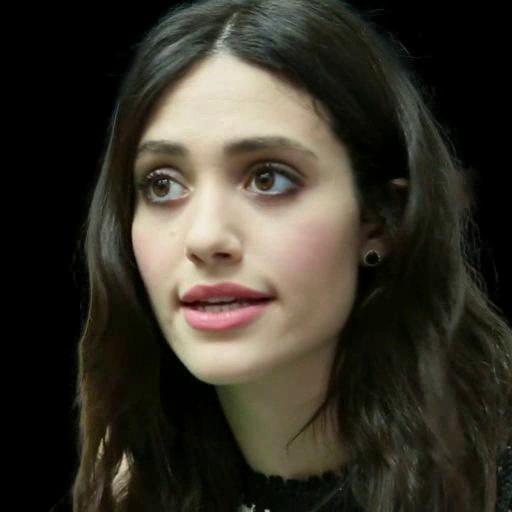} & 
        \hspace{\mrg}
        \includegraphics[width=\wid]{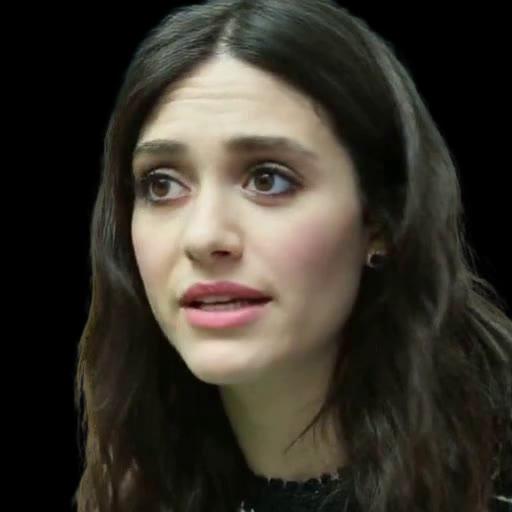} & 
        \hspace{\mrg}
        \includegraphics[width=\wid]{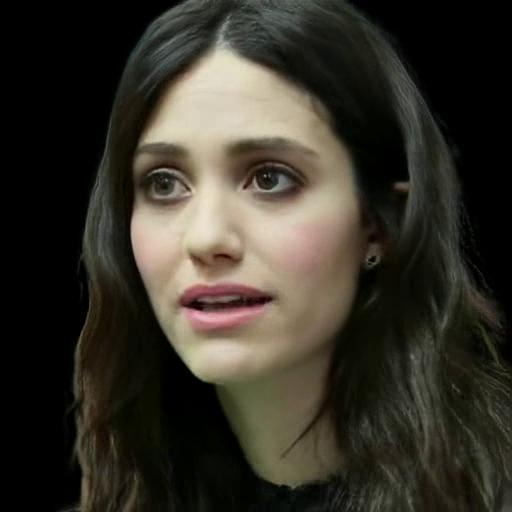}
        \\ %
        \textbf{Source} & 
        \hspace{\mrg} 
        \textbf{Driver} & 
        \hspace{\mrg} 
        \textbf{Face-V2V}~\cite{Wang2021OneShotFN} & 
        \hspace{\mrg} 
        \textbf{Ours} 
    \end{tabular}}
    \vspace{-0.4cm}
    \caption{A qualitative comparison of head avatar systems in cross-reenactment scenario (top two rows) and self-reenactment scenario (bottom row) at 512px resolution. In cross-reenactment, we can see that our approach achieves better preservation of motion and appearance than the previous state-of-the-art (Face-V2V). In self-reenactment, we achieve the results of comparable quality with the state-of-the-art. For more examples, please refer to the supplementary materials.}
    \label{fig:512px_qual}
\end{figure*}
\begin{figure*}[!ht]
    \centering    
    \setlength{\wid}{0.18\textwidth}
    \setlength{\mrg}{-0.3cm}
    \resizebox{1\linewidth}{!}{
    \begin{tabular}{ccccc}
        \includegraphics[width=\wid]{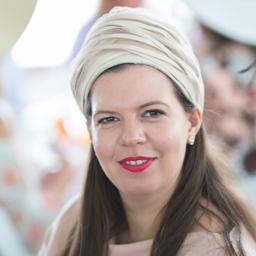} & 
        \hspace{\mrg}
        \includegraphics[width=\wid]{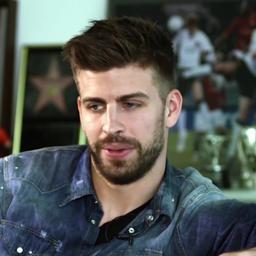} & 
        \hspace{\mrg}
        \includegraphics[width=\wid]{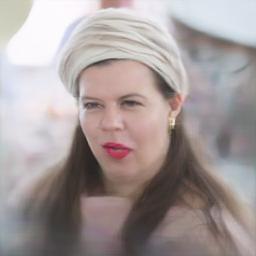} & 
        \hspace{\mrg}
        \includegraphics[width=\wid]{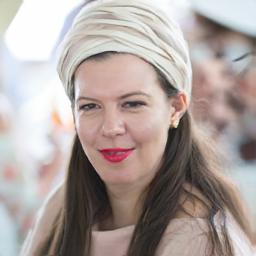} & 
        \hspace{\mrg}
        \includegraphics[width=\wid]{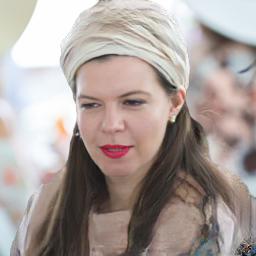} 
        \\ %
        \includegraphics[width=\wid]{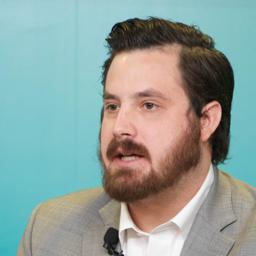} & 
        \hspace{\mrg}
        \includegraphics[width=\wid]{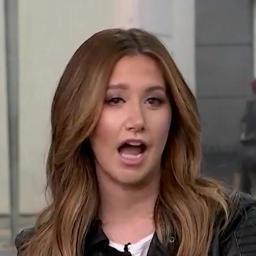} & 
        \hspace{\mrg}
        \includegraphics[width=\wid]{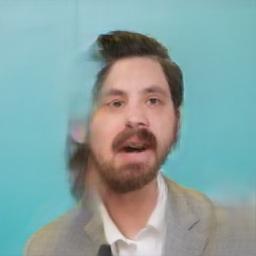} & 
        \hspace{\mrg}
        \includegraphics[width=\wid]{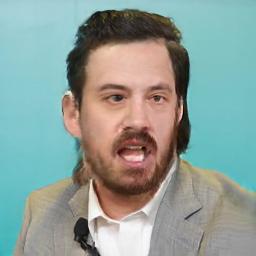} & 
        \hspace{\mrg}
        \includegraphics[width=\wid]{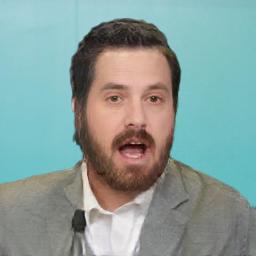} 
        \\
        \includegraphics[width=\wid]{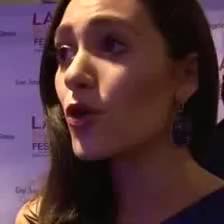} & 
        \hspace{\mrg}
        \includegraphics[width=\wid]{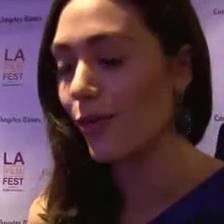} & 
        \hspace{\mrg}
        \includegraphics[width=\wid]{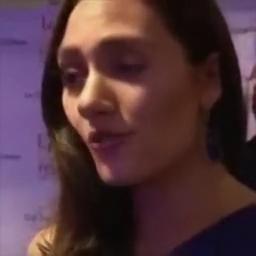} & 
        \hspace{\mrg}
        \includegraphics[width=\wid]{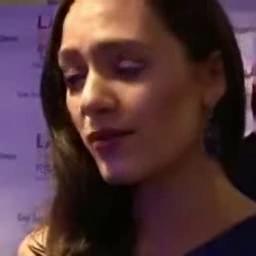} &
        \hspace{\mrg}
        \includegraphics[width=\wid]{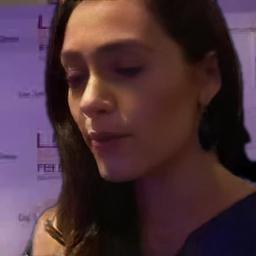} 
        \\
        \textbf{Source} & 
        \hspace{\mrg} 
        \textbf{Driver} & 
        \hspace{\mrg} 
        \textbf{FOMM}~\cite{Siarohin2019FirstOM} & 
        \hspace{\mrg} 
        \textbf{HeadGAN}~\cite{Doukas2021HeadGANON} & 
        \hspace{\mrg} 
        \textbf{Ours}
    \end{tabular}}
    \vspace{-0.4cm}
    \caption{A qualitative comparison of head avatar systems in cross-reenactment scenario (top two rows) and self-reenactment scenario (bottom row) at $256 \times 256$ resolution. Our system significantly outperforms the competitors in cross-reenactment, achieving more faithful motion and appearance preservation in the generated images. We also show that our system achieves similar results in self-reenactment. For more examples, please refer to the supplementary materials.}
    \label{fig:256px_qual}
\end{figure*}

\section{Experiments}

We use multiple datasets to train and evaluate our model: VoxCeleb2~\cite{Chung2018VoxCeleb2DS} and VoxCeleb2HQ video datasets, and FFHQ~\cite{Karras2019ASG} image dataset. We have obtained a high-quality version of the VoxCeleb2 dataset, which we refer to as VoxCeleb2HQ, by downloading the original videos and filtering them using both bitrate and image quality assessment~\cite{hyperIQA}. This leaves approximately one-tenth of the original dataset (15,000 videos). We use this dataset to train and evaluate our base model at $512 \times 512$ resolution while using the original VoxCeleb2 dataset, filtered using bitrate, for the $256 \times 256$ resolution. For training a high-resolution model, we used a filtered version of the FFHQ dataset, which consists of 20,000 images and has no frames that contain multiple people or children. Lastly, we use a proprietary dataset of 20,000 selfie videos and 100,000 selfie pictures to train the student model.

\subsection{Training details}

We trained the $256 \times 256$ model for 200,000 iterations with the batch size of 24, and the $512 \times 512$ model for 300,000 iterations with the batch size of 16. We used AdamW~\cite{Loshchilov2019DecoupledWD} optimizer with cosine learning rate scheduling. The initial learning rate was reduced from $2*10^{-4}$ to $10^{-6}$ during training iterations. We used the following hyperparameters for the losses: $w_\text{IN} = 20, w_\text{face} = 4, w_\text{gaze} = 5, w_\text{adv} = 1, w_\text{FM} = 40,$ and $w_\text{cos} = 2$. We also set $s = 5$ and $m = 0.2$ in the cosine loss.

We trained the high-resolution enhancer model for 50,000 iterations with the batch size of 16. We used the same optimizer and the learning rate scheduling. We set the loss weights to $w_\text{MAE} = 100, w_\text{adv}^\text{c} = 1, w_\text{FM} = 100$ and $w_\text{cyc}^\text{c} = 10$. Finally, for the student model we distilled 100 avatars. We trained it for 170,000 iterations with the batch size of 8. For detailed descriptions of all architectures, please refer to the supplementary material.

\subsection{Baseline methods}

We compare our base model with the following systems.

Face Vid-to-vid (Face-V2V)~\cite{Wang2021OneShotFN} is a state-of-the-art system in self-reenactment, i.e. when the source and driving images have the same appearance and identity. Its main features are the volumetric encoding of the avatar's appearance and the explicit representation of the head motion with 3D keypoints, which are learned in an unsupervised way. In our base model, we utilize a similar volumetric encoding of the appearance but instead encode the face motion implicitly, which improves cross-reenactment performance.

First Order Motion Model (FOMM)~\cite{Siarohin2019FirstOM} uses 2D keypoints to represent motion and is another strong baseline for the task of self-reenactment. Similar to Face-V2V, these keypoints are trained in an unsupervised way. However, as shown in our evaluation, this method fails to generate realistic images in the cross-reenactment scenario.

Lastly, we compare against the HeadGAN~\cite{Doukas2021HeadGANON} system, in which the expression coefficients of the 3D morphable model~\cite{Blanz1999AMM} are used as a motion representation. These coefficients are calculated using a pre-trained dense 3D keypoints regressor~\cite{Deng2020RetinaFaceSM}. Effectively, this approach disentangles motion data from the appearance in the 3D keypoints, but limits the space of possible motions (for example, it does not allow the control of the gaze direction).

\subsection{Cross-reenactment evaluation}

Since pre-trained models of FOMM and HeadGAN are only available at $256 \times 256$ resolution, we compare them against our base model trained on a bitrate-filtered VoxCeleb2 dataset. For Face-V2V, we compare the $512 \times 512$ model pre-trained on the TalkingHead-1KH~\cite{Wang2021OneShotFN} dataset to our base model trained on the VoxCeleb2HQ. For the evaluation, we use samples from the VoxCeleb2HQ and FFHQ datasets, downscaled to the training resolution.

For quantitative evaluation, we use the following metrics. \textit{Frechet Inception Distance} (FID)~\cite{fid} is used to compare the distributions of predicted images and the images in the dataset. \textit{Cosine similarity} between the embeddings of a face recognition network (CSIM)~\cite{Zakharov2019FewShotAL} is used to evaluate the preservation of a person's appearance in the predicted image. Finally, we conduct two \textit{user studies} (denoted as UMTN and UAPP) to evaluate the motion and appearance preservation. We show the crowd-sourced users a random triplet of images: a driving example to evaluate motion preservation or a source example to evaluate the appearance, alongside the outputs of two random methods. We then ask each user to pick one of the two outputs with the better-preserved motion or appearance. We then measure the percentage of examples where each method was picked. We conducted our experiment on approximately 2,000 crowd-sourced people, and each evaluation sample was shown, on average, to twenty different users.

The qualitative results are shown in Figures~\ref{fig:512px_qual}-\ref{fig:256px_qual}, and the quantitative metrics are presented in Table~\ref{tab:base_quant}. Overall, we can see that our method outperforms all competitors by some margin. Furthermore, the first two rows in Figure~\ref{fig:256px_qual} suggest that our approach is better at preserving the shape and appearance of the source image and the motion of the driver image, including gaze direction, than the FOMM and HeadGAN systems. Compared to the Face V2V system (Figure~\ref{fig:512px_qual}, first two rows), our implicit pose representation approach prevents appearance leakage through the driving image, leading to better preservation of the source image appearance, as well as driver motion. These observations are confirmed by the quantitative evaluation, in which we outperform our competitors across all cross-reenactment metrics (Table~\ref{tab:base_quant}), including both user studies.

\begin{table}
\setlength{\mrg}{0.1cm}
\setlength\tabcolsep{3pt}
\centering
\resizebox{0.9\linewidth}{!}{
\begin{tabular}{l cccc cccc cccc}
    \multicolumn{13}{c}{\textbf{Cross-reenactment}}
    \\
    Method
    &
    \multicolumn{3}{c}{FID$\downarrow$}
    &
    \multicolumn{3}{c}{CSIM$\uparrow$}
    &
    \multicolumn{3}{c}{UMTN$\uparrow$}
    &
    \multicolumn{3}{c}{UAPP$\uparrow$}
    \\
    \hline
    \multicolumn{13}{c}{VoxCeleb2HQ \& FFHQ ($256 \times 256$)}
    \\
    \hline
    FOMM    & \multicolumn{3}{c}{79.1}
    & \multicolumn{3}{c}{0.63} & \multicolumn{3}{c}{24.0} & \multicolumn{3}{c}{27.9}
    \\
    HeadGAN & \multicolumn{3}{c}{70.0}
    & \multicolumn{3}{c}{0.66} & \multicolumn{3}{c}{23.6} & \multicolumn{3}{c}{32.1} 
    \\
    Ours    & \multicolumn{3}{c}{68.9}
    & \multicolumn{3}{c}{0.72} & \multicolumn{3}{c}{52.4} & \multicolumn{3}{c}{40.0}
    \\
    \hline
    \multicolumn{13}{c}{VoxCeleb2HQ \& FFHQ ($512 \times 512$)}
    \\
    \hline
    Face-V2V & \multicolumn{3}{c}{63.4}
    & \multicolumn{3}{c}{0.70} & \multicolumn{3}{c}{34.4} & \multicolumn{3}{c}{45.4}
    \\
    Ours     & \multicolumn{3}{c}{58.8}
    & \multicolumn{3}{c}{0.73} & \multicolumn{3}{c}{65.6} & \multicolumn{3}{c}{54.6}
    \\
    \multicolumn{13}{c}{\textbf{Self-reenactment (raw / masked)}}
    \\
    Method
    &
    \multicolumn{4}{c}{PSNR$\uparrow$}
    &
    \multicolumn{4}{c}{SSIM$\uparrow$}
    &
    \multicolumn{4}{c}{LPIPS$\downarrow$}
    \\
    \hline
    \multicolumn{13}{c}{VoxCeleb2 ($256 \times 256$)}
    \\
    \hline
    FOMM    & \multicolumn{4}{c}{20.6 / 27.5} & \multicolumn{4}{c}{0.74 / 0.90} & \multicolumn{4}{c}{0.18 / 0.06}
    \\
    HeadGAN & \multicolumn{4}{c}{18.6 / 26.5} & \multicolumn{4}{c}{0.68 / 0.88} & \multicolumn{4}{c}{0.20 / 0.07}
    \\
    Ours    & \multicolumn{4}{c}{18.3 / 27.0} & \multicolumn{4}{c}{0.67 / 0.89} & \multicolumn{4}{c}{0.23 / 0.07}
    \\
    \hline
    \multicolumn{13}{c}{VoxCeleb2HQ ($512 \times 512$)}
    \\
    \hline
    Face-V2V & \multicolumn{4}{c}{21.9 / 31.2} & \multicolumn{4}{c}{0.76 / 0.90} & \multicolumn{4}{c}{0.18 / 0.06} 
    \\
    Ours     & \multicolumn{4}{c}{20.2 / 30.2} & \multicolumn{4}{c}{0.72 / 0.89} & \multicolumn{4}{c}{0.22 / 0.07}
\end{tabular}
}
\caption{Quantitative results for cross and self-reenactment. To evaluate cross-reenactment performance, we measure FID (lower the better), CSIM (higher the better), and user preference scores (UMTN measures motion preservation and UAPP -- appearance, both are higher the better). Our method outperforms its competitors across all metrics at both resolutions, achieving state-of-the-art results in the cross-reenactment scenario. The gap is especially noticeable in the user study, where we achieve significantly better motion preservation. We use standard PSNR, SSIM (higher the better), and LPIPS (lower the better) metrics to evaluate the self-reenactment. We measure each metric using either raw or masked images. Our method performs similarly to the competitors when face masking is applied while achieving reasonable results in the unmasked (raw) scenario.}
\label{tab:base_quant}
\end{table}

\begin{table}
    \centering
    \resizebox{0.8\linewidth}{!}{
    \begin{tabular}{l ccc}
        \multicolumn{4}{c}{\textbf{Cross-reenactment}}
        \\
        Method 
        &
        FID$\downarrow$
        &
        CSIM$\uparrow$
        &
        IQA$\uparrow$
        \\
        \hline
        Base w/ bicubic   & 51.4 & 0.67 & 35.1 \\
        HiFaceGAN        & 49.4 & 0.65 & 43.9 \\
        Ours             & 39.2 & 0.67 & 49.3 \\
    \end{tabular}
    }
    \caption{Quantitative results on the FFHQ dataset in the cross-reenactment mode at $1024 \times 1024$ resolution. Besides the standard cross-reenactment metrics, we additionally perform an image quality assessment (IQA, higher the better). Our super-resolution method improves the resulting image quality compared to the base model with bicubic upsampling and the super-resolution baseline (HiFaceGAN), as seen from the FID and IQA metrics. At the same time, we preserve the source image appearance, which results in the same CSIM as the base model.}
    \label{tab:hr_quant}
\end{table}

\subsection{Self-reenactment evaluation}

We use the same pre-trained models for the self-reenactment experiments as for the cross reenactment and evaluate them on the samples from the VoxCeleb2 and VoxCeleb2HQ evaluation sets. In addition, we use the following standard metrics to measure the difference between the synthesized and ground-truth images: Peak Signal-to-Noise Ratio (PSNR), Structural Similarity Index Measure (SSIM)~\cite{Wang2004ImageQA}, and the Learned Perceptual Image Patch Similarity (LPIPS)~\cite{Zhang2018TheUE}. 

We notice that qualitatively we achieve similar performance to the competitors, especially in the face and hair regions (Figures~\ref{fig:512px_qual}-\ref{fig:256px_qual}, third row). To quantitatively verify that, we have conducted an evaluation using masked data. The masks include the face, ears, and hair regions and are applied to both the target and the predicted images before calculating the metrics. In this scenario, we achieve comparable performance to the baseline methods  (\tab{base_quant}) but have an inferior performance when the unmasked (raw) images are used.

This difference could be caused, among other reasons, by the lack of shoulders motion modeling in our method. It results in the misalignment between our predictions and ground truth in the corresponding regions. We further discuss this issue in the limitations section. Also, our method's high degree of disentanglement between motion and appearance descriptors prevents it from leaking the appearance data directly from the driver, which generally contributes to the reduced performance in self-reenactment.

\subsection{High-resolution evaluation}
\begin{figure*}
    \centering    
    \setlength{\wid}{0.19\textwidth}
    \setlength{\mrg}{-0.3cm}
    \setlength{\mrgv}{0cm}
    \resizebox{1.0\linewidth}{!}{
    \begin{tabular}{ccccc}
        \includegraphics[width=\wid]{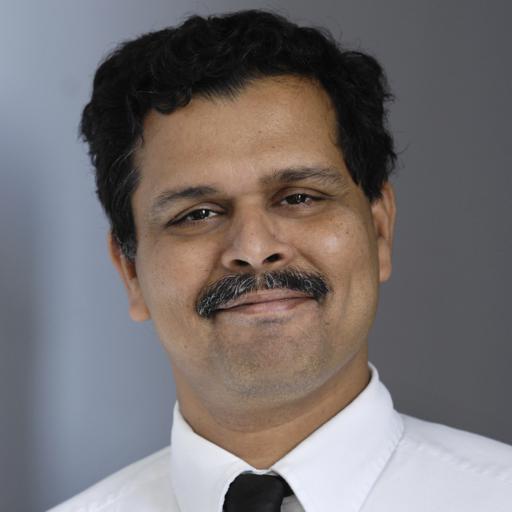} & 
        \hspace{\mrg}
        \vspace{\mrgv}
        \includegraphics[width=\wid]{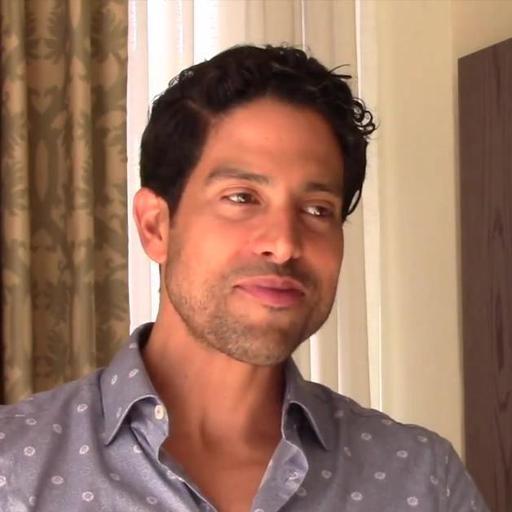} & 
        \hspace{\mrg}
        \includegraphics[width=\wid]{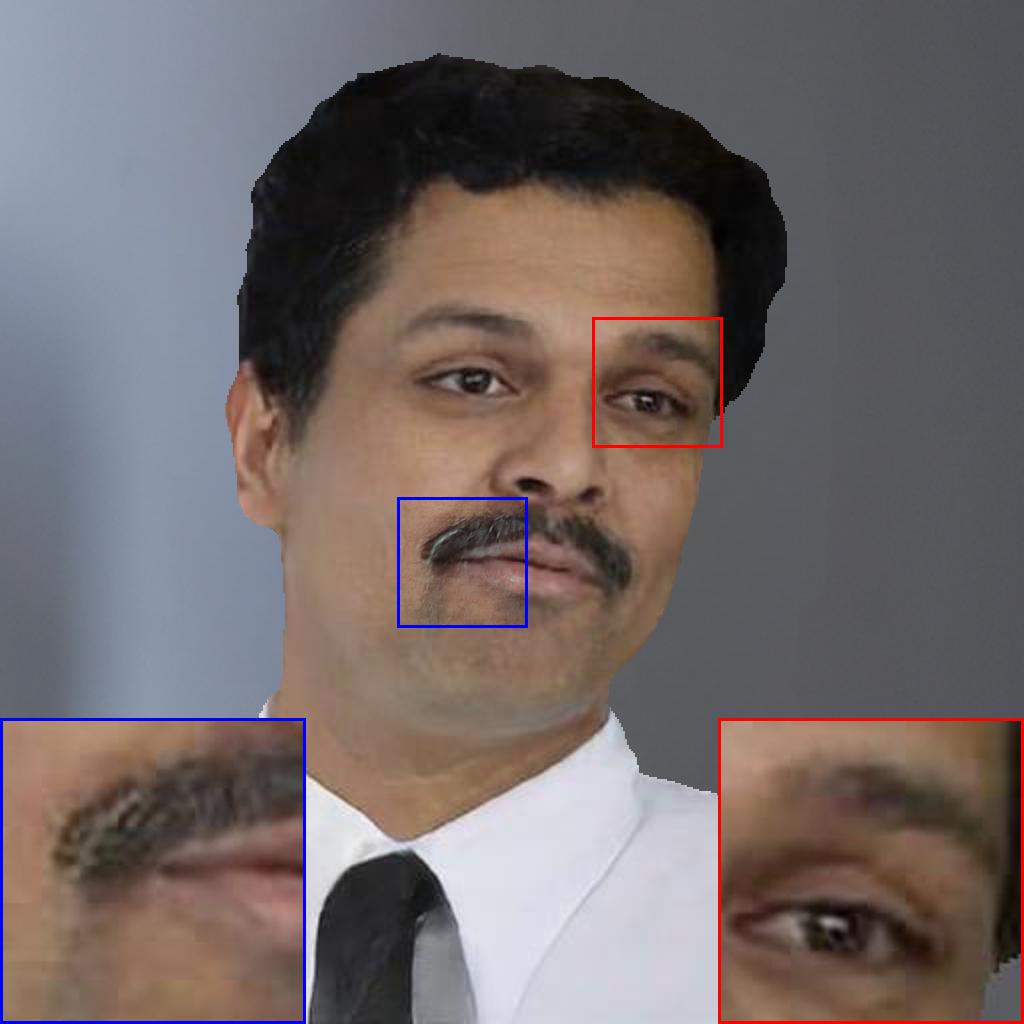} & 
        \hspace{\mrg}
        \includegraphics[width=\wid]{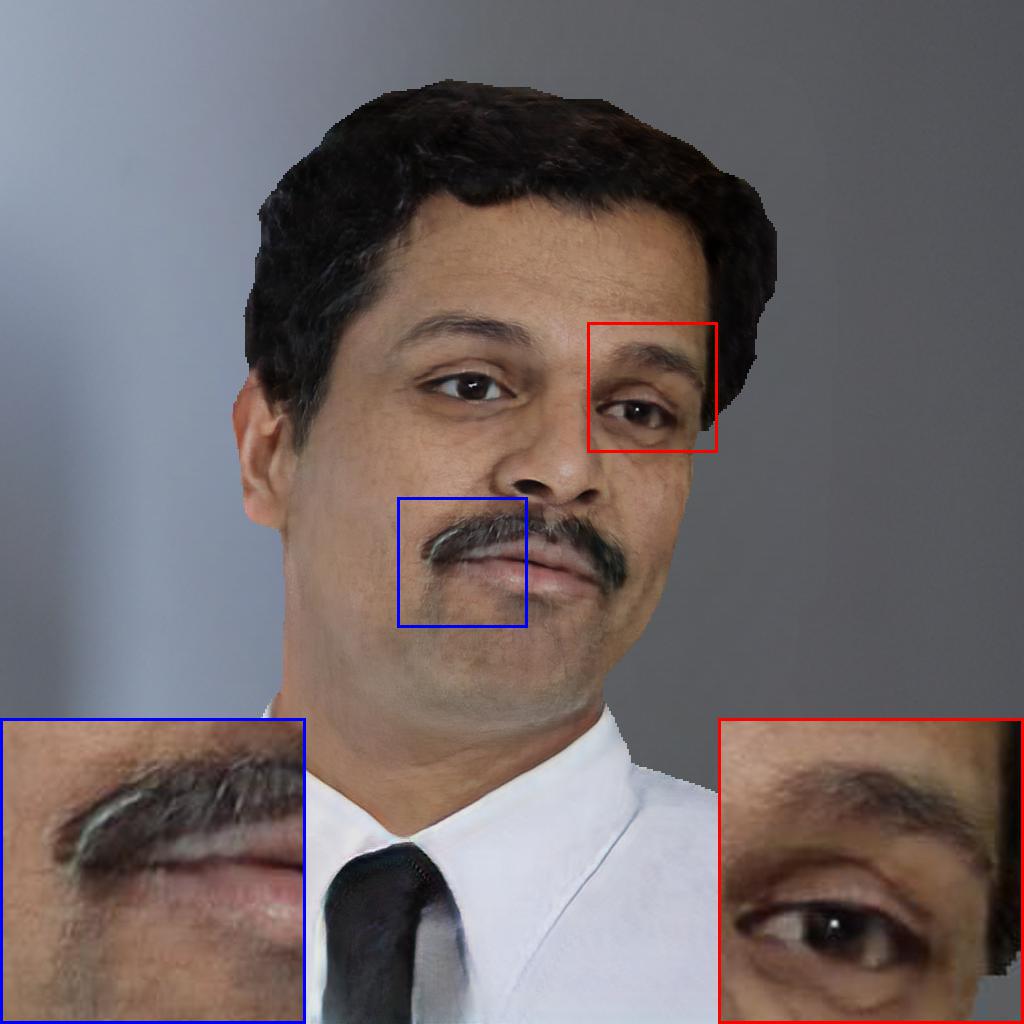} & 
        \hspace{\mrg}
        \includegraphics[width=\wid]{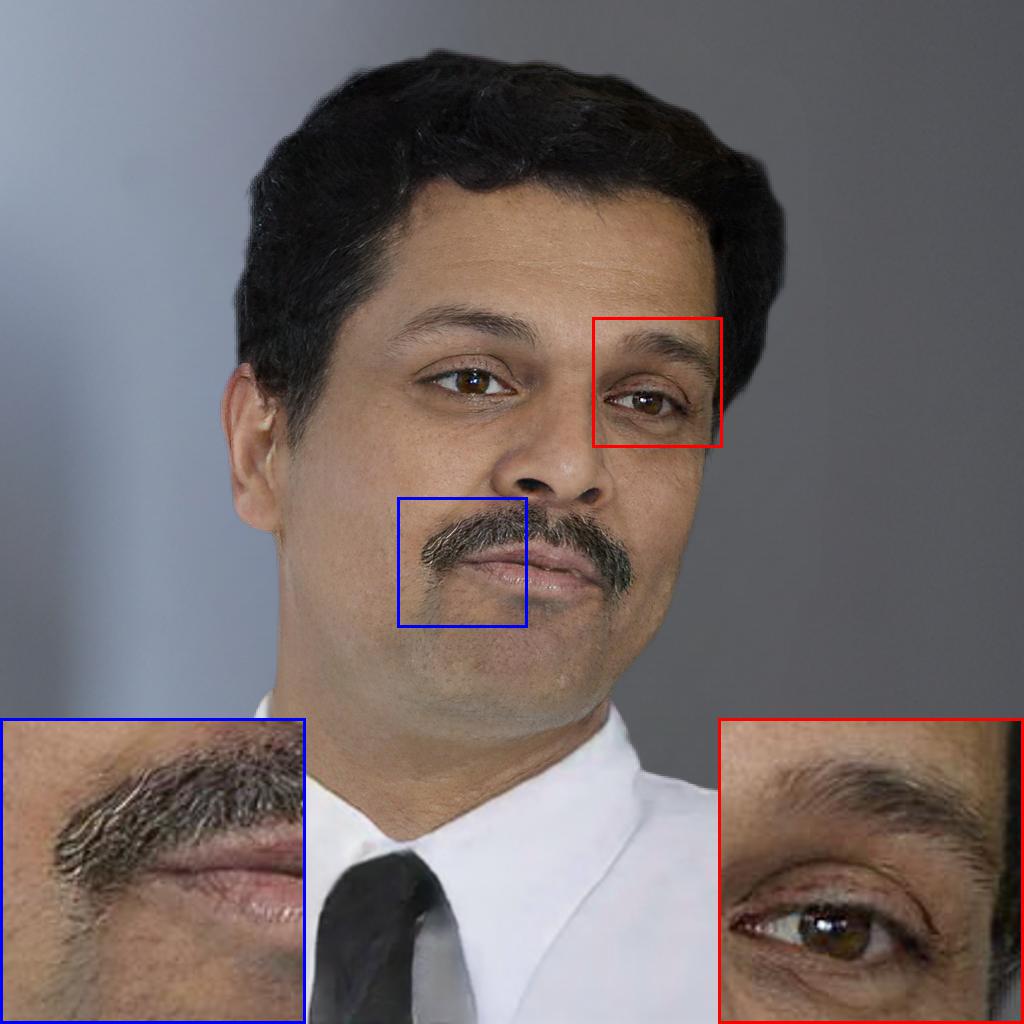} 
        \\ %
        \includegraphics[width=\wid]{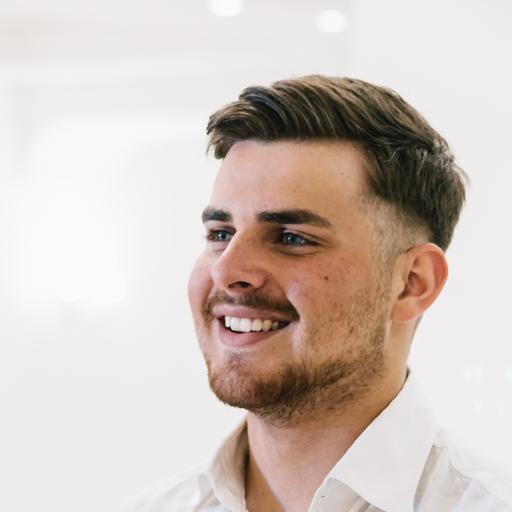} & 
        \hspace{\mrg}
        \vspace{\mrgv}
        \includegraphics[width=\wid]{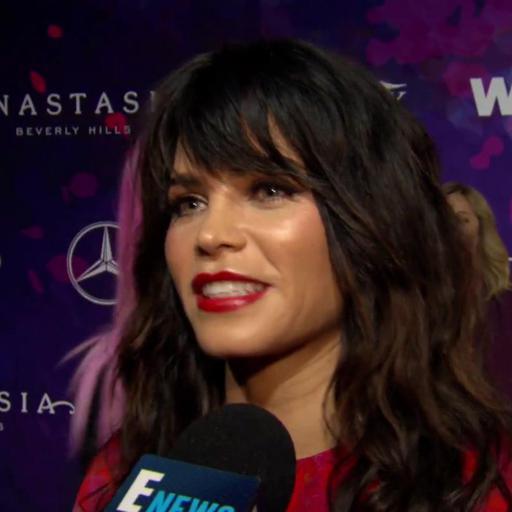} & 
        \hspace{\mrg}
        \includegraphics[width=\wid]{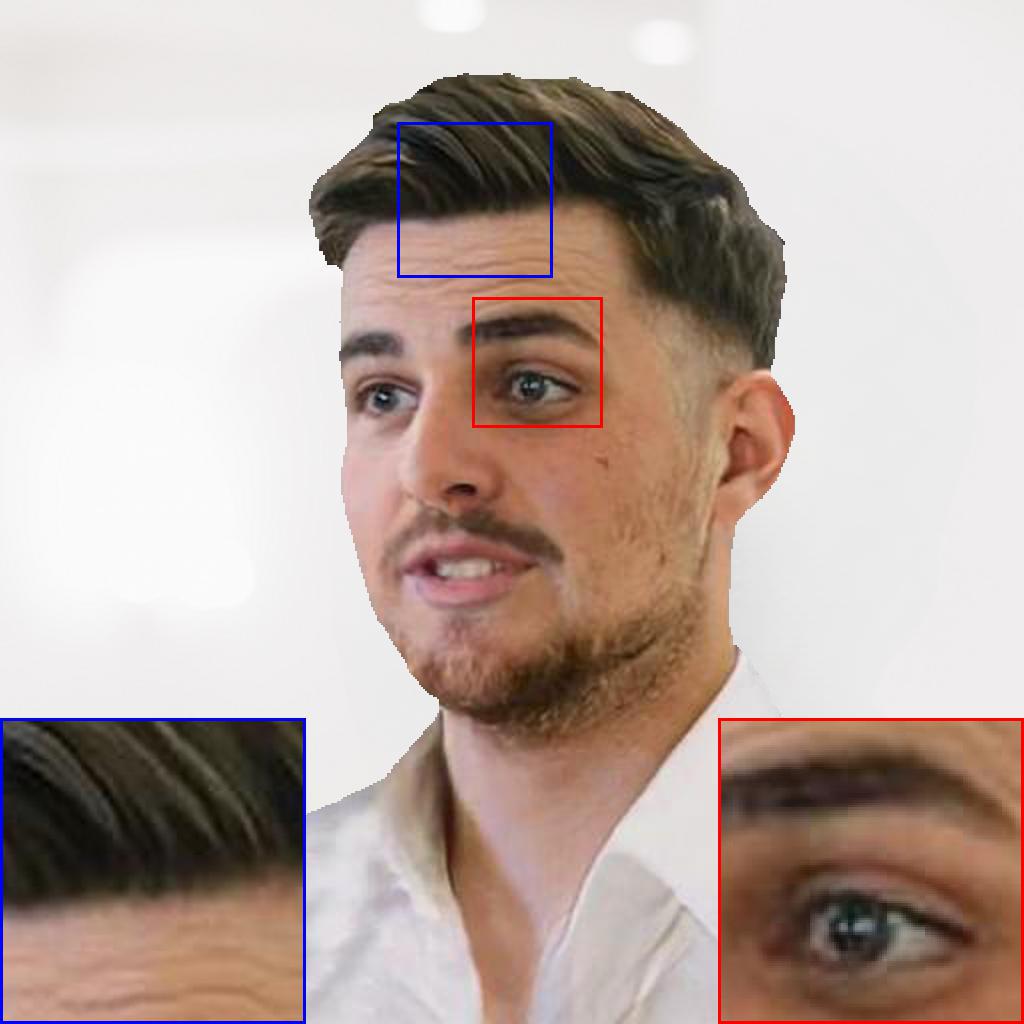} & 
        \hspace{\mrg}
        \includegraphics[width=\wid]{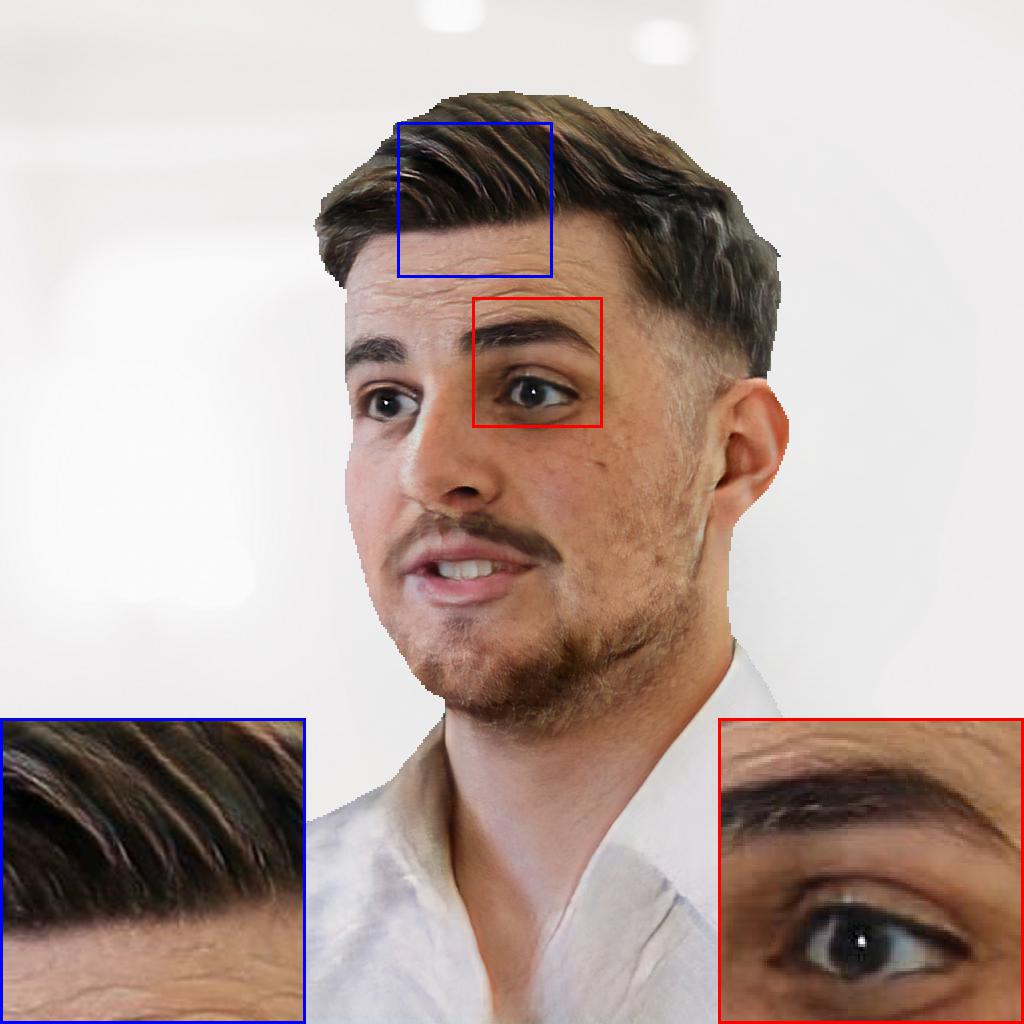} & 
        \hspace{\mrg}
        \includegraphics[width=\wid]{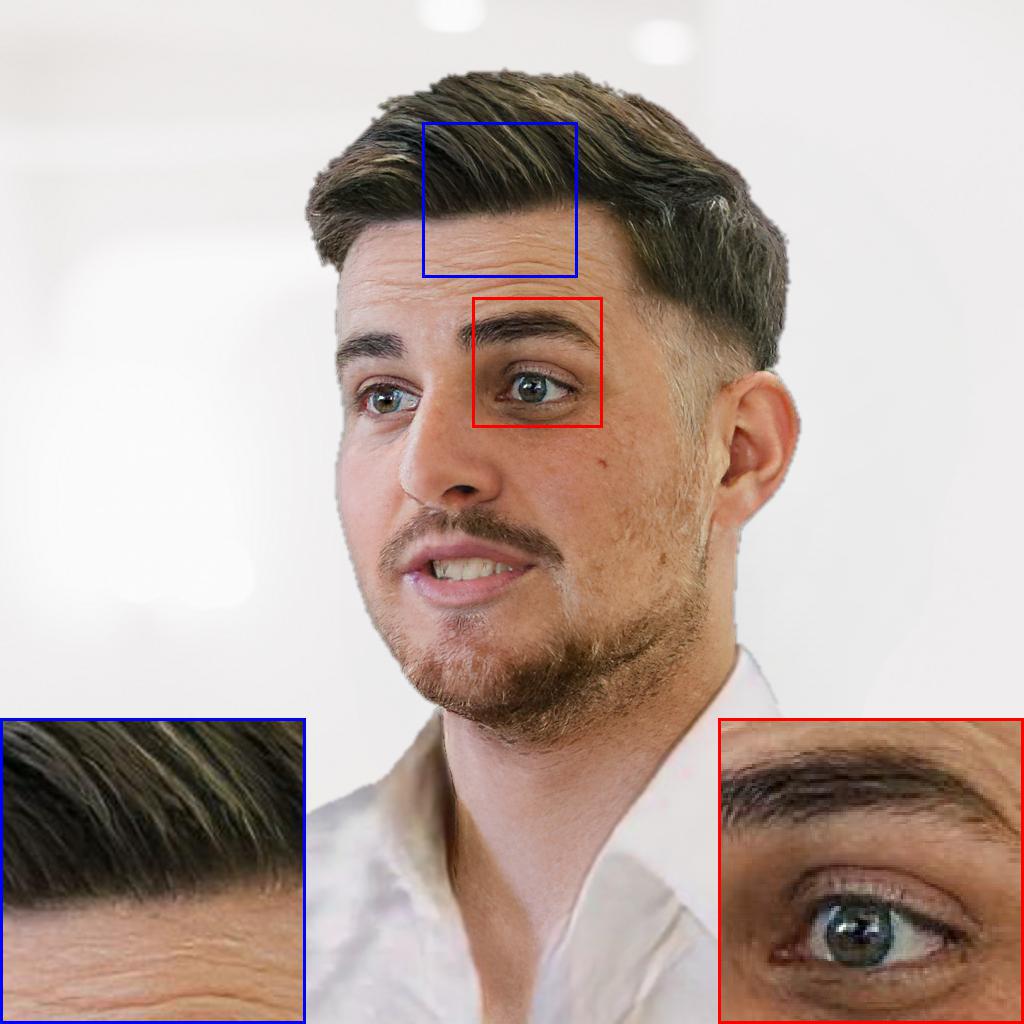} 
        \\ %
        \textbf{Source} & 
        \hspace{\mrg} 
        \textbf{Driver} & 
        \hspace{\mrg} 
        \textbf{Ours (base w/ bicubic)} & 
        \hspace{\mrg} 
        \textbf{HiFaceGAN}~\cite{Yang2020HiFaceGANFR} & 
        \hspace{\mrg} 
        \textbf{Ours (HR)}
    \end{tabular}}
    \vspace{-0.4cm}
    \caption{A qualitative comparison of different super-resolution methods applied to the output of our base model. While performing better than a baseline bicubic upsampling, we can see that the state-of-the-art super-resolution method (HiFaceGAN) cannot achieve the same level of high-frequency details fidelity as our approach. Digital zoom-in is recommended.}
    \label{fig:1024px_qual}
\end{figure*}

We evaluate high-resolution synthesis only in cross-reenactment mode since data for the self-reenactment scenario is missing. We use subsets of a filtered FFHQ dataset for training and evaluation. We train both our and the baseline super-resolution approaches using an output of a pre-trained base model $\G_\text{base}$ as input and by sampling two random augmented versions of the training image as a source and a driver. We use random crops and rotations since other augmentations could change person-specific traits (e.g.~head width).

We compare against two baselines. First, we consider bicubic upsampling of the output of the base model, and second, we evaluate a state-of-the-art face super-resolution system (HiFaceGAN)~\cite{Yang2020HiFaceGANFR}. The results are presented in \fig{1024px_qual}, and \tab{hr_quant}. In the quantitative comparison, we use an additional image quality assessment metric (IQA)~\cite{hyperIQA} to measure the resulting image quality. Our method outperforms its competitors both qualitatively and quantitatively by generating more high-frequency details and, at the same time, preserving the identity of the source image.

\begin{figure}
    \centering    
    \setlength{\wid}{0.111\textwidth}
    \setlength{\mrg}{-0.4cm}
    \setlength{\mrgv}{-0.09cm}
    \begin{tabular}{cccc}
        \vspace{\mrgv}
        \includegraphics[width=\wid]{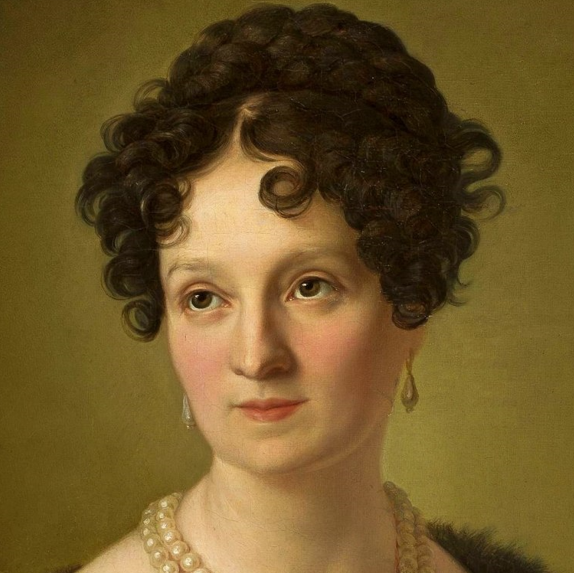} & 
        \hspace{\mrg}
        \includegraphics[width=\wid]{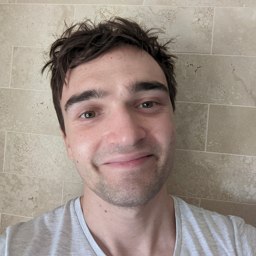} & 
        \hspace{\mrg}
        \includegraphics[width=\wid]{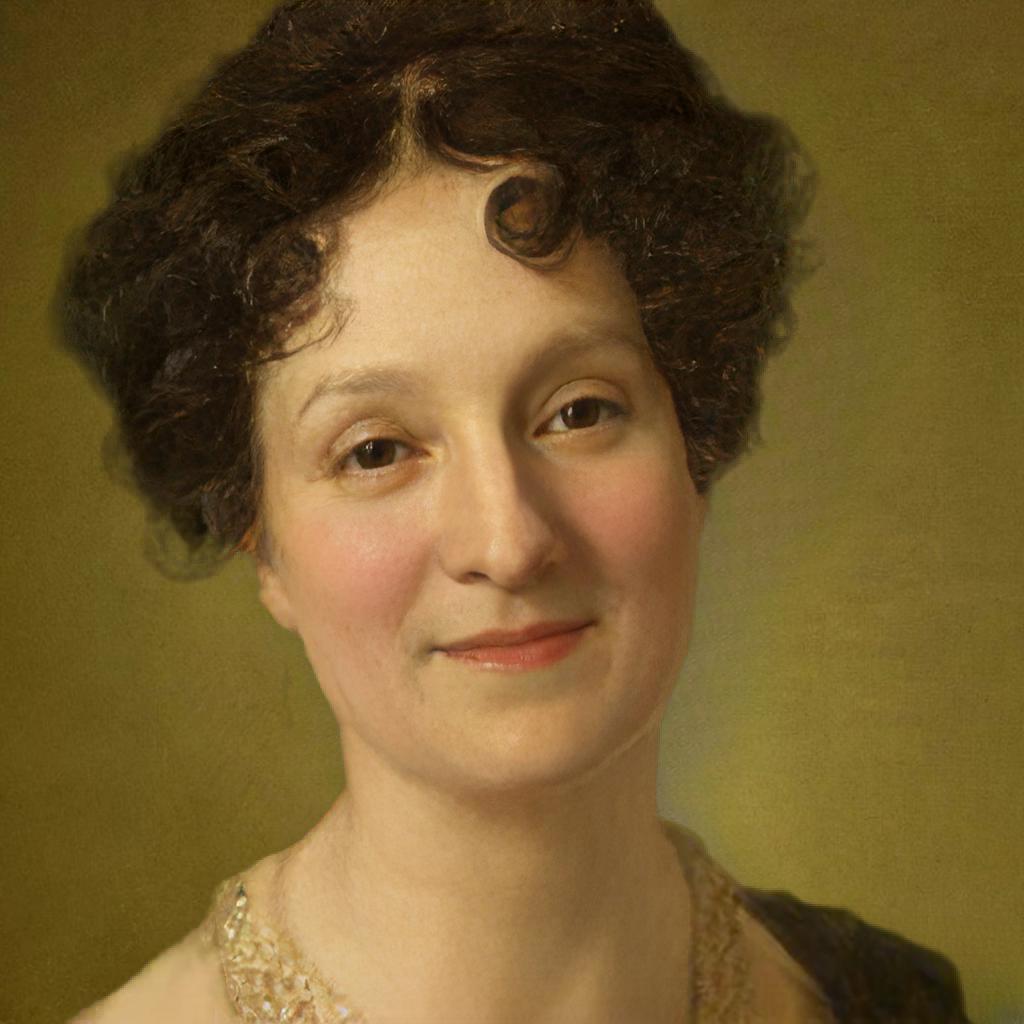} & 
        \hspace{\mrg}
        \includegraphics[width=\wid]{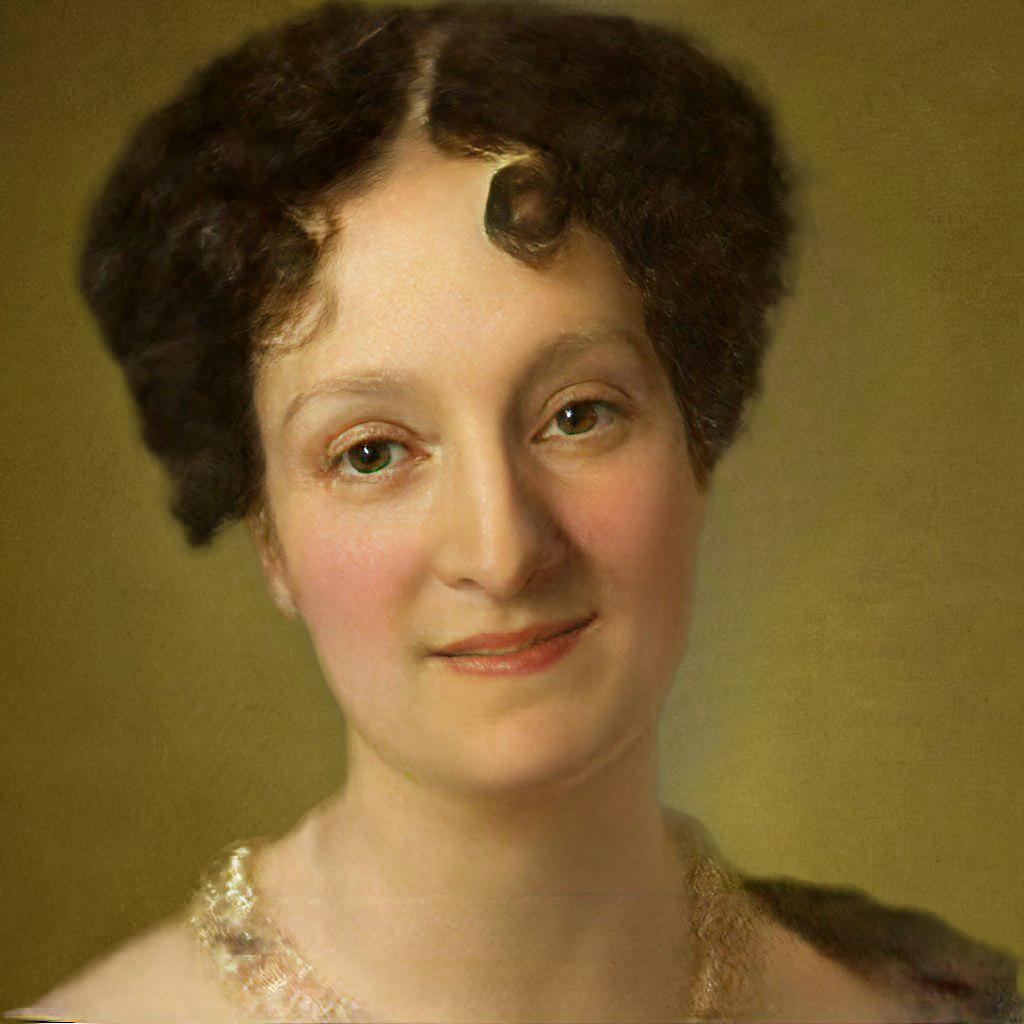}
        \\ %
        \includegraphics[width=\wid]{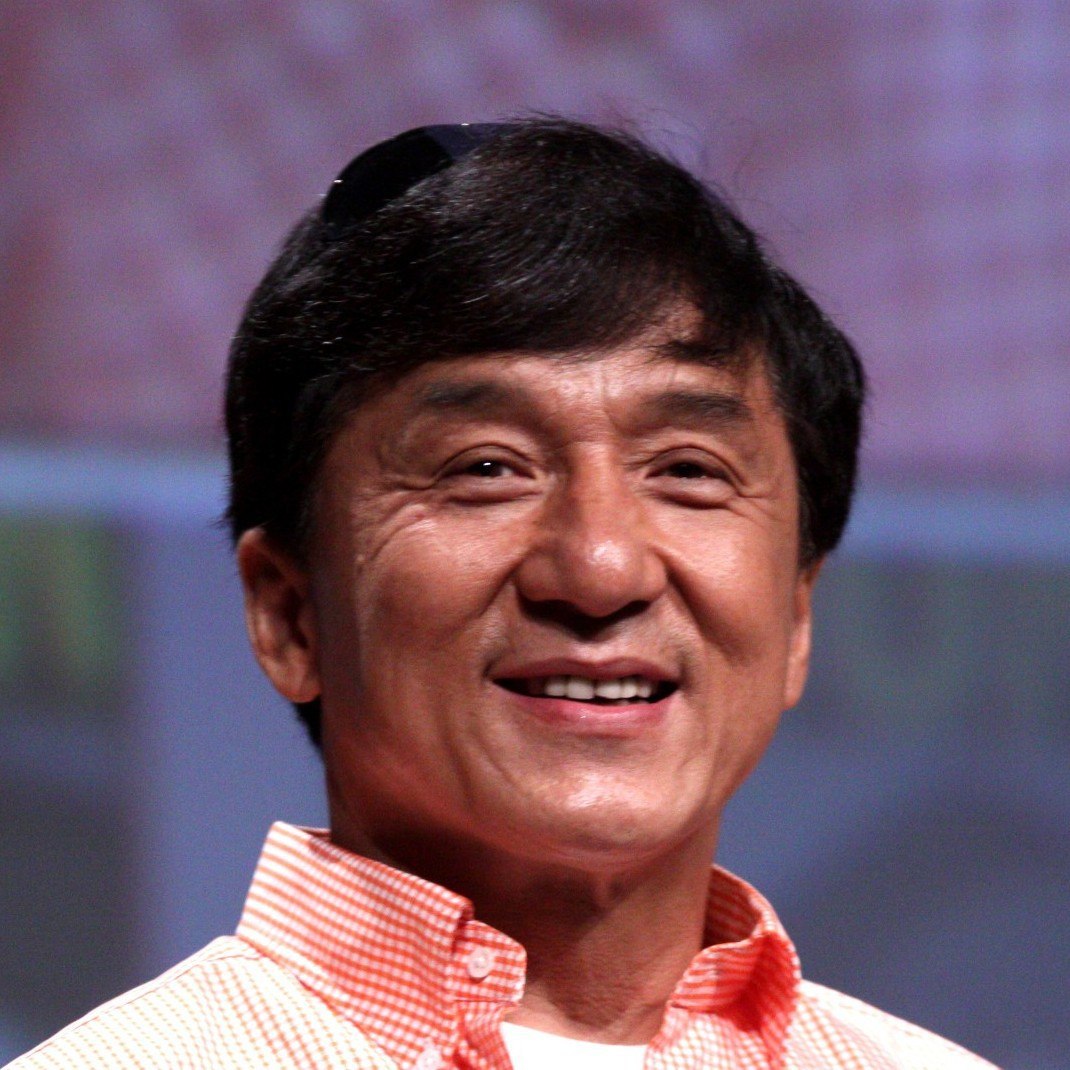} & 
        \hspace{\mrg}
        \includegraphics[width=\wid]{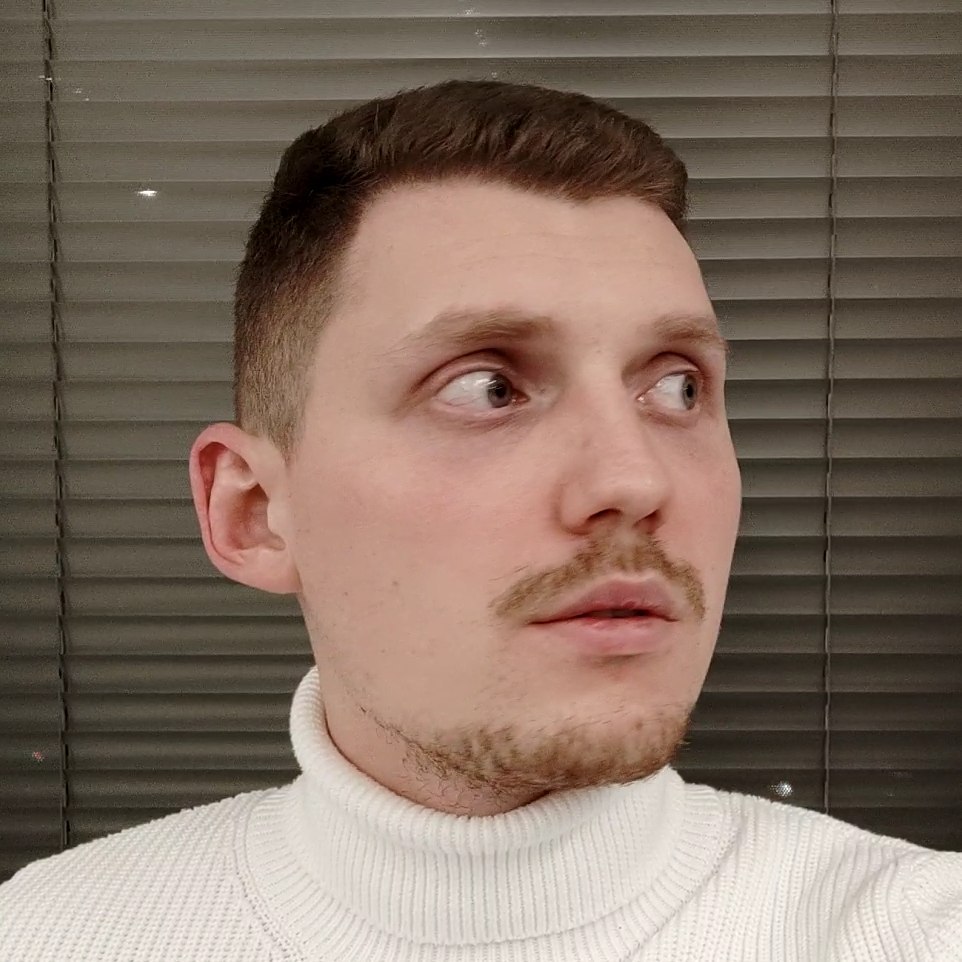} & 
        \hspace{\mrg}
        \includegraphics[width=\wid]{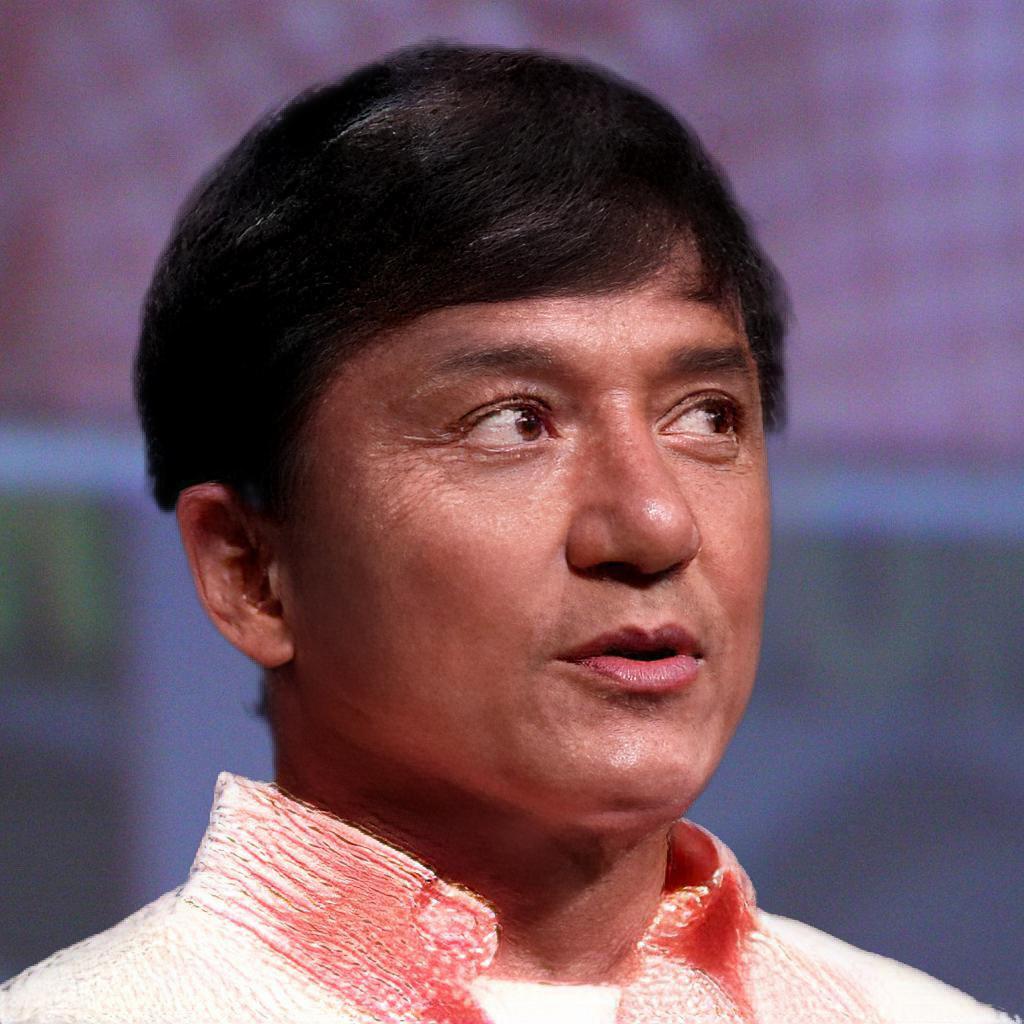} & 
        \hspace{\mrg}
        \includegraphics[width=\wid]{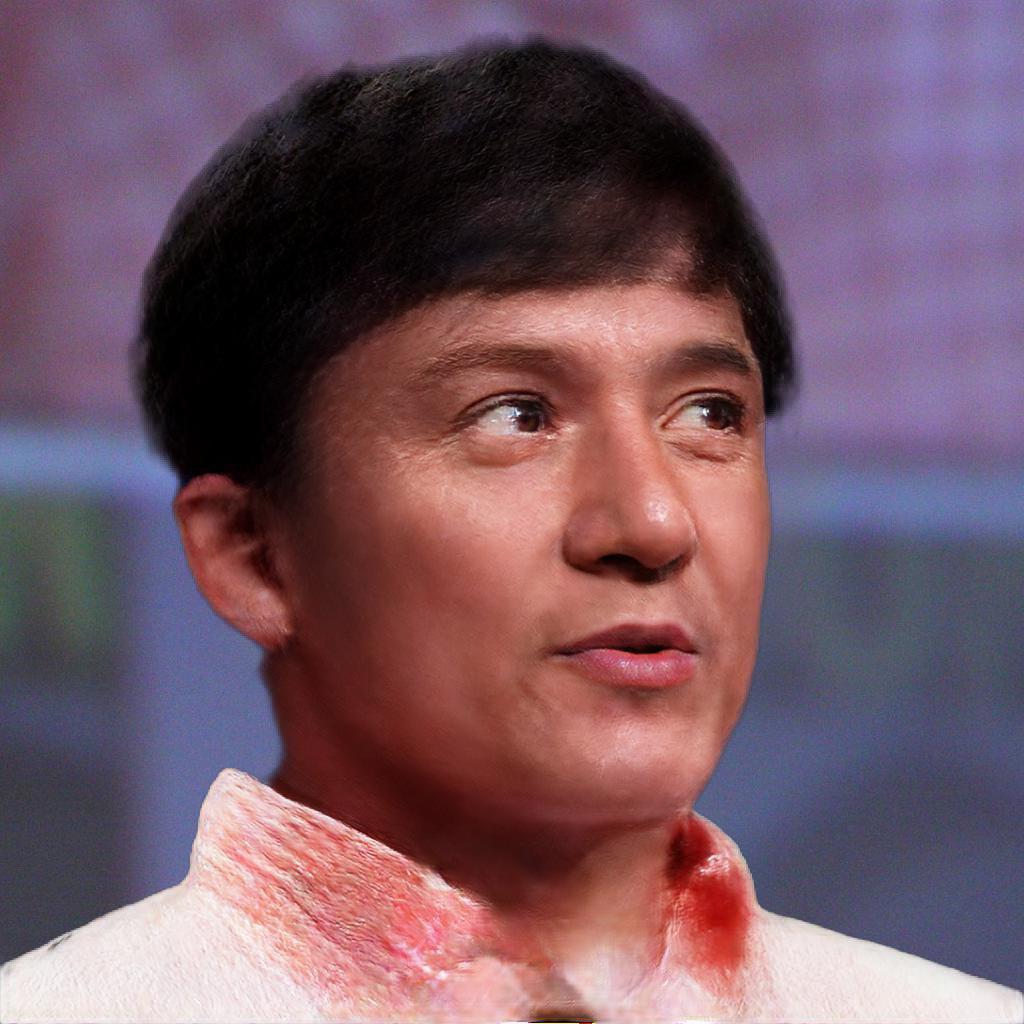}
        \\
        \textbf{Source} & 
        \hspace{\mrg} 
        \textbf{Driver} & 
        \hspace{\mrg} 
        \textbf{Ours (HR)} & 
        \hspace{\mrg} 
        \textbf{Ours (dist)} 
    \end{tabular}
    \vspace{-0.4cm}
    \caption{Results of the distilled version of our system trained for 100 avatars. It closely matches the prediction of the teacher model while being approximately ten times faster at the inference, achieving up to 130 FPS on a modern GPU.}
    \label{fig:distill}
\end{figure}

Finally, in \fig{distill} we show the results for the distillation of our base and high-resolution models into a small student network designed to work for a limited number of avatars. The architecture we chose for the distillation achieves 130 frames per second on the NVIDIA RTX 3090 graphics card in the FP16 mode. The total model size for the student containing 100 avatars is 800 megabytes. This model can closely match the performance of the teacher model. It thus achieves a PSNR of 23.14 and LPIPS of 0.208 (w.r.t. the teacher model) averaged across all avatars.

\begin{figure}
    \centering    
    \setlength{\wid}{0.09\textwidth}
    \setlength{\mrg}{-0.4cm}
    \begin{tabular}{ccccc}
        \includegraphics[width=\wid]{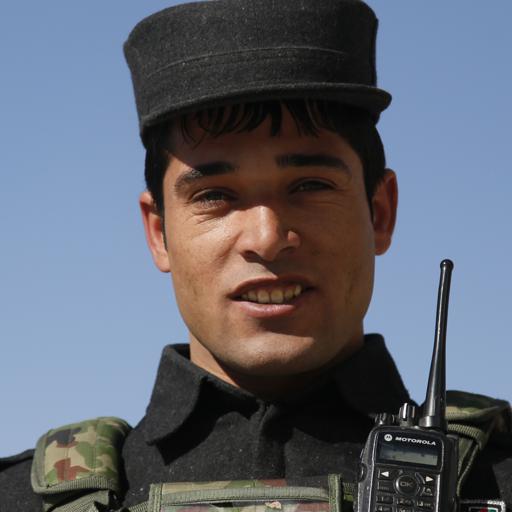} & 
        \hspace{\mrg}
        \includegraphics[width=\wid]{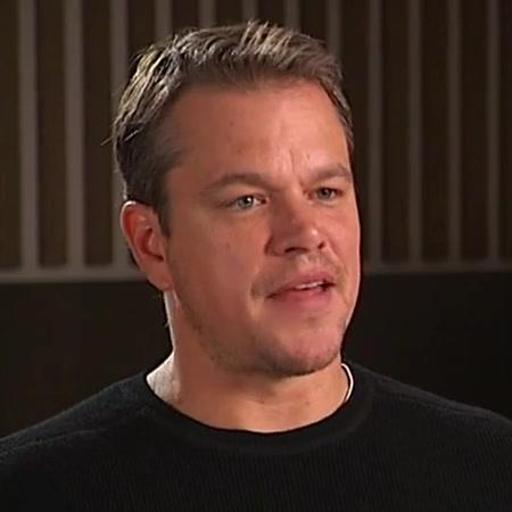} & 
        \hspace{\mrg}
        \includegraphics[width=\wid]{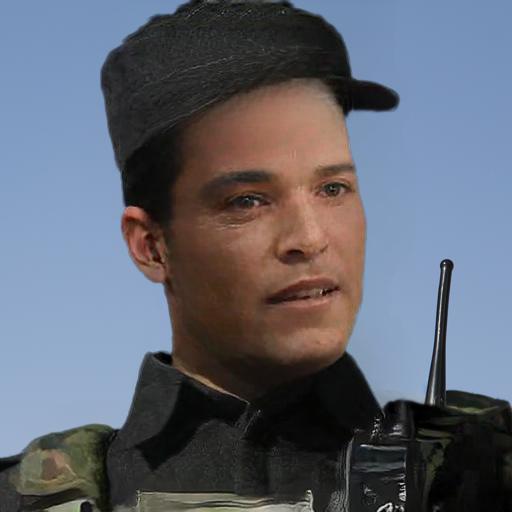} & 
        \hspace{\mrg}
        \includegraphics[width=\wid]{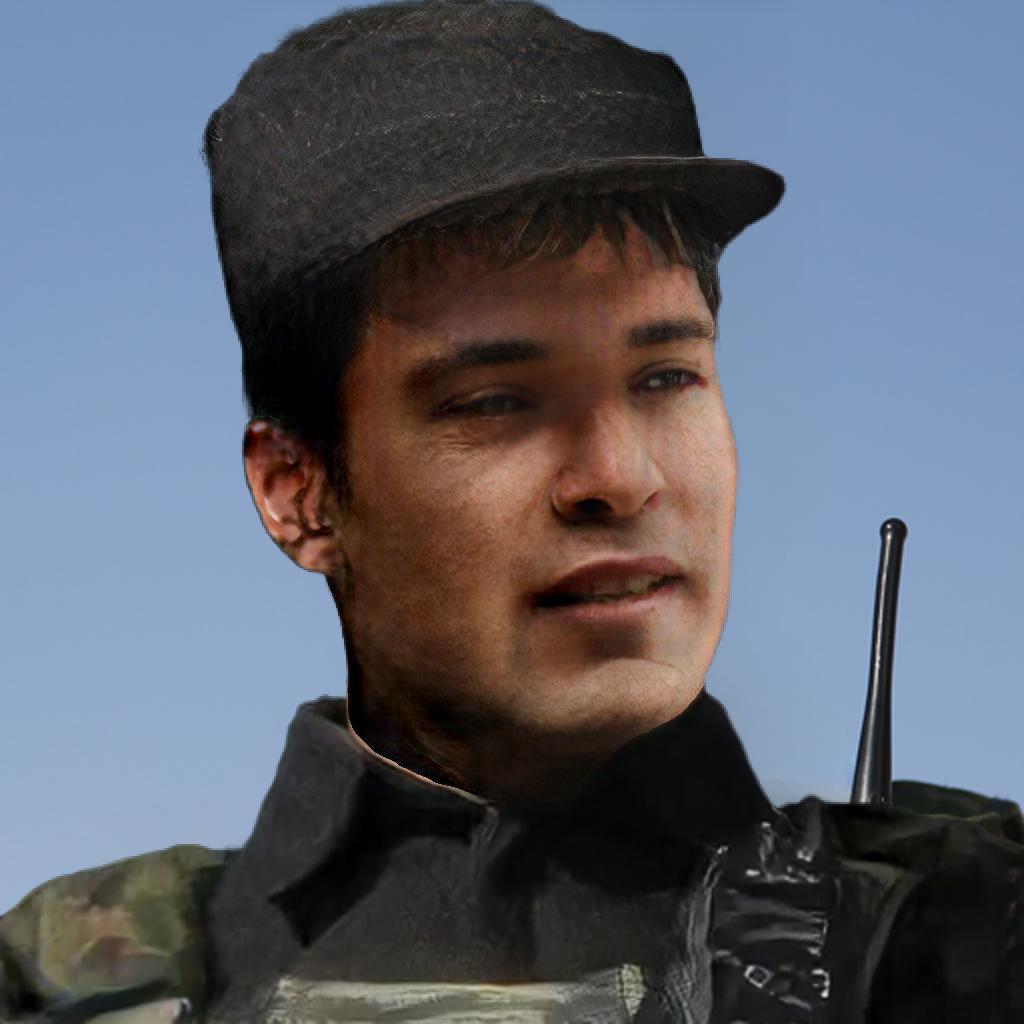} & 
        \hspace{\mrg}
        \includegraphics[width=\wid]{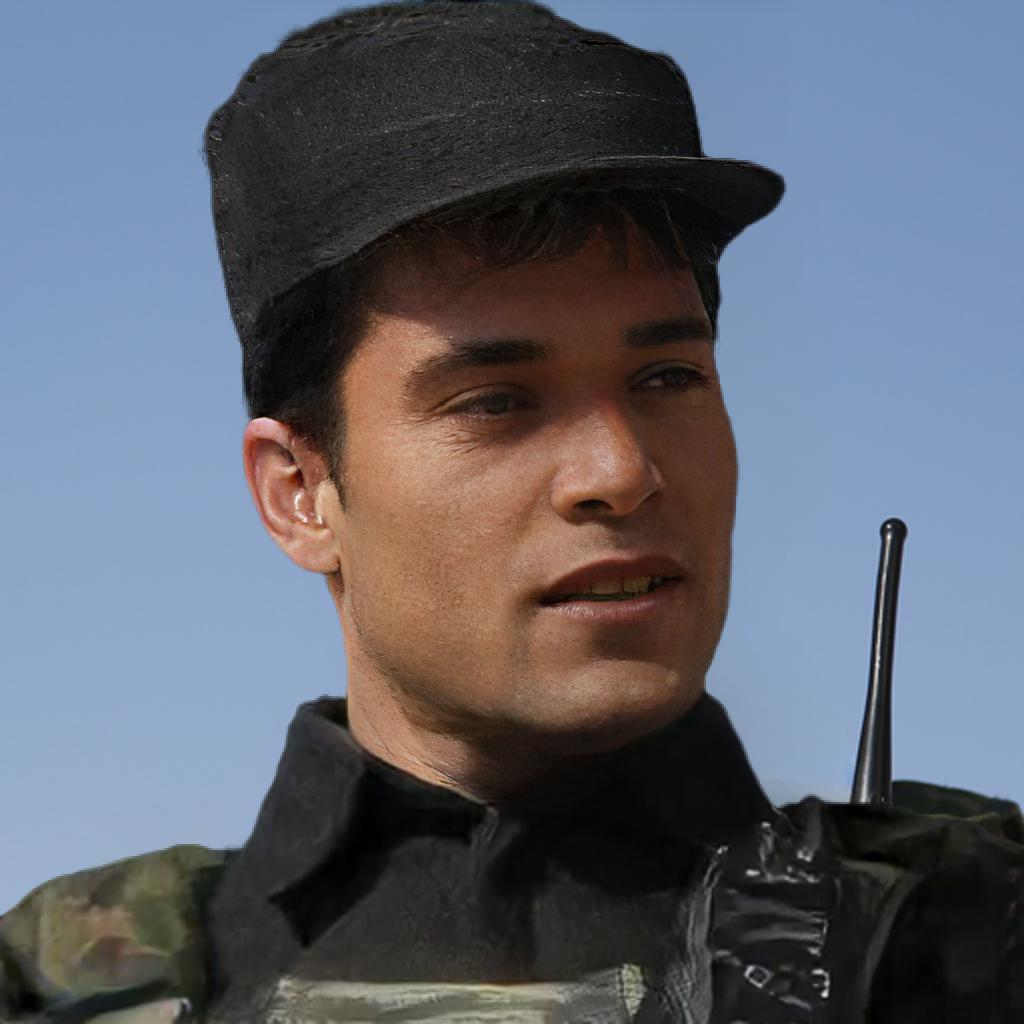}
        \\
        \textbf{Source} & 
        \hspace{\mrg} 
        \textbf{Driver} & 
        \hspace{\mrg} 
        \textbf{w/o $\mathcal{L}_\text{cos}$} & 
        \hspace{\mrg} 
        \textbf{w/o $\mathcal{L}^\text{c}_*$} & 
        \hspace{\mrg} 
        \textbf{Ours (HR)} 
    \end{tabular}
    \vspace{-0.4cm}
    \caption{Ablation study. Both contrastive loss $\mathcal{L}_\text{cos}$ and unsupervised super-resolution losses $\mathcal{L}_\text{adv}^\text{c}$ and $\mathcal{L}_\text{cyc}^\text{c}$ (denoted as $\mathcal{L}_*^\text{c}$) improve the performance of our method in the cross-driving scenario.}
    \label{fig:ablation}
    \vspace{-0.2cm}
\end{figure}
\begin{figure}
    \centering    
    \setlength{\wid}{0.09\textwidth}
    \setlength{\mrg}{-0.4cm}
    \begin{tabular}{ccccc}
        \includegraphics[width=\wid]{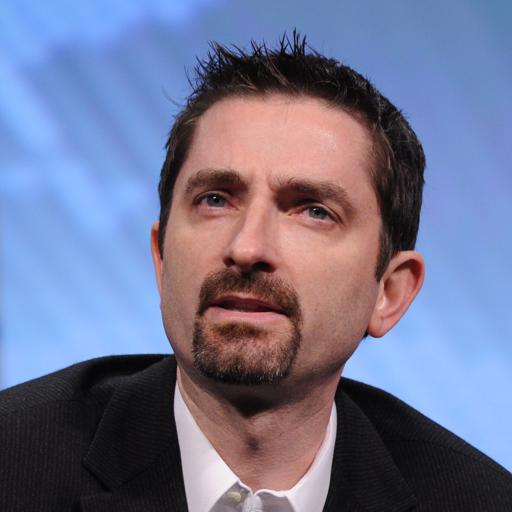} & 
        \hspace{\mrg}
        \includegraphics[width=\wid]{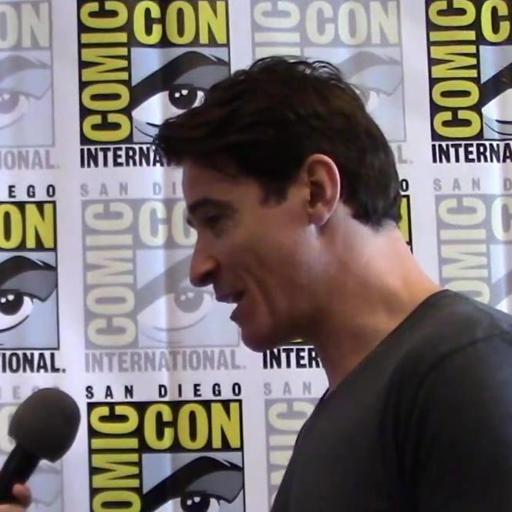} & 
        \hspace{\mrg}
        \includegraphics[width=\wid]{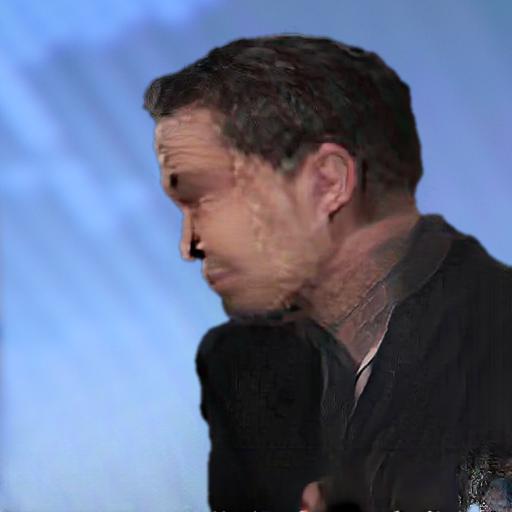} & 
        \hspace{\mrg}
        \includegraphics[width=\wid]{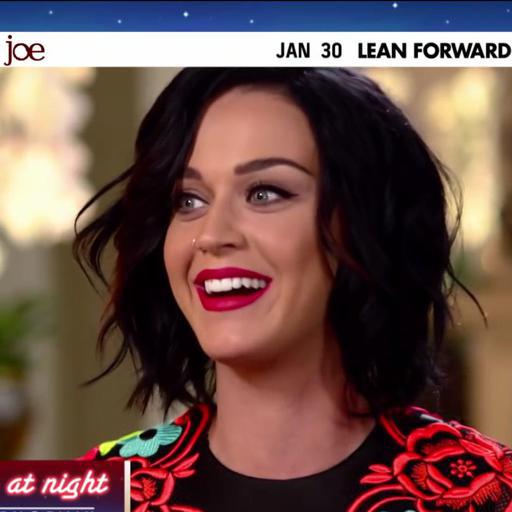} & 
        \hspace{\mrg}
        \includegraphics[width=\wid]{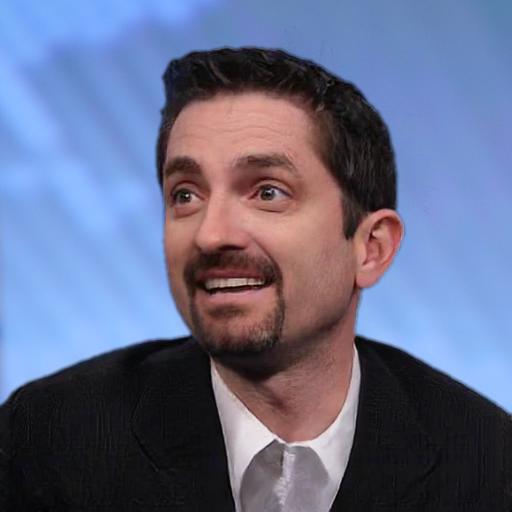}
        \\
        \textbf{Source} & 
        \hspace{\mrg} 
        \textbf{Driver 1} & 
        \hspace{\mrg} 
        \textbf{Ours 1} & 
        \hspace{\mrg} 
        \textbf{Driver 2} & 
        \hspace{\mrg} 
        \textbf{Ours 2} 
    \end{tabular}
    \vspace{-0.4cm}
    \caption{The limitations of our method include the inability to model large head rotations, which stems from the near frontal views distribution in the training data (1st example), and the lack of shoulders motion modeling (2nd example).}
    \label{fig:limitations}
\end{figure}

\subsection{Ablation study}

We conducted an extensive ablation study to evaluate the contributions of individual components within our method. Therefore, we evaluate the importance of the proposed cycle consistency losses for the base and high-resolution models. The qualitative results are shown in \fig{ablation}. Overall, both losses substantially improve the disentanglement between the motion and appearance. The quantitative evaluation confirms this: the base model without $\mathcal{L}_\text{cos}$ achieves an FID of 34.8, compared to the final 28.6, and the high-resolution model without cycle losses has an FID of 39.6, compared to the final FID of 39.2. We also provide an in-depth evaluation of the architectural choices in the supplementary materials.
\section{Conclusion}
\label{sect:conclusion}

We have presented a new approach for synthesizing high-resolution neural avatars. To the best of our knowledge, this approach is the first to achieve megapixel resolution. We have also explored a possible application of the proposed method in practice, which involves locking the identities of the avatars by training a dedicated student network. Using the student network also increases the rendering speed while achieving similar quality of renders to our full one-shot model.

Two main limitations of our system stem from the properties of our training set. First, both the VoxCeleb2 and the FFHQ datasets that we use for training tend to have near frontal views, which degrades the quality of rendering for strongly non-frontal head poses (\fig{limitations}). Secondly, as only static views are available at high resolution, a certain amount of temporal flicker is present in our results (see supplementary video). Ideally, this needs to be tackled with special losses or architectural choices. Lastly, our system lacks the modeling of shoulders motion. Addressing the issues mentioned above remains our future work.

\section*{Acknowledgements}

We sincerely thank Michail Christos Doukas for providing us inference results of HeadGAN \cite{Doukas2021HeadGANON} system and Ting-Chun Wang for providing us inference results of Face-V2V \cite{Wang2021OneShotFN} system. We also thank Roman Suvorov for their comments and suggestions regarding the text contents and clarity, as well as Julia Churkina for helping us with proof-reading. Furthermore, we thank Dmitry Nikulin for providing code for his gaze estimation model, on the top of which we built our gaze model. The computational resources for this work were mainly provided by Samsung ML Platform.

\bibliographystyle{ACM-Reference-Format}
\balance
\bibliography{refs}


\begin{thebibliography}{42}


\ifx \showCODEN    \undefined \def \showCODEN     #1{\unskip}     \fi
\ifx \showDOI      \undefined \def \showDOI       #1{#1}\fi
\ifx \showISBNx    \undefined \def \showISBNx     #1{\unskip}     \fi
\ifx \showISBNxiii \undefined \def \showISBNxiii  #1{\unskip}     \fi
\ifx \showISSN     \undefined \def \showISSN      #1{\unskip}     \fi
\ifx \showLCCN     \undefined \def \showLCCN      #1{\unskip}     \fi
\ifx \shownote     \undefined \def \shownote      #1{#1}          \fi
\ifx \showarticletitle \undefined \def \showarticletitle #1{#1}   \fi
\ifx \showURL      \undefined \def \showURL       {\relax}        \fi
\providecommand\bibfield[2]{#2}
\providecommand\bibinfo[2]{#2}
\providecommand\natexlab[1]{#1}
\providecommand\showeprint[2][]{arXiv:#2}

\bibitem[\protect\citeauthoryear{Blanz and Vetter}{Blanz and Vetter}{1999}]%
        {Blanz1999AMM}
\bibfield{author}{\bibinfo{person}{Volker Blanz} {and} \bibinfo{person}{Thomas
  Vetter}.} \bibinfo{year}{1999}\natexlab{}.
\newblock \showarticletitle{A morphable model for the synthesis of 3D faces}.
  In \bibinfo{booktitle}{\emph{SIGGRAPH '99}}.
\newblock


\bibitem[\protect\citeauthoryear{Bulat and Tzimiropoulos}{Bulat and
  Tzimiropoulos}{2017}]%
        {Bulat2017HowFA}
\bibfield{author}{\bibinfo{person}{Adrian Bulat} {and}
  \bibinfo{person}{Georgios Tzimiropoulos}.} \bibinfo{year}{2017}\natexlab{}.
\newblock \showarticletitle{How Far are We from Solving the 2D \& 3D Face
  Alignment Problem? (and a Dataset of 230,000 3D Facial Landmarks)}.
\newblock \bibinfo{journal}{\emph{2017 IEEE International Conference on
  Computer Vision (ICCV)}} (\bibinfo{year}{2017}), \bibinfo{pages}{1021--1030}.
\newblock


\bibitem[\protect\citeauthoryear{Burkov, Pasechnik, Grigorev, and
  Lempitsky}{Burkov et~al\mbox{.}}{2020}]%
        {Burkov_2020_CVPR}
\bibfield{author}{\bibinfo{person}{Egor Burkov}, \bibinfo{person}{I.
  Pasechnik}, \bibinfo{person}{Artur Grigorev}, {and}
  \bibinfo{person}{Victor~S. Lempitsky}.} \bibinfo{year}{2020}\natexlab{}.
\newblock \showarticletitle{Neural Head Reenactment with Latent Pose
  Descriptors}.
\newblock \bibinfo{journal}{\emph{2020 IEEE/CVF Conference on Computer Vision
  and Pattern Recognition (CVPR)}} (\bibinfo{year}{2020}),
  \bibinfo{pages}{13783--13792}.
\newblock


\bibitem[\protect\citeauthoryear{Chung, Nagrani, and Zisserman}{Chung
  et~al\mbox{.}}{2018}]%
        {Chung2018VoxCeleb2DS}
\bibfield{author}{\bibinfo{person}{Joon~Son Chung}, \bibinfo{person}{Arsha
  Nagrani}, {and} \bibinfo{person}{Andrew Zisserman}.}
  \bibinfo{year}{2018}\natexlab{}.
\newblock \showarticletitle{VoxCeleb2: Deep Speaker Recognition}. In
  \bibinfo{booktitle}{\emph{INTERSPEECH}}.
\newblock


\bibitem[\protect\citeauthoryear{Cortacero, Fischer, and Demiris}{Cortacero
  et~al\mbox{.}}{2019}]%
        {Cortacero_2019_ICCV}
\bibfield{author}{\bibinfo{person}{Kevin Cortacero}, \bibinfo{person}{Tobias
  Fischer}, {and} \bibinfo{person}{Yiannis Demiris}.}
  \bibinfo{year}{2019}\natexlab{}.
\newblock \showarticletitle{RT-BENE: A Dataset and Baselines for Real-Time
  Blink Estimation in Natural Environments}. In
  \bibinfo{booktitle}{\emph{Proceedings of the IEEE/CVF International
  Conference on Computer Vision (ICCV) Workshops}}.
\newblock


\bibitem[\protect\citeauthoryear{Deng, Dong, Socher, Li, Li, and Fei-Fei}{Deng
  et~al\mbox{.}}{2009}]%
        {Deng2009ImageNetAL}
\bibfield{author}{\bibinfo{person}{Jia Deng}, \bibinfo{person}{Wei Dong},
  \bibinfo{person}{Richard Socher}, \bibinfo{person}{Li-Jia Li},
  \bibinfo{person}{K. Li}, {and} \bibinfo{person}{Li Fei-Fei}.}
  \bibinfo{year}{2009}\natexlab{}.
\newblock \showarticletitle{ImageNet: A large-scale hierarchical image
  database}. In \bibinfo{booktitle}{\emph{CVPR}}.
\newblock


\bibitem[\protect\citeauthoryear{Deng, Guo, Ververas, Kotsia, Zafeiriou, and
  FaceSoft}{Deng et~al\mbox{.}}{2020}]%
        {Deng2020RetinaFaceSM}
\bibfield{author}{\bibinfo{person}{Jiankang Deng}, \bibinfo{person}{J. Guo},
  \bibinfo{person}{Evangelos Ververas}, \bibinfo{person}{Irene Kotsia},
  \bibinfo{person}{Stefanos Zafeiriou}, {and} \bibinfo{person}{InsightFace
  FaceSoft}.} \bibinfo{year}{2020}\natexlab{}.
\newblock \showarticletitle{RetinaFace: Single-Shot Multi-Level Face
  Localisation in the Wild}.
\newblock \bibinfo{journal}{\emph{2020 IEEE/CVF Conference on Computer Vision
  and Pattern Recognition (CVPR)}} (\bibinfo{year}{2020}),
  \bibinfo{pages}{5202--5211}.
\newblock


\bibitem[\protect\citeauthoryear{Doukas, Zafeiriou, and Sharmanska}{Doukas
  et~al\mbox{.}}{2021}]%
        {Doukas2021HeadGANON}
\bibfield{author}{\bibinfo{person}{Michail~Christos Doukas},
  \bibinfo{person}{Stefanos Zafeiriou}, {and} \bibinfo{person}{Viktoriia
  Sharmanska}.} \bibinfo{year}{2021}\natexlab{}.
\newblock \showarticletitle{HeadGAN: One-shot Neural Head Synthesis and
  Editing}.
\newblock \bibinfo{journal}{\emph{2021 IEEE/CVF International Conference on
  Computer Vision (ICCV)}}.
\newblock


\bibitem[\protect\citeauthoryear{Fischer, Chang, and Demiris}{Fischer
  et~al\mbox{.}}{2018}]%
        {Fischer2018RTGENERE}
\bibfield{author}{\bibinfo{person}{Tobias Fischer}, \bibinfo{person}{Hyung~Jin
  Chang}, {and} \bibinfo{person}{Y. Demiris}.} \bibinfo{year}{2018}\natexlab{}.
\newblock \showarticletitle{RT-GENE: Real-Time Eye Gaze Estimation in Natural
  Environments}. In \bibinfo{booktitle}{\emph{ECCV}}.
\newblock


\bibitem[\protect\citeauthoryear{Gafni, Thies, Zollhofer, and
  Nie{\ss}ner}{Gafni et~al\mbox{.}}{2021}]%
        {Gafni2021DynamicNR}
\bibfield{author}{\bibinfo{person}{Guy Gafni}, \bibinfo{person}{Justus Thies},
  \bibinfo{person}{Michael Zollhofer}, {and} \bibinfo{person}{Matthias
  Nie{\ss}ner}.} \bibinfo{year}{2021}\natexlab{}.
\newblock \showarticletitle{Dynamic Neural Radiance Fields for Monocular 4D
  Facial Avatar Reconstruction}.
\newblock \bibinfo{journal}{\emph{2021 IEEE/CVF Conference on Computer Vision
  and Pattern Recognition (CVPR)}}, \bibinfo{pages}{8645--8654}.
\newblock


\bibitem[\protect\citeauthoryear{Gong, Gao, Liang, Shen, Wang, and Lin}{Gong
  et~al\mbox{.}}{2019}]%
        {Gong2019GraphonomyUH}
\bibfield{author}{\bibinfo{person}{Ke Gong}, \bibinfo{person}{Yiming Gao},
  \bibinfo{person}{Xiaodan Liang}, \bibinfo{person}{Xiaohui Shen},
  \bibinfo{person}{M. Wang}, {and} \bibinfo{person}{Liang Lin}.}
  \bibinfo{year}{2019}\natexlab{}.
\newblock \showarticletitle{Graphonomy: Universal Human Parsing via Graph
  Transfer Learning}.
\newblock \bibinfo{journal}{\emph{2019 IEEE/CVF Conference on Computer Vision
  and Pattern Recognition (CVPR)}} (\bibinfo{year}{2019}),
  \bibinfo{pages}{7442--7451}.
\newblock


\bibitem[\protect\citeauthoryear{Ha, Kersner, Kim, Seo, and Kim}{Ha
  et~al\mbox{.}}{2020}]%
        {Ha2020MarioNETteFF}
\bibfield{author}{\bibinfo{person}{Sungjoo Ha}, \bibinfo{person}{Martin
  Kersner}, \bibinfo{person}{Beomsu Kim}, \bibinfo{person}{Seokjun Seo}, {and}
  \bibinfo{person}{Dongyoung Kim}.} \bibinfo{year}{2020}\natexlab{}.
\newblock \showarticletitle{MarioNETte: Few-shot Face Reenactment Preserving
  Identity of Unseen Targets}. In \bibinfo{booktitle}{\emph{AAAI}}.
\newblock


\bibitem[\protect\citeauthoryear{Heusel, Ramsauer, Unterthiner, Nessler, and
  Hochreiter}{Heusel et~al\mbox{.}}{2017}]%
        {fid}
\bibfield{author}{\bibinfo{person}{Martin Heusel}, \bibinfo{person}{Hubert
  Ramsauer}, \bibinfo{person}{Thomas Unterthiner}, \bibinfo{person}{Bernhard
  Nessler}, {and} \bibinfo{person}{Sepp Hochreiter}.}
  \bibinfo{year}{2017}\natexlab{}.
\newblock \showarticletitle{GANs Trained by a Two Time-Scale Update Rule
  Converge to a Local Nash Equilibrium}. In \bibinfo{booktitle}{\emph{Advances
  in Neural Information Processing Systems}}.
\newblock


\bibitem[\protect\citeauthoryear{Johnson, Alahi, and Fei-Fei}{Johnson
  et~al\mbox{.}}{2016}]%
        {Johnson2016PerceptualLF}
\bibfield{author}{\bibinfo{person}{Justin Johnson}, \bibinfo{person}{Alexandre
  Alahi}, {and} \bibinfo{person}{Li Fei-Fei}.} \bibinfo{year}{2016}\natexlab{}.
\newblock \showarticletitle{Perceptual Losses for Real-Time Style Transfer and
  Super-Resolution}. In \bibinfo{booktitle}{\emph{ECCV}}.
\newblock


\bibitem[\protect\citeauthoryear{Karras, Laine, and Aila}{Karras
  et~al\mbox{.}}{2019}]%
        {Karras2019ASG}
\bibfield{author}{\bibinfo{person}{Tero Karras}, \bibinfo{person}{Samuli
  Laine}, {and} \bibinfo{person}{Timo Aila}.} \bibinfo{year}{2019}\natexlab{}.
\newblock \showarticletitle{A Style-Based Generator Architecture for Generative
  Adversarial Networks}.
\newblock \bibinfo{journal}{\emph{2019 IEEE/CVF Conference on Computer Vision
  and Pattern Recognition (CVPR)}}, \bibinfo{pages}{4396--4405}.
\newblock


\bibitem[\protect\citeauthoryear{Ke, Sun, Li, Yan, and Lau}{Ke
  et~al\mbox{.}}{2022}]%
        {MODNet}
\bibfield{author}{\bibinfo{person}{Zhanghan Ke}, \bibinfo{person}{Jiayu Sun},
  \bibinfo{person}{Kaican Li}, \bibinfo{person}{Qiong Yan}, {and}
  \bibinfo{person}{Rynson~W.H. Lau}.} \bibinfo{year}{2022}\natexlab{}.
\newblock \showarticletitle{MODNet: Real-Time Trimap-Free Portrait Matting via
  Objective Decomposition}. In \bibinfo{booktitle}{\emph{AAAI}}.
\newblock


\bibitem[\protect\citeauthoryear{Kim, Garrido, Tewari, Xu, Thies, Nie{\ss}ner,
  P{\'e}rez, Richardt, Zollh{\"o}fer, and Theobalt}{Kim et~al\mbox{.}}{2018}]%
        {Kim2018DeepVP}
\bibfield{author}{\bibinfo{person}{Hyeongwoo Kim}, \bibinfo{person}{Pablo
  Garrido}, \bibinfo{person}{Ayush Tewari}, \bibinfo{person}{Weipeng Xu},
  \bibinfo{person}{Justus Thies}, \bibinfo{person}{Matthias Nie{\ss}ner},
  \bibinfo{person}{Patrick P{\'e}rez}, \bibinfo{person}{Christian Richardt},
  \bibinfo{person}{Michael Zollh{\"o}fer}, {and} \bibinfo{person}{Christian
  Theobalt}.} \bibinfo{year}{2018}\natexlab{}.
\newblock \showarticletitle{Deep video portraits}.
\newblock \bibinfo{journal}{\emph{ACM Transactions on Graphics (TOG)}}
  \bibinfo{volume}{37} (\bibinfo{year}{2018}), \bibinfo{pages}{1 -- 14}.
\newblock


\bibitem[\protect\citeauthoryear{Lombardi, Saragih, Simon, and Sheikh}{Lombardi
  et~al\mbox{.}}{2018}]%
        {Lombardi2018DeepAM}
\bibfield{author}{\bibinfo{person}{Stephen Lombardi}, \bibinfo{person}{Jason~M.
  Saragih}, \bibinfo{person}{Tomas Simon}, {and} \bibinfo{person}{Yaser
  Sheikh}.} \bibinfo{year}{2018}\natexlab{}.
\newblock \showarticletitle{Deep appearance models for face rendering}.
\newblock \bibinfo{journal}{\emph{ACM Transactions on Graphics (TOG)}}
  \bibinfo{volume}{37} (\bibinfo{year}{2018}), \bibinfo{pages}{1 -- 13}.
\newblock


\bibitem[\protect\citeauthoryear{Lombardi, Simon, Saragih, Schwartz, Lehrmann,
  and Sheikh}{Lombardi et~al\mbox{.}}{2019}]%
        {Lombardi2019NeuralV}
\bibfield{author}{\bibinfo{person}{Stephen Lombardi}, \bibinfo{person}{Tomas
  Simon}, \bibinfo{person}{Jason~M. Saragih}, \bibinfo{person}{Gabriel
  Schwartz}, \bibinfo{person}{Andreas~M. Lehrmann}, {and}
  \bibinfo{person}{Yaser Sheikh}.} \bibinfo{year}{2019}\natexlab{}.
\newblock \showarticletitle{Neural volumes}.
\newblock \bibinfo{journal}{\emph{ACM Transactions on Graphics (TOG)}}
  \bibinfo{volume}{38} (\bibinfo{year}{2019}), \bibinfo{pages}{1 -- 14}.
\newblock


\bibitem[\protect\citeauthoryear{Loshchilov and Hutter}{Loshchilov and
  Hutter}{2019}]%
        {Loshchilov2019DecoupledWD}
\bibfield{author}{\bibinfo{person}{Ilya Loshchilov} {and}
  \bibinfo{person}{Frank Hutter}.} \bibinfo{year}{2019}\natexlab{}.
\newblock \showarticletitle{Decoupled Weight Decay Regularization}. In
  \bibinfo{booktitle}{\emph{ICLR}}.
\newblock


\bibitem[\protect\citeauthoryear{Mildenhall, Srinivasan, Tancik, Barron,
  Ramamoorthi, and Ng}{Mildenhall et~al\mbox{.}}{2020}]%
        {Mildenhall2020NeRFRS}
\bibfield{author}{\bibinfo{person}{Ben Mildenhall}, \bibinfo{person}{Pratul~P.
  Srinivasan}, \bibinfo{person}{Matthew Tancik}, \bibinfo{person}{Jonathan~T.
  Barron}, \bibinfo{person}{Ravi Ramamoorthi}, {and} \bibinfo{person}{Ren Ng}.}
  \bibinfo{year}{2020}\natexlab{}.
\newblock \showarticletitle{NeRF: Representing Scenes as Neural Radiance Fields
  for View Synthesis}. In \bibinfo{booktitle}{\emph{ECCV}}.
\newblock


\bibitem[\protect\citeauthoryear{Park, Sinha, Barron, Bouaziz, Goldman, Seitz,
  and Martin-Brualla}{Park et~al\mbox{.}}{2021a}]%
        {Park2021NerfiesDN}
\bibfield{author}{\bibinfo{person}{Keunhong Park}, \bibinfo{person}{U. Sinha},
  \bibinfo{person}{Jonathan~T. Barron}, \bibinfo{person}{Sofien Bouaziz},
  \bibinfo{person}{Dan~B. Goldman}, \bibinfo{person}{Steven~M. Seitz}, {and}
  \bibinfo{person}{Ricardo Martin-Brualla}.} \bibinfo{year}{2021}\natexlab{a}.
\newblock \showarticletitle{Nerfies: Deformable Neural Radiance Fields}.
\newblock \bibinfo{journal}{\emph{2021 IEEE/CVF International Conference on
  Computer Vision (ICCV)}}.
\newblock


\bibitem[\protect\citeauthoryear{Park, Sinha, Hedman, Barron, Bouaziz, Goldman,
  Martin-Brualla, and Seitz}{Park et~al\mbox{.}}{2021b}]%
        {Park2021HyperNeRFAH}
\bibfield{author}{\bibinfo{person}{Keunhong Park}, \bibinfo{person}{U. Sinha},
  \bibinfo{person}{Peter Hedman}, \bibinfo{person}{Jonathan~T. Barron},
  \bibinfo{person}{Sofien Bouaziz}, \bibinfo{person}{Dan~B. Goldman},
  \bibinfo{person}{Ricardo Martin-Brualla}, {and} \bibinfo{person}{Steven~M.
  Seitz}.} \bibinfo{year}{2021}\natexlab{b}.
\newblock \showarticletitle{HyperNeRF: A Higher-Dimensional Representation for
  Topologically Varying Neural Radiance Fields}.
\newblock \bibinfo{journal}{\emph{ArXiv}}.
\newblock


\bibitem[\protect\citeauthoryear{Parkhi, Vedaldi, and Zisserman}{Parkhi
  et~al\mbox{.}}{2015}]%
        {Parkhi2015DeepFR}
\bibfield{author}{\bibinfo{person}{Omkar~M. Parkhi}, \bibinfo{person}{Andrea
  Vedaldi}, {and} \bibinfo{person}{Andrew Zisserman}.}
  \bibinfo{year}{2015}\natexlab{}.
\newblock \showarticletitle{Deep Face Recognition}. In
  \bibinfo{booktitle}{\emph{BMVC}}.
\newblock


\bibitem[\protect\citeauthoryear{Siarohin, Lathuili{\`e}re, Tulyakov, Ricci,
  and Sebe}{Siarohin et~al\mbox{.}}{2019a}]%
        {Siarohin2019AnimatingAO}
\bibfield{author}{\bibinfo{person}{Aliaksandr Siarohin},
  \bibinfo{person}{St{\'e}phane Lathuili{\`e}re}, \bibinfo{person}{S.
  Tulyakov}, \bibinfo{person}{Elisa Ricci}, {and} \bibinfo{person}{N. Sebe}.}
  \bibinfo{year}{2019}\natexlab{a}.
\newblock \showarticletitle{Animating Arbitrary Objects via Deep Motion
  Transfer}.
\newblock \bibinfo{journal}{\emph{2019 IEEE/CVF Conference on Computer Vision
  and Pattern Recognition (CVPR)}} (\bibinfo{year}{2019}),
  \bibinfo{pages}{2372--2381}.
\newblock


\bibitem[\protect\citeauthoryear{Siarohin, Lathuili{\`e}re, Tulyakov, Ricci,
  and Sebe}{Siarohin et~al\mbox{.}}{2019b}]%
        {Siarohin2019FirstOM}
\bibfield{author}{\bibinfo{person}{Aliaksandr Siarohin},
  \bibinfo{person}{St{\'e}phane Lathuili{\`e}re}, \bibinfo{person}{S.
  Tulyakov}, \bibinfo{person}{Elisa Ricci}, {and} \bibinfo{person}{N. Sebe}.}
  \bibinfo{year}{2019}\natexlab{b}.
\newblock \showarticletitle{First Order Motion Model for Image Animation}.
\newblock \bibinfo{journal}{\emph{ArXiv}}  \bibinfo{volume}{abs/2003.00196}
  (\bibinfo{year}{2019}).
\newblock


\bibitem[\protect\citeauthoryear{Simonyan and Zisserman}{Simonyan and
  Zisserman}{2015}]%
        {Simonyan2015VeryDC}
\bibfield{author}{\bibinfo{person}{Karen Simonyan} {and}
  \bibinfo{person}{Andrew Zisserman}.} \bibinfo{year}{2015}\natexlab{}.
\newblock \showarticletitle{Very Deep Convolutional Networks for Large-Scale
  Image Recognition}.
\newblock \bibinfo{journal}{\emph{CoRR}}  \bibinfo{volume}{abs/1409.1556}
  (\bibinfo{year}{2015}).
\newblock


\bibitem[\protect\citeauthoryear{Su, Yan, Zhu, Zhang, Ge, Sun, and Zhang}{Su
  et~al\mbox{.}}{2020}]%
        {hyperIQA}
\bibfield{author}{\bibinfo{person}{Shaolin Su}, \bibinfo{person}{Qingsen Yan},
  \bibinfo{person}{Yu Zhu}, \bibinfo{person}{Cheng Zhang}, \bibinfo{person}{Xin
  Ge}, \bibinfo{person}{Jinqiu Sun}, {and} \bibinfo{person}{Yanning Zhang}.}
  \bibinfo{year}{2020}\natexlab{}.
\newblock \showarticletitle{Blindly Assess Image Quality in the Wild Guided by
  a Self-Adaptive Hyper Network}. In \bibinfo{booktitle}{\emph{IEEE/CVF
  Conference on Computer Vision and Pattern Recognition (CVPR)}}.
\newblock


\bibitem[\protect\citeauthoryear{Suvorov, Logacheva, Mashikhin, Remizova,
  Ashukha, Silvestrov, Kong, Goka, Park, and Lempitsky}{Suvorov
  et~al\mbox{.}}{2021}]%
        {suvorov2021resolution}
\bibfield{author}{\bibinfo{person}{Roman Suvorov}, \bibinfo{person}{Elizaveta
  Logacheva}, \bibinfo{person}{Anton Mashikhin}, \bibinfo{person}{Anastasia
  Remizova}, \bibinfo{person}{Arsenii Ashukha}, \bibinfo{person}{Aleksei
  Silvestrov}, \bibinfo{person}{Naejin Kong}, \bibinfo{person}{Harshith Goka},
  \bibinfo{person}{Kiwoong Park}, {and} \bibinfo{person}{Victor Lempitsky}.}
  \bibinfo{year}{2021}\natexlab{}.
\newblock \showarticletitle{Resolution-robust Large Mask Inpainting with
  Fourier Convolutions}.
\newblock \bibinfo{journal}{\emph{arXiv preprint arXiv:2109.07161}}
  (\bibinfo{year}{2021}).
\newblock


\bibitem[\protect\citeauthoryear{Thies, Zollh{\"o}fer, Stamminger, Theobalt,
  and Nie{\ss}ner}{Thies et~al\mbox{.}}{2019}]%
        {Thies2019Face2FaceRF}
\bibfield{author}{\bibinfo{person}{Justus Thies}, \bibinfo{person}{Michael
  Zollh{\"o}fer}, \bibinfo{person}{Marc Stamminger}, \bibinfo{person}{Christian
  Theobalt}, {and} \bibinfo{person}{Matthias Nie{\ss}ner}.}
  \bibinfo{year}{2019}\natexlab{}.
\newblock \showarticletitle{Face2Face: real-time face capture and reenactment
  of RGB videos}.
\newblock \bibinfo{journal}{\emph{ArXiv}}  \bibinfo{volume}{abs/2007.14808}
  (\bibinfo{year}{2019}).
\newblock


\bibitem[\protect\citeauthoryear{Wang, Wang, Zhou, Ji, Li, Gong, Zhou, and
  Liu}{Wang et~al\mbox{.}}{2018c}]%
        {Wang2018CosFaceLM}
\bibfield{author}{\bibinfo{person}{H. Wang}, \bibinfo{person}{Yitong Wang},
  \bibinfo{person}{Zheng Zhou}, \bibinfo{person}{Xing Ji},
  \bibinfo{person}{Zhifeng Li}, \bibinfo{person}{Dihong Gong},
  \bibinfo{person}{Jin Zhou}, {and} \bibinfo{person}{Wenyu Liu}.}
  \bibinfo{year}{2018}\natexlab{c}.
\newblock \showarticletitle{CosFace: Large Margin Cosine Loss for Deep Face
  Recognition}.
\newblock \bibinfo{journal}{\emph{2018 IEEE/CVF Conference on Computer Vision
  and Pattern Recognition}} (\bibinfo{year}{2018}),
  \bibinfo{pages}{5265--5274}.
\newblock


\bibitem[\protect\citeauthoryear{Wang, Liu, Zhu, Liu, Tao, Kautz, and
  Catanzaro}{Wang et~al\mbox{.}}{2018a}]%
        {wang2018vid2vid}
\bibfield{author}{\bibinfo{person}{Ting-Chun Wang}, \bibinfo{person}{Ming-Yu
  Liu}, \bibinfo{person}{Jun-Yan Zhu}, \bibinfo{person}{Guilin Liu},
  \bibinfo{person}{Andrew Tao}, \bibinfo{person}{Jan Kautz}, {and}
  \bibinfo{person}{Bryan Catanzaro}.} \bibinfo{year}{2018}\natexlab{a}.
\newblock \showarticletitle{Video-to-Video Synthesis}. In
  \bibinfo{booktitle}{\emph{Advances in Neural Information Processing Systems
  (NeurIPS)}}.
\newblock


\bibitem[\protect\citeauthoryear{Wang, Liu, Zhu, Tao, Kautz, and
  Catanzaro}{Wang et~al\mbox{.}}{2018b}]%
        {wang2018pix2pixHD}
\bibfield{author}{\bibinfo{person}{Ting-Chun Wang}, \bibinfo{person}{Ming-Yu
  Liu}, \bibinfo{person}{Jun-Yan Zhu}, \bibinfo{person}{Andrew Tao},
  \bibinfo{person}{Jan Kautz}, {and} \bibinfo{person}{Bryan Catanzaro}.}
  \bibinfo{year}{2018}\natexlab{b}.
\newblock \showarticletitle{High-Resolution Image Synthesis and Semantic
  Manipulation with Conditional GANs}. In \bibinfo{booktitle}{\emph{Proceedings
  of the IEEE Conference on Computer Vision and Pattern Recognition}}.
\newblock


\bibitem[\protect\citeauthoryear{Wang, Mallya, and Liu}{Wang
  et~al\mbox{.}}{2021}]%
        {Wang2021OneShotFN}
\bibfield{author}{\bibinfo{person}{Ting-Chun Wang}, \bibinfo{person}{Arun
  Mallya}, {and} \bibinfo{person}{Ming-Yu Liu}.}
  \bibinfo{year}{2021}\natexlab{}.
\newblock \showarticletitle{One-Shot Free-View Neural Talking-Head Synthesis
  for Video Conferencing}.
\newblock \bibinfo{journal}{\emph{2021 IEEE/CVF Conference on Computer Vision
  and Pattern Recognition (CVPR)}}.
\newblock


\bibitem[\protect\citeauthoryear{Wang, Bovik, Sheikh, and Simoncelli}{Wang
  et~al\mbox{.}}{2004}]%
        {Wang2004ImageQA}
\bibfield{author}{\bibinfo{person}{Zhou Wang}, \bibinfo{person}{Alan~Conrad
  Bovik}, \bibinfo{person}{Hamid~R. Sheikh}, {and} \bibinfo{person}{Eero~P.
  Simoncelli}.} \bibinfo{year}{2004}\natexlab{}.
\newblock \showarticletitle{Image quality assessment: from error visibility to
  structural similarity}.
\newblock \bibinfo{journal}{\emph{IEEE Transactions on Image Processing}}
  \bibinfo{volume}{13} (\bibinfo{year}{2004}), \bibinfo{pages}{600--612}.
\newblock


\bibitem[\protect\citeauthoryear{Yang, Vo, Neverova, Ramanan, Vedaldi, and
  Joo}{Yang et~al\mbox{.}}{2021}]%
        {Yang2021BANMoBA}
\bibfield{author}{\bibinfo{person}{Gengshan Yang}, \bibinfo{person}{Minh Vo},
  \bibinfo{person}{Natalia Neverova}, \bibinfo{person}{Deva Ramanan},
  \bibinfo{person}{Andrea Vedaldi}, {and} \bibinfo{person}{Hanbyul Joo}.}
  \bibinfo{year}{2021}\natexlab{}.
\newblock \showarticletitle{BANMo: Building Animatable 3D Neural Models from
  Many Casual Videos}.
\newblock \bibinfo{journal}{\emph{ArXiv}}.
\newblock


\bibitem[\protect\citeauthoryear{Yang, Liu, Wang, Wang, Ren, Ma, and Gao}{Yang
  et~al\mbox{.}}{2020}]%
        {Yang2020HiFaceGANFR}
\bibfield{author}{\bibinfo{person}{Lingbo Yang}, \bibinfo{person}{C. Liu},
  \bibinfo{person}{P. Wang}, \bibinfo{person}{Shanshe Wang},
  \bibinfo{person}{P. Ren}, \bibinfo{person}{Siwei Ma}, {and}
  \bibinfo{person}{W. Gao}.} \bibinfo{year}{2020}\natexlab{}.
\newblock \showarticletitle{HiFaceGAN: Face Renovation via Collaborative
  Suppression and Replenishment}.
\newblock \bibinfo{journal}{\emph{Proceedings of the 28th ACM International
  Conference on Multimedia}} (\bibinfo{year}{2020}).
\newblock


\bibitem[\protect\citeauthoryear{Zakharov, Ivakhnenko, Shysheya, and
  Lempitsky}{Zakharov et~al\mbox{.}}{2020}]%
        {Zakharov2020FastBN}
\bibfield{author}{\bibinfo{person}{Egor Zakharov}, \bibinfo{person}{Aleksei
  Ivakhnenko}, \bibinfo{person}{Aliaksandra Shysheya}, {and}
  \bibinfo{person}{Victor~S. Lempitsky}.} \bibinfo{year}{2020}\natexlab{}.
\newblock \showarticletitle{Fast Bi-layer Neural Synthesis of One-Shot
  Realistic Head Avatars}. In \bibinfo{booktitle}{\emph{ECCV}}.
\newblock


\bibitem[\protect\citeauthoryear{Zakharov, Shysheya, Burkov, and
  Lempitsky}{Zakharov et~al\mbox{.}}{2019}]%
        {Zakharov2019FewShotAL}
\bibfield{author}{\bibinfo{person}{Egor Zakharov}, \bibinfo{person}{Aliaksandra
  Shysheya}, \bibinfo{person}{Egor Burkov}, {and} \bibinfo{person}{Victor~S.
  Lempitsky}.} \bibinfo{year}{2019}\natexlab{}.
\newblock \showarticletitle{Few-Shot Adversarial Learning of Realistic Neural
  Talking Head Models}.
\newblock \bibinfo{journal}{\emph{2019 IEEE/CVF International Conference on
  Computer Vision (ICCV)}}.
\newblock


\bibitem[\protect\citeauthoryear{Zhang, Isola, Efros, Shechtman, and
  Wang}{Zhang et~al\mbox{.}}{2018}]%
        {Zhang2018TheUE}
\bibfield{author}{\bibinfo{person}{Richard Zhang}, \bibinfo{person}{Phillip
  Isola}, \bibinfo{person}{Alexei~A. Efros}, \bibinfo{person}{Eli Shechtman},
  {and} \bibinfo{person}{Oliver Wang}.} \bibinfo{year}{2018}\natexlab{}.
\newblock \showarticletitle{The Unreasonable Effectiveness of Deep Features as
  a Perceptual Metric}.
\newblock \bibinfo{journal}{\emph{2018 IEEE/CVF Conference on Computer Vision
  and Pattern Recognition}} (\bibinfo{year}{2018}), \bibinfo{pages}{586--595}.
\newblock


\bibitem[\protect\citeauthoryear{Zhang, Zhu, Lei, Shi, Wang, and Li}{Zhang
  et~al\mbox{.}}{2017}]%
        {Zhang2017S3FDSS}
\bibfield{author}{\bibinfo{person}{Shifeng Zhang}, \bibinfo{person}{Xiangyu
  Zhu}, \bibinfo{person}{Zhen Lei}, \bibinfo{person}{Hailin Shi},
  \bibinfo{person}{Xiaobo Wang}, {and} \bibinfo{person}{S. Li}.}
  \bibinfo{year}{2017}\natexlab{}.
\newblock \showarticletitle{S3FD: Single Shot Scale-Invariant Face Detector}.
\newblock \bibinfo{journal}{\emph{2017 IEEE International Conference on
  Computer Vision (ICCV)}} (\bibinfo{year}{2017}), \bibinfo{pages}{192--201}.
\newblock


\bibitem[\protect\citeauthoryear{Zhu, Park, Isola, and Efros}{Zhu
  et~al\mbox{.}}{2017}]%
        {Zhu2017UnpairedIT}
\bibfield{author}{\bibinfo{person}{Jun-Yan Zhu}, \bibinfo{person}{Taesung
  Park}, \bibinfo{person}{Phillip Isola}, {and} \bibinfo{person}{Alexei~A.
  Efros}.} \bibinfo{year}{2017}\natexlab{}.
\newblock \showarticletitle{Unpaired Image-to-Image Translation Using
  Cycle-Consistent Adversarial Networks}.
\newblock \bibinfo{journal}{\emph{2017 IEEE International Conference on
  Computer Vision (ICCV)}} (\bibinfo{year}{2017}), \bibinfo{pages}{2242--2251}.
\newblock


\end{thebibliography}

\appendix
\section{Network architectures}

Here we describe the architectures of our networks, conduct an in-depth evaluation of the architectural choices and provide details about preprocessing of the dataset.

\begin{figure*}[!h]
    \centering    
    \setlength{\wid}{0.19\textwidth}
    \setlength{\mrg}{-0.3cm}
    \setlength{\mrgv}{0cm}
    \begin{tabular}{cccc}
        \includegraphics[width=\wid]{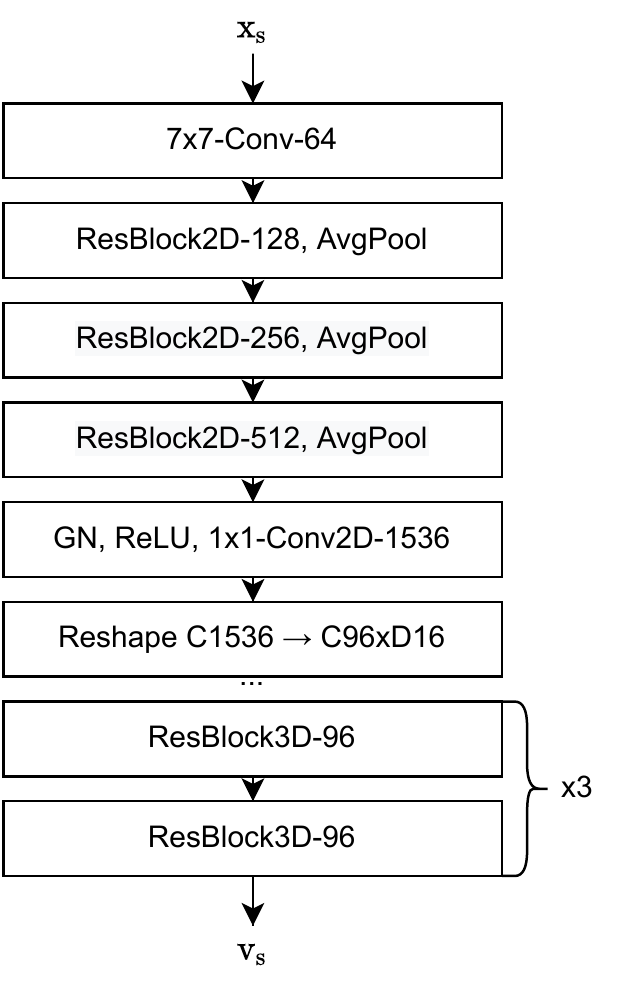} & 
        \hspace{\mrg}
        \vspace{\mrgv}
        \includegraphics[width=\wid]{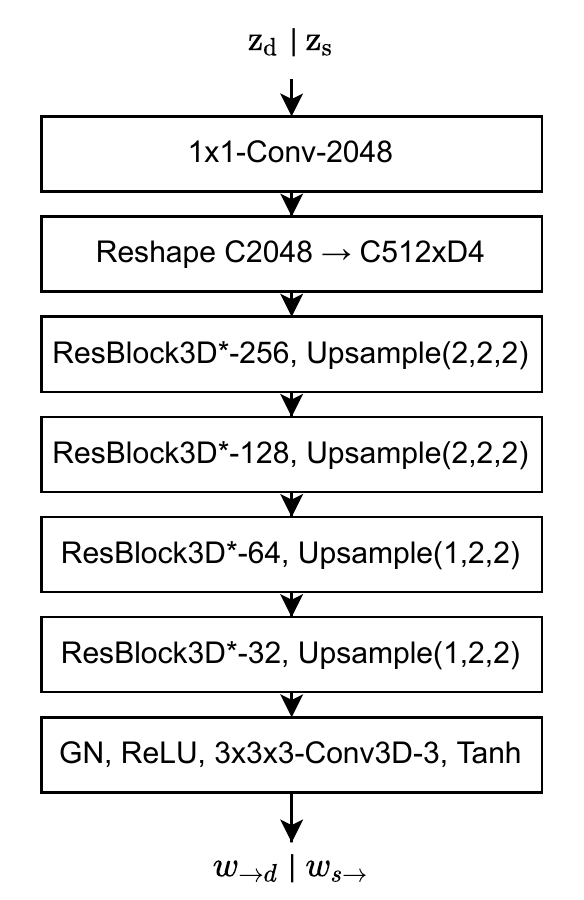} & 
        \hspace{\mrg}
        \includegraphics[width=\wid]{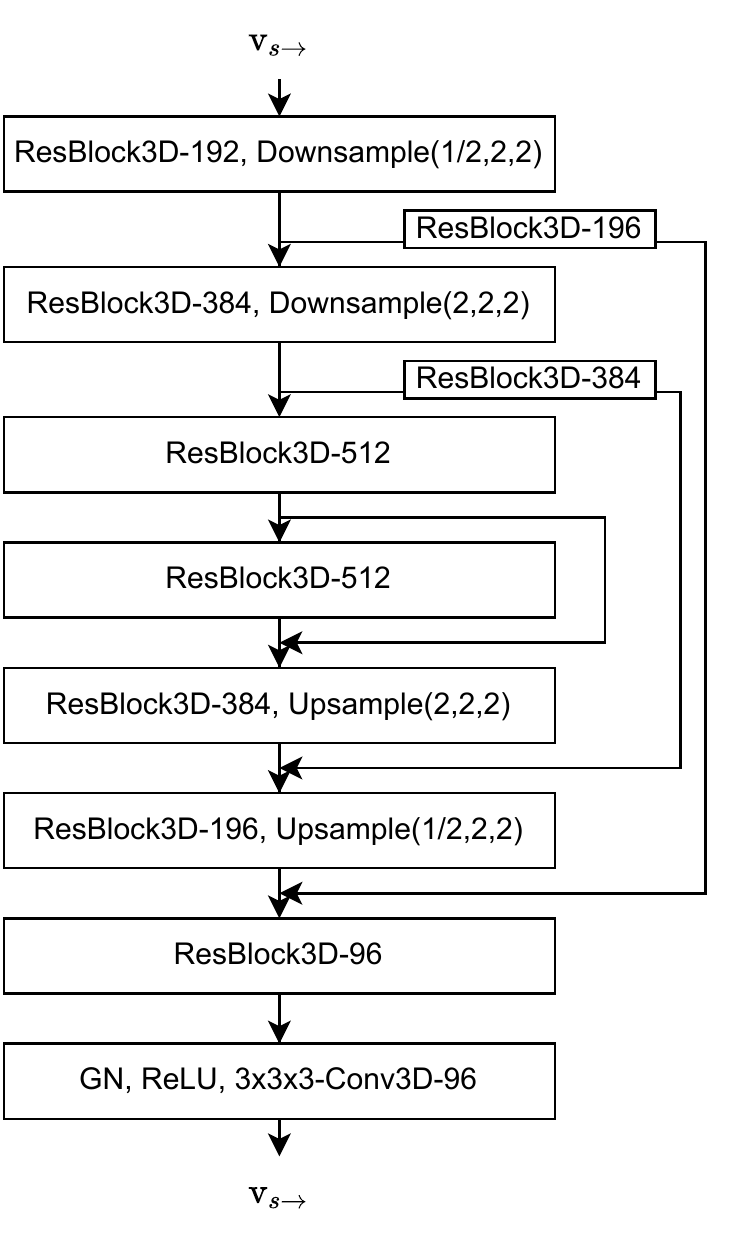} & 
        \hspace{\mrg}
        \includegraphics[width=\wid]{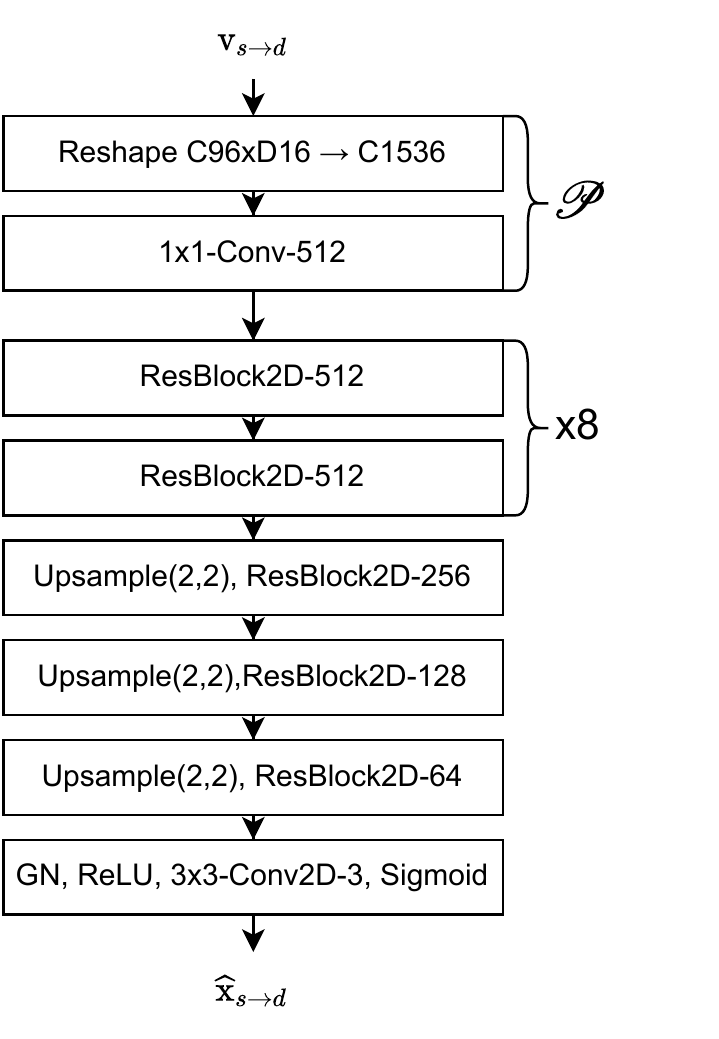}
        \\ %
        \textbf{$\E_\text{app}$ (a)} & 
        \hspace{\mrg} 
        \textbf{$\W_{\rightarrow d}$ |  $\W_{s \rightarrow}$ (b)} & 
        \hspace{\mrg} 
        \textbf{$\G_\text{3D}$ (c)} & 
        \hspace{\mrg} 
        \textbf{$\G_\text{2D}$ (d)}
    \end{tabular}
    \vspace{-0.4cm}
    \caption{Architectures of components of $\G_\text{base}$. }
    \label{fig:schemes_base}
\end{figure*}

\subsection{Base model}

In the architecture of our base model, we replace all BatchNorms with GroupNorms, and all convolutional layers, except the first and the last ones, are used with weight standardization. 

\noindent
\newline \textbf{Appearance encoder $\E_\text{app}$.} 
The network consists of two parts. The first part produces a 4D tensor of volumetric features $v_s$ that represent the person's appearance from the source image. It includes several residual blocks followed by average pooling. We reshape the resulting 2D features to 3D features and then apply several 3D residual blocks to compute the final volumetric representation. The scheme shown in Figure~\ref{fig:schemes_base} (a). 

The second part produces a global descriptor $\e_s$ that helps retain the appearance of the output image. We use a ResNet-50 architecture with custom residual blocks. The architecture of our residual block can be seen in Figure~\ref{fig:schemes_hr} (c), where $n$ denotes the dimension of a convolutional layer (either 2D or 3D) and $\x$ denotes the number of output channels.  

\noindent
\newline  \textbf{Motion encoder $\E_\text{mtn}$.}
We use two separate ResNet-18 networks as encoders to separately predict the head pose and expression vector. The head pose prediction network is pre-trained, while the expression prediction network is trained from scratch.

\noindent
\newline  \textbf{Warping generators $\W_{s\rightarrow}$ and  $\W_{\rightarrow d}$}.
When both source and driver tuples $(\R_{s/d}, \t_{s/d}, \z_{s/d}, \e_{s})$ are predicted, we can produce 3D warping  $\w_{s\rightarrow}$ and $\w_{\rightarrow d}$ . Both warpings consist of two parts: one in charge of rotation and translation ($\w_{...}^\text{rt}$) and another one in charge of emotion changing ($\w_{...}^\text{em}$). 

To get the first part: for $\w_{\rightarrow d}^\text{rt}$ we multiply identity grid on transformation matrix and for $\w_{s\rightarrow}^\text{rt}$ we multiply identity grid on inversed transformation matrix.

To get $\w_{s\rightarrow}^\text{em}$ and $\w_{\rightarrow d}^\text{em}$ we use two separate warping generators (see Figure~\ref{fig:schemes_base} (b)) with the same architecture contain several 3D residual blocks where all GroupNorms changed on Adaptive GroupNorms (marked as ResBlock3D* on the scheme), whereas inputs we use sums $\z_{s}+\e_{s}$ and $\z_{d}+\e_{s}$ respectively. To generate adaptive parameters, we multiply the foregoing sums and additionally learned matrices for each pair of parameters.

\noindent
\newline  \textbf{3D convolutional network $\G_\text{3D}$}. 
Next, we process volumetric representation after the first warping to get canonical volume where source motion removed from the appearance features. For this, we apply Unet-like architecture with several downsample units consists of 3D residual block and downsample operation, followed by the same number of upsample units consists of 3D residual block and upsample operation. The scheme shown in Figure~\ref{fig:schemes_base} (c). $\text{Sample(z, x, y)}$ mean sample operation that changes depth, height, width in z, x, y times respectively. For example, z=1/2 means downsample along depth dimension in 2 times and z=2 means upsample along depth dimension in 2 times.  

\noindent
\newline  \textbf{2D convolutional network $\G_\text{2D}$}. 
Finally, to predict an output image from a processed volume, we first use orthographically projection $\mathcal{P}$ that is consists of reshape operation and 1x1 convolution. While more complex projection operators could be used (like volumetric ray marching), we found such simple approach is sufficient for our applications, the same way as it has been done in ~\cite{wang2018vid2vid}. Then we utilize the network includes 8 residual blocks on the same resolution and number of feature maps, then gradually apply units contain upsampling and residual blocks with successively decreasing number of output channels. The scheme shown in Figure~\ref{fig:schemes_base} (d).

\begin{figure*}[!h]
    \centering    
    \setlength{\wid}{0.19\textwidth}
    \setlength{\mrg}{-0.3cm}
    \setlength{\mrgv}{0cm}
    \begin{tabular}{ccc}
        \includegraphics[width=\wid]{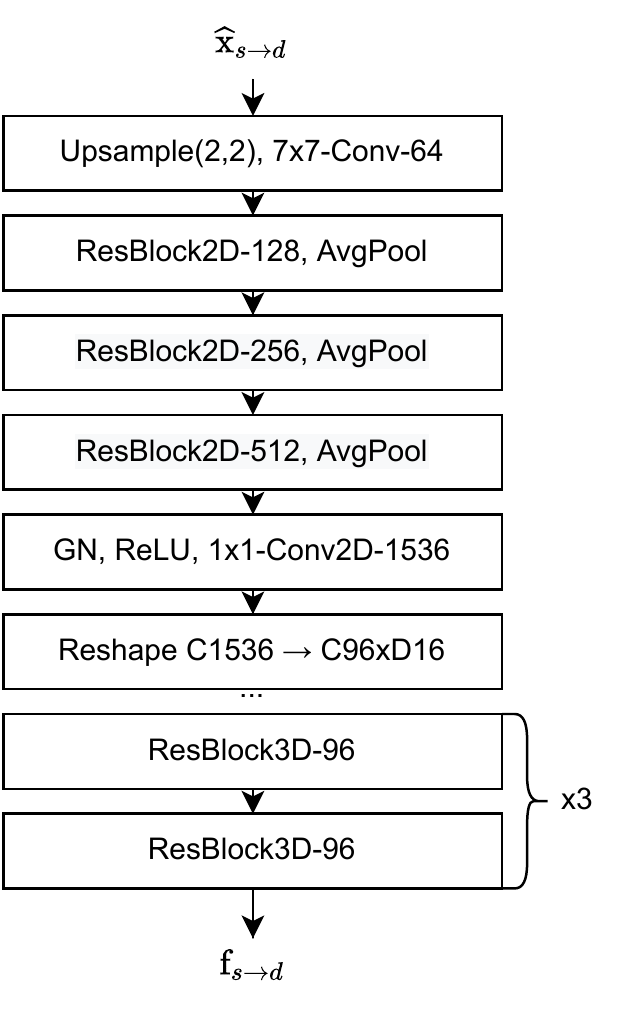} & 
        \hspace{\mrg}
        \includegraphics[width=\wid]{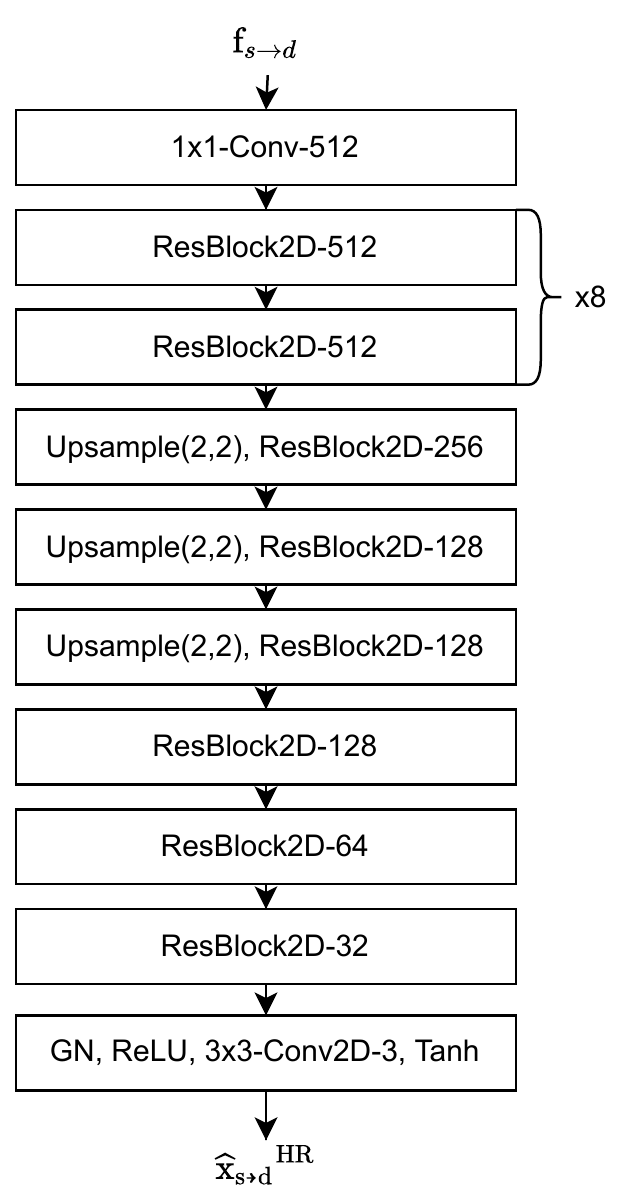} &
        \hspace{\mrg}
        \includegraphics[width=\wid]{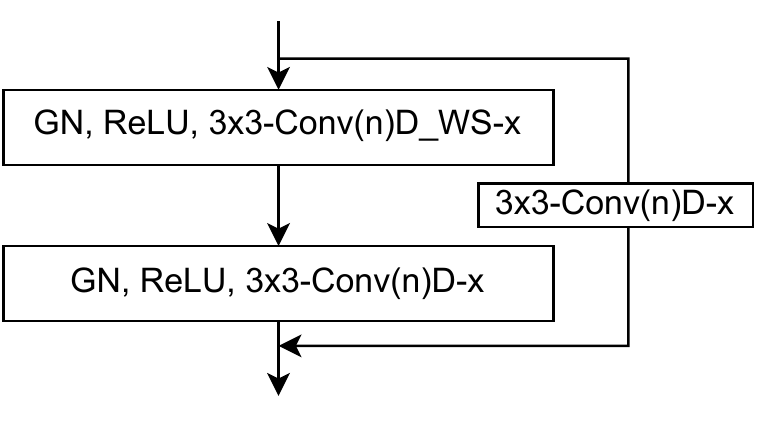}
        \\ %
        \textbf{Encoder (a)} & 
        \hspace{\mrg} 
        \textbf{Decoder (b)} & 
        \hspace{\mrg} 
        \textbf{Resblock (c)}
        \hspace{\mrg} 
    \end{tabular}
    \vspace{-0.4cm}
    \caption{Architectures of components of $\G_\text{enh}$.}
    \label{fig:schemes_hr}
\end{figure*}

\subsection{High-resolution model}
High-resolution model contain 2 parts: encoder and decoder. Both of them you can see on the scheme shown in Figure~\ref{fig:schemes_hr}. \textbf{Encoder}, that takes $\hat{\x}_{s->d}$ as an input, contain just conv layer followed by 2 residual blocks and produce 3D feature tensor $\text{f}_{s->d}$. \textbf{Decoder}, that takes output features $\text{f}_{s->d}$ and produce hi-resolution version of input $\hat{\x}_{s->d}^{\text{HR}}$, recalls 2D convolutional network from the base model, it also includes 8 residual blocks on the same resolution and number of feature maps, followed by two upsampling with residual blocks and three residual blocks on high-resolution.

\begin{figure*}[!h]
    \centering    
    \setlength{\wid}{0.19\textwidth}
    \setlength{\mrg}{-0.3cm}
    \setlength{\mrgv}{0cm}
    \resizebox{10cm}{!}{
    \begin{tabular}{ccc}
        \includegraphics[scale=1]{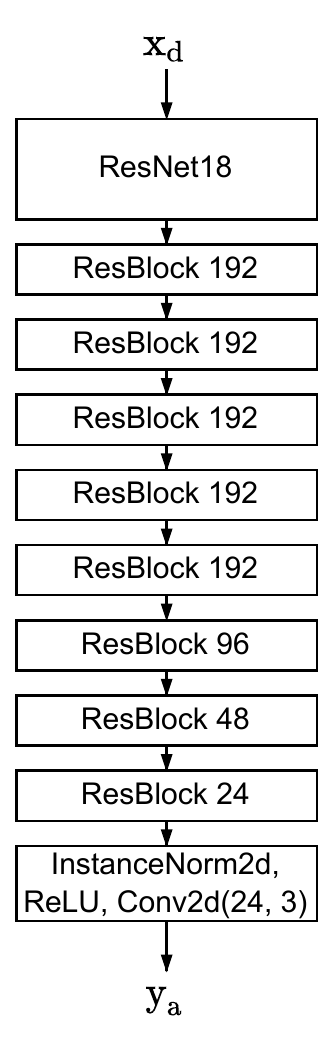} & 
        \hspace{\mrg}
        \includegraphics[scale=1]{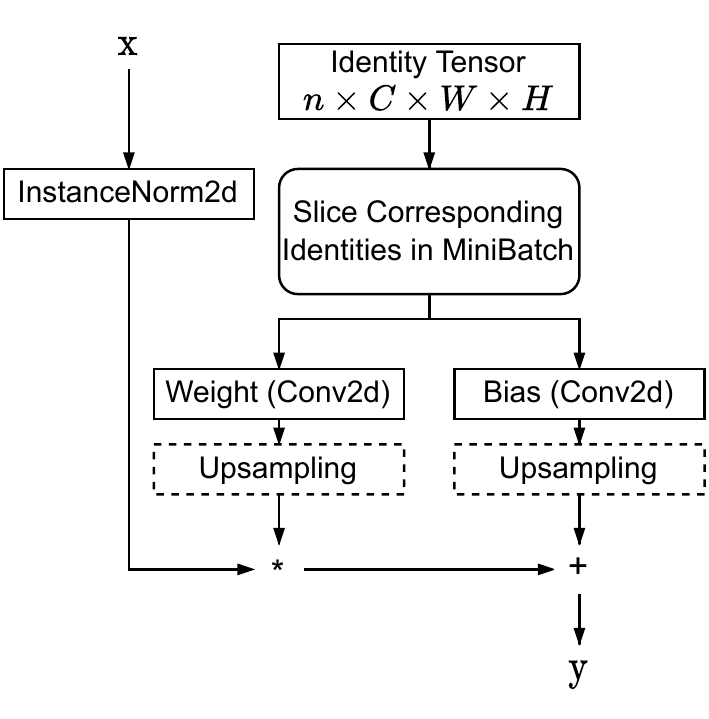} &
        \hspace{\mrg}
        \includegraphics[scale=1]{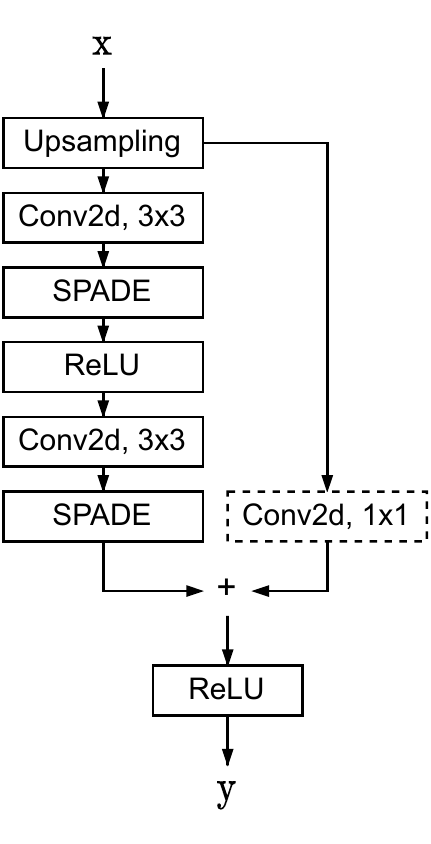} \\
        \textbf{Student model (a)} & 
        \hspace{\mrg} 
        \textbf{Adaptive normalization (SPADE) (b)} & 
        \hspace{\mrg} 
        \textbf{Residual block (c)}
        \hspace{\mrg} 
    \end{tabular}}
    
    \caption{Architectures of components of the student model. Dashed lines correspond to the optional blocks (i.e. used only if channel/resolution configuration needs to apply some transform, either upsampling or change the number of channels). Each SPADE tensor shape $W \times H$ is at most $64 \times 64$.}
    \label{fig:schemes_hr}
\end{figure*}

\subsection{Student model}

 The encoder of a student model is ResNet18, and the generator consists of residual blocks with SPADE normalization layers, in each SPADE block a tensor used for normalization is fixed for a specific avatar. During the forward pass we select which tensor to use in normalization layer to switch between predefined avatars. Using such procedure during the training, we force our model to store all the identity-specific information into SPADE blocks. Also, to compress the final model we tweak a size of spatial dimension of normalization tensors (which dominate the size of the whole model) in SPADE blocks: by default these tensors must be of the same shape as a corresponding input feature map, instead, we compress them spatially and use bilinear upsampling to output the feature map of the right size. More precisely, we bound the resolution of an inner identity tensor by 64.
\section{Additional information}
\subsection{Training details}

\textbf{$\G_\text{base}$ and $\G_\text{HR}$}.  As augmentation for both source and target images, we use color jitter and random flip. As for driving image, before sending it to $\E_\text{mtn}$ we do a center crop around the face of a person. Next, we augment it using a random warping based on thin-plate-splines, which severely degrades the shape of the facial features, yet keeps the expression intact (ex., it cannot close or open eyes or change the eyes' direction). Finally, we apply a severe color jitter.

For $\mathcal{L}_\text{GAN}$ loss we use multi-resolution patchGAN where the discriminator produces the patch-level prediction. We apply spectral normalization for both $\G_\text{base}$ and $\G_\text{HR}$.

For AdamW optimizer we used the following parameters: betas=(0.5, 0.999), eps=1e-8, weight decay=1e-2 for both $\G_\text{base}$ and $\G_\text{HR}$ and correspond discriminator.

\begin{figure*}
    \centering    
    \setlength{\wid}{0.24\textwidth}
    \setlength{\mrg}{-0.3cm}
    \begin{tabular}{cccc}
        \includegraphics[width=\wid]{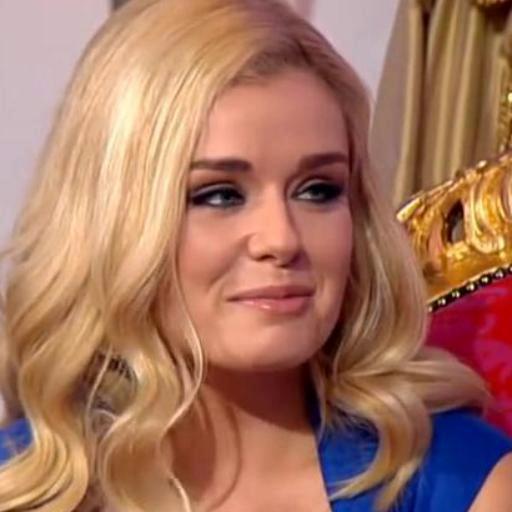} & 
        \hspace{\mrg}
        \vspace{\mrgv}
        \includegraphics[width=\wid]{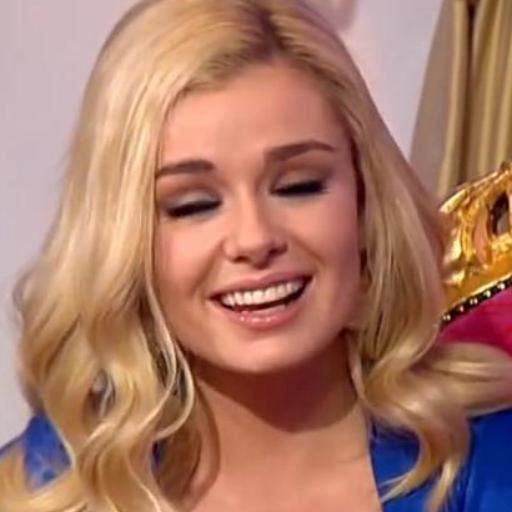} & 
        \hspace{\mrg}
        \includegraphics[width=\wid]{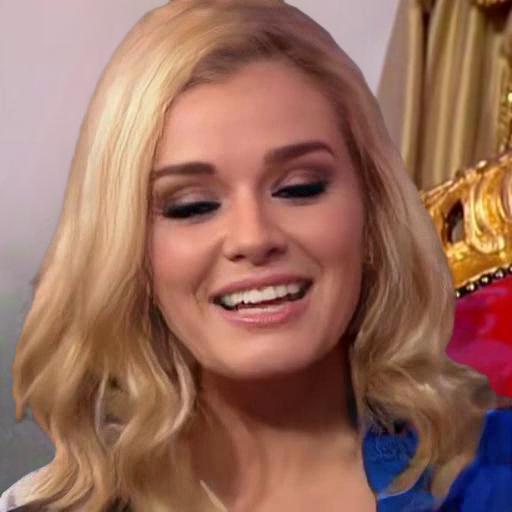} & 
        \hspace{\mrg}
        \includegraphics[width=\wid]{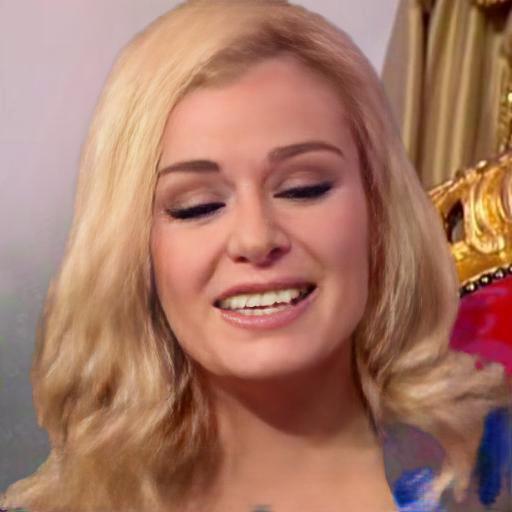}
        \\ %
        \textbf{Source} & 
        \hspace{\mrg} 
        \textbf{Driver} & 
        \hspace{\mrg} 
        \textbf{Teacher} & 
        \hspace{\mrg} 
        \textbf{Student}
    \end{tabular}
    \vspace{-0.4cm}
    \caption{Result of student model in self-reenactment mode.}
    \label{fig:student_self}
\end{figure*}

\noindent
\newline \textbf{Student}
\newline
The student model was trained to predict the corresponding prediction of the teacher model for a fixed set of identities. We used a standard set of losses for such setup, specifically, adversarial and three kinds of perceptual losses. Adversarial training was done with multiscale discriminator on four resolutions. Perceptual losses are the same that were used to train teacher model: standard VGG19 loss, gaze loss and VGG Face loss. Additionally, to check how student model handle self-reenactment mode, we train separate student model on 10 avatars, using persons from 10 random test videos. Student model achieved PSNR of 19.25 and SSIM of 0.682 (while teacher model achieved 21.34 and 0.768 correspondingly). You can see an example in Figure~\ref{fig:student_self}.

\subsection{Two stage training}
Initially, we have evaluated some of the feasible configurations for the end-to-end training. First of all, end-to-end training with the full enhancer network or even a single decoder layer at 1024x1024 resolution would not fit into the memory of our available GPUs. We, therefore, tried freezing a pre-trained encoder and fine-tuning a decoder with an additional upsampling block, combining both of the objectives and with some weighting coefficient. We have observed a significant decrease in the quality of the results, compared to a separately trained network, across three different weights. Since it is effectively doing super-resolution without high-resolution conditioning, maintaining its high capacity is crucial for the network to generate the missing high-frequency details. You can see some comparison in Figure~\ref{fig:schemes_base}.

\subsection{Datasets preprocessing}
We obtain the VoxCeleb2HQ dataset by first downloading the original videos from the VoxCeleb2~\cite{Chung2018VoxCeleb2DS} dataset. These videos are processed using an off-the-shelf face~\cite{Zhang2017S3FDSS} and keypoints~\cite{Bulat2017HowFA} detectors and cropped frame-by-frame around the head regions. Then, the obtained cropped frames are first filtered by their resolution, to exclude all crops that are smaller than $512 \times 512$. We call this process a bitrade filtering. Then, we additionally rank the frames from the remaining videos by their image quality assessment (IQA) scores, calculated using a pre-trained system~\cite{hyperIQA}. We then remove the bottom 50\% of the videos using the mean IQA score across its frames, thus arriving at the 15,000 videos that we use for training.

\subsection{Evaluation of the architectural choices}
\begin{figure*}[!ht]
    \centering
    \includegraphics[width=\textwidth]{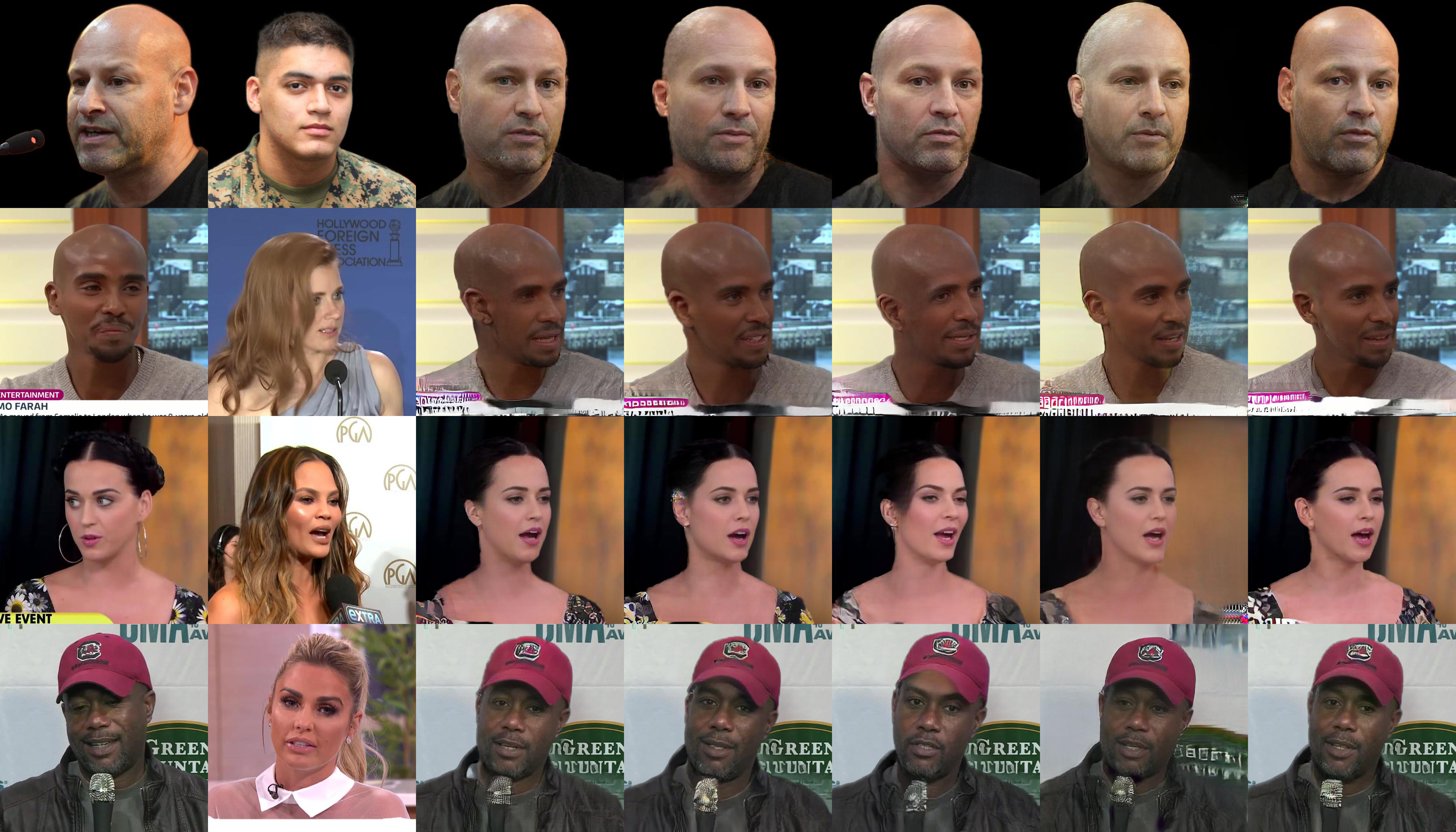}

    \setlength{\mrg}{0.7cm}
    \begin{tabular}{cccccccc}
        \textbf{Source} \hspace{0.8cm} & 
        \textbf{Driver} \hspace{0.8cm} & 
        \textbf{w/o $\e_{s}$ (a)} \hspace{0.8cm} & 
        \textbf{w/o $\w_{s\rightarrow}$ (b)} \hspace{0.8cm} & 
        \textbf{w/o augs (c)} \hspace{0.8cm} & 
        \textbf{w/ BG (d)} \hspace{0.8cm} & 
        \textbf{Ours}
    \end{tabular}
    
    \caption{Additional ablation study. We qualitatively evaluate the individual components of our base model (last column). We observe the positive influence of crucial part of our method. The details of the evaluation described in Section 2.3. }
    \label{fig:ablation_scheme}
\end{figure*}

 In addition to the ablations we describe in the main part of the paper, we decided to conduct a series of additional experiments to evaluate the impact of the key features in our method. We demonstrate results of an additional ablation study in Figure~\ref{fig:ablation_scheme}. The following parts were eliminated separately: (a) architecture without encoder $\e_{s}$, (b) the warping for source image $\w_{s\rightarrow}$, (c) driver augmentation, (d) added block to predict background directly with person appearance, (e) base model.
 We show that $\e_{s}$ helps to preserve identity information, especially on tight turns. Without warping generator $\w_{s\rightarrow}$ the preservation of whole structure of the shoulders and head worsen and artifacts appeared on ears.
 If turn off driver augmentation, apparently the model shows significant worsening of results in terms of identity preservation (see the eyes and ears zones, and  artifacts on the cap and temples).
 Additionally, we train our model to predict background and person together. The preservation of the identity dropped as long as the whole image quality. Mainly because the capacity spent on the background modeling.

\begin{figure}
    \centering    
    \setlength{\wid}{0.111\textwidth}
    \setlength{\mrg}{-0.4cm}
    \setlength{\mrgv}{-0.09cm}
    \begin{tabular}{cccc}
        \vspace{\mrgv}
        \includegraphics[width=\wid]{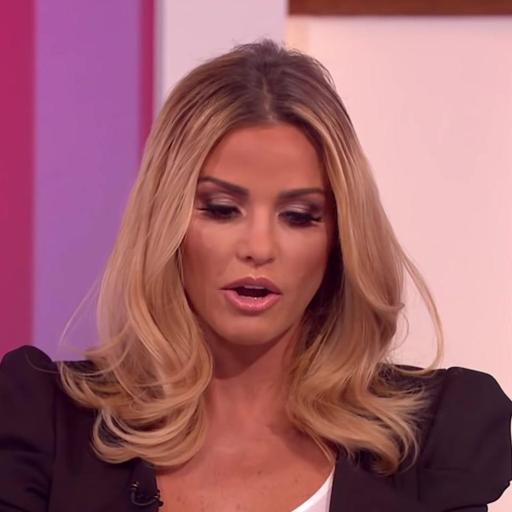} & 
        \hspace{\mrg}
        \includegraphics[width=\wid]{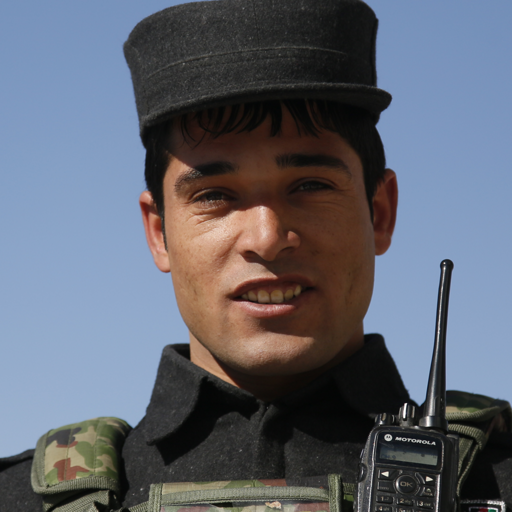} & 
        \hspace{\mrg}
        \includegraphics[width=\wid]{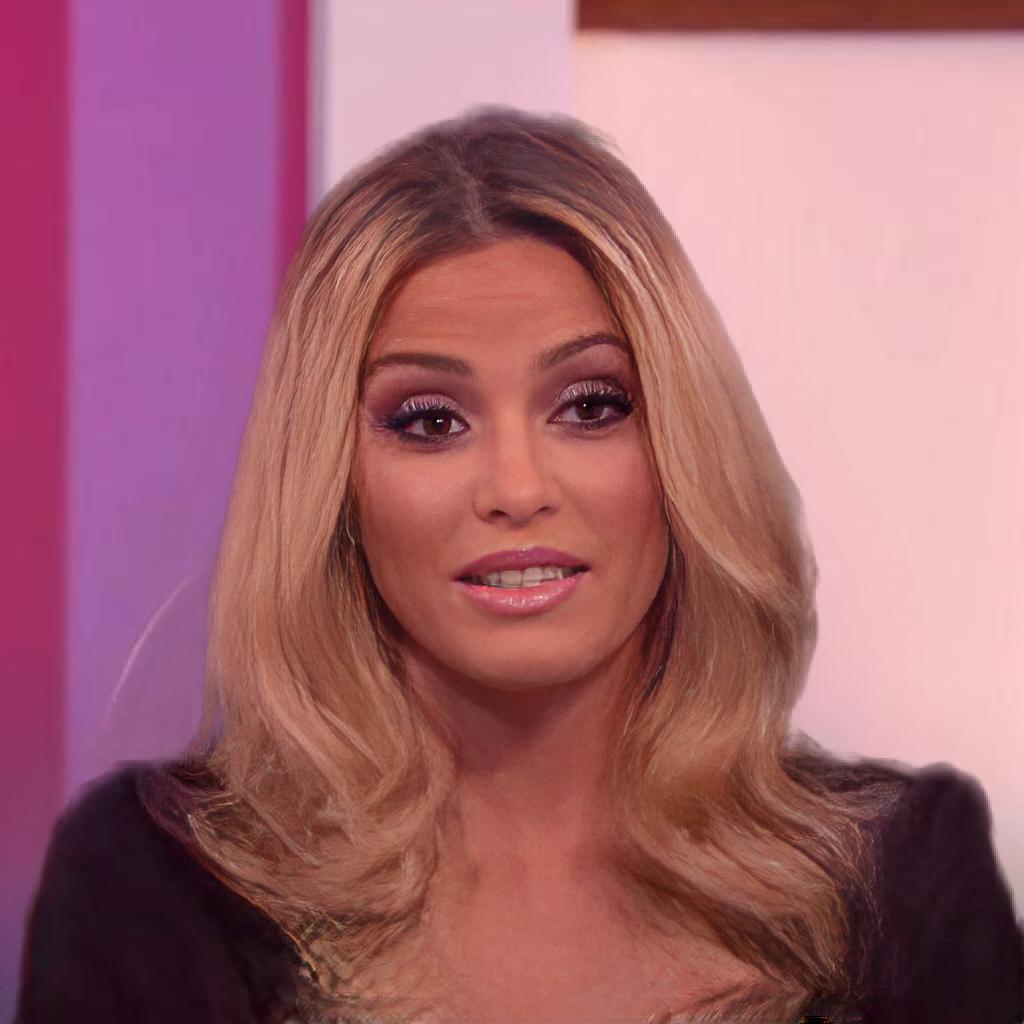} & 
        \hspace{\mrg}
        \includegraphics[width=\wid]{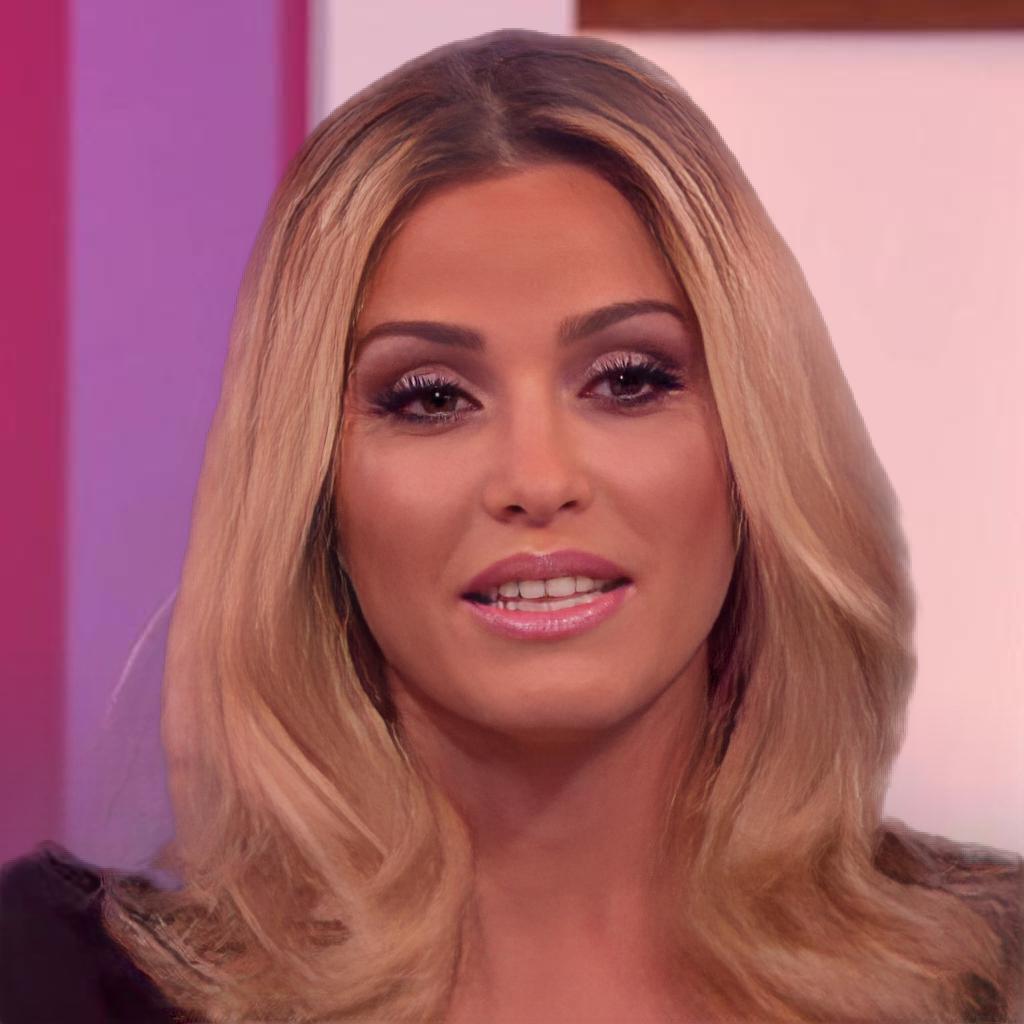}
        
        \\ %
        \includegraphics[width=\wid]{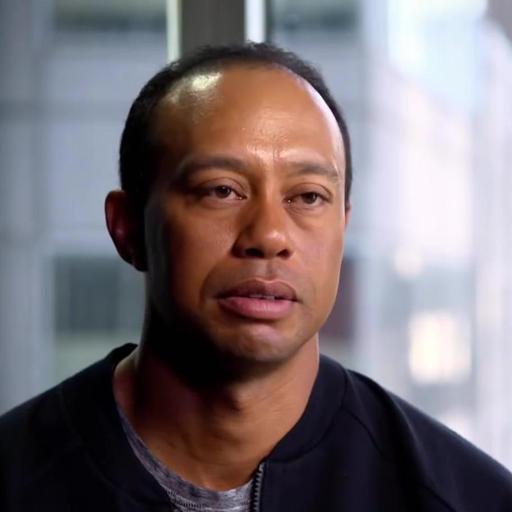} & 
        \hspace{\mrg}
        \includegraphics[width=\wid]{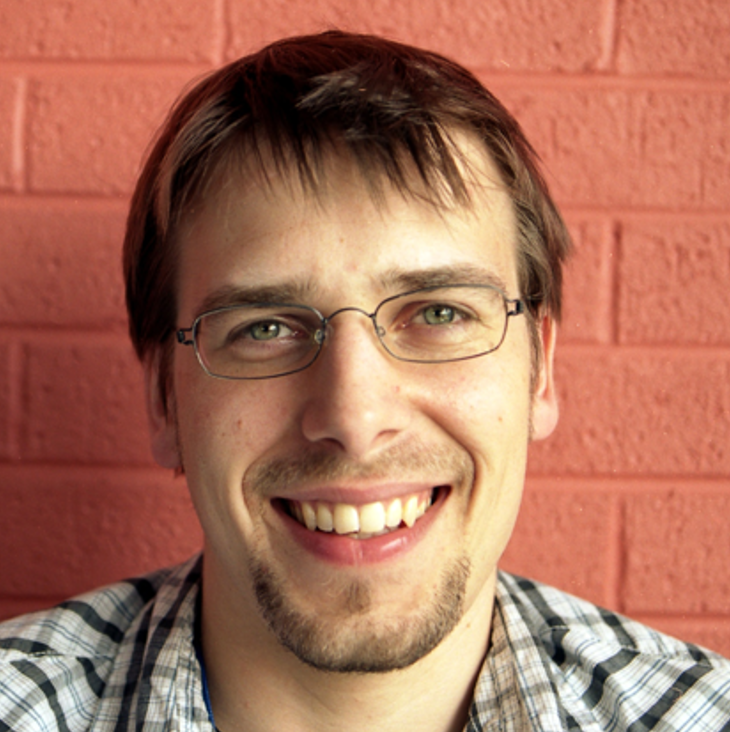} & 
        \hspace{\mrg}
        \includegraphics[width=\wid]{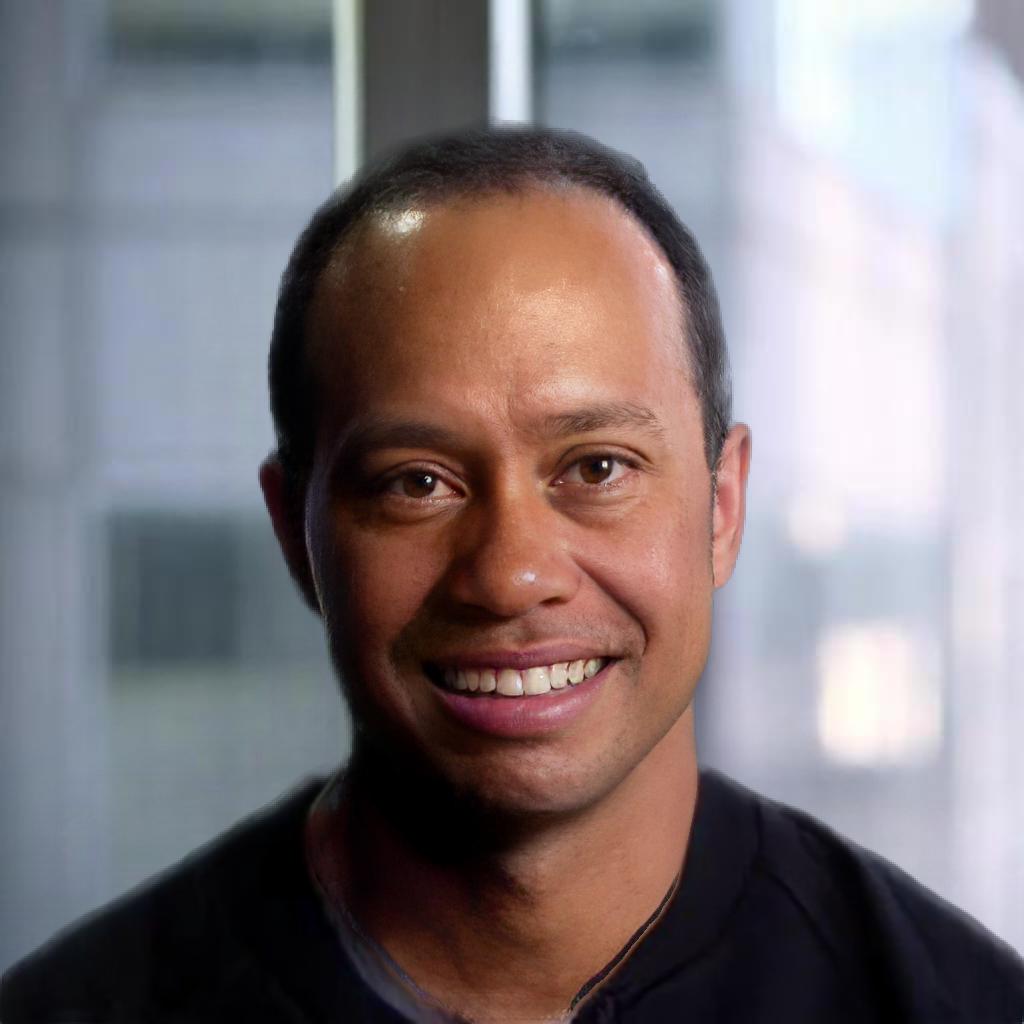} & 
        \hspace{\mrg}
        \includegraphics[width=\wid]{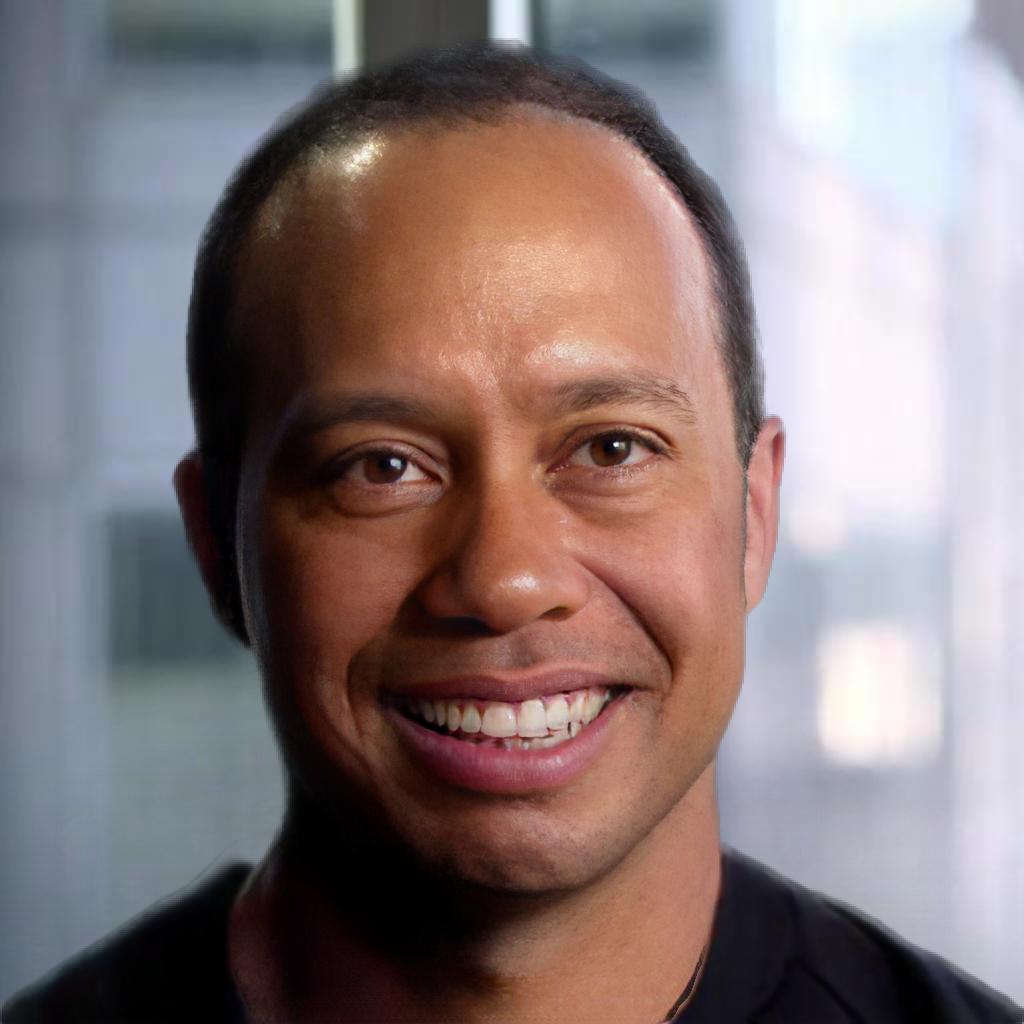} 
        \\
        \textbf{Source} & 
        \hspace{\mrg} 
        \textbf{Driver} & 
        \hspace{\mrg} 
        \textbf{Source $s \text{\small\&} t$} & 
        \hspace{\mrg} 
        \textbf{Driver $s \text{\small\&} t$} 
    \end{tabular}
    \vspace{-0.3cm}
    \caption{Results with different scales and translations }
    \label{fig:scale}
\end{figure}

\begin{figure}
    \centering    
    \setlength{\wid}{0.15\textwidth}
    \setlength{\mrg}{-0.4cm}
    \setlength{\mrgv}{-0.09cm}
    \begin{tabular}{ccc}
        \vspace{\mrgv}
        \includegraphics[width=\wid]{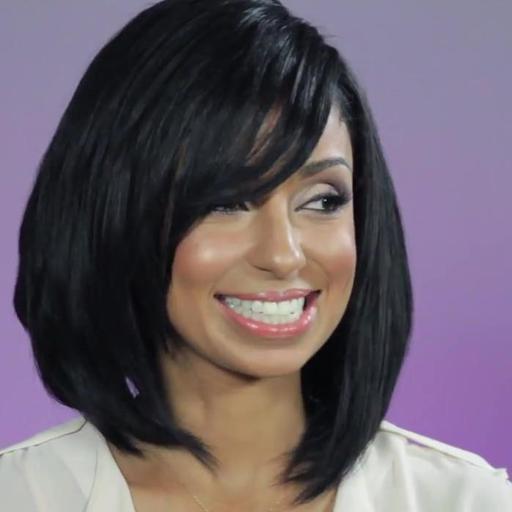} & 
        \hspace{\mrg}
        \includegraphics[width=\wid]{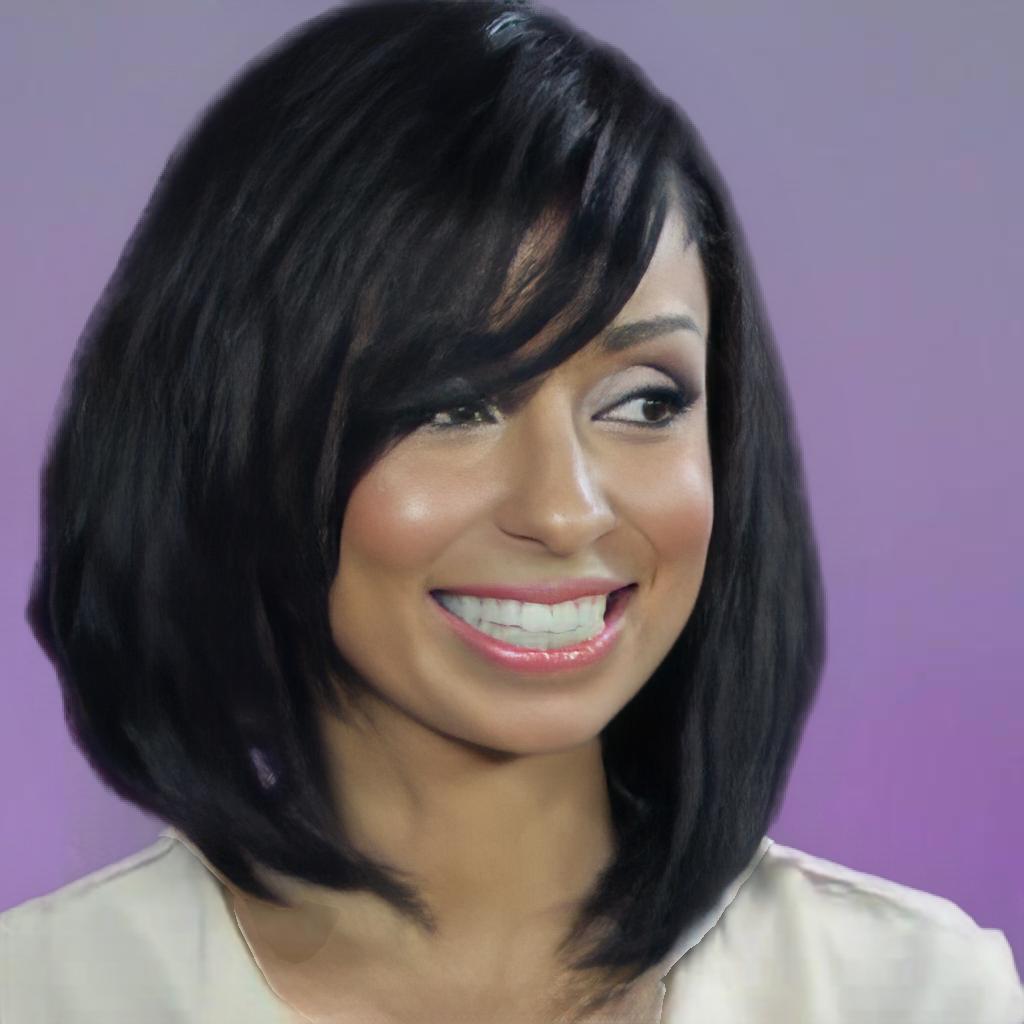} & 
        \hspace{\mrg}
        \includegraphics[width=\wid]{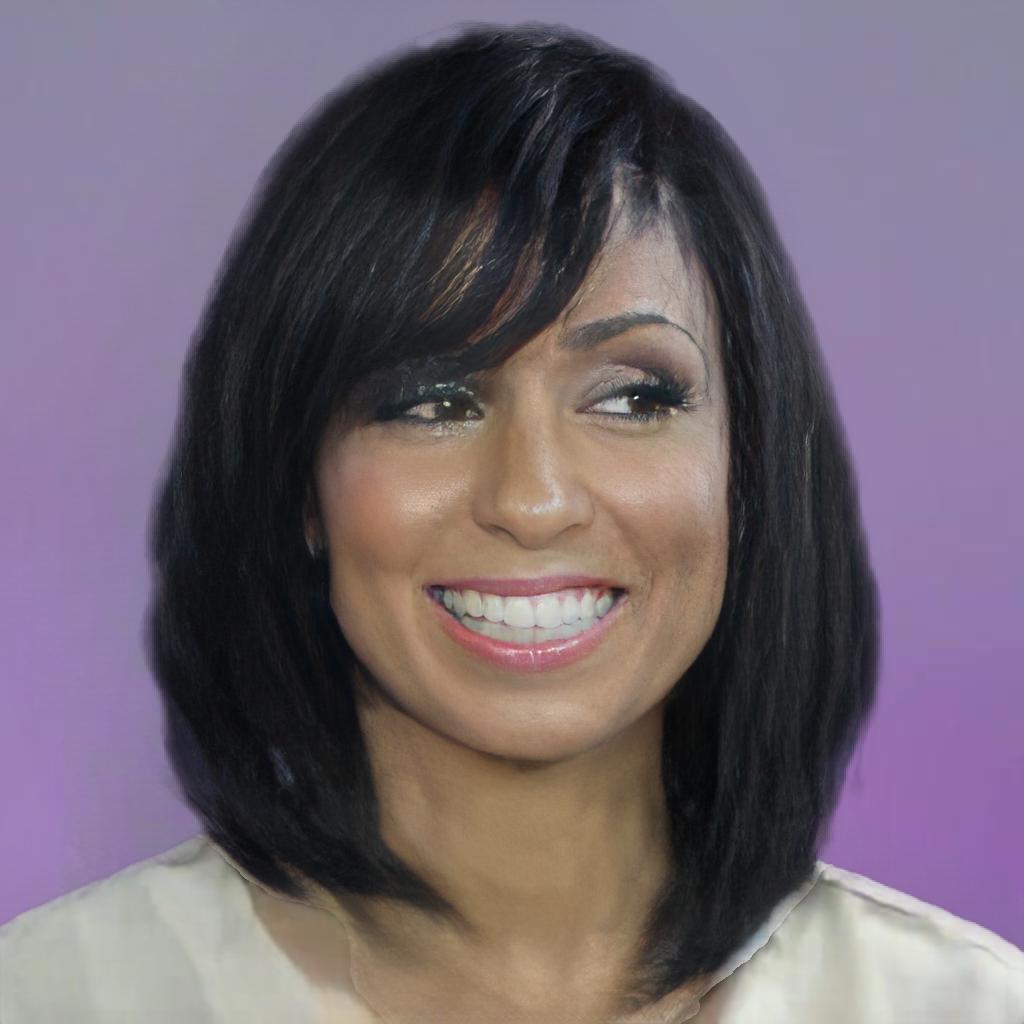}
        \\ %
        \textbf{Source} & 
        \hspace{\mrg} 
        \textbf{Reconstructed} & 
        \hspace{\mrg} 
        \textbf{Frontalized}
    \end{tabular}
    \vspace{-0.4cm}
    \caption{Result of frontalization}
    \label{fig:front}
\end{figure}

\begin{figure*}
    \centering    
    \setlength{\wid}{0.14\textwidth}
    \setlength{\mrg}{-0.4cm}
    \setlength{\mrgv}{-0.09cm}
    \begin{tabular}{ccccccc}
        \vspace{\mrgv}
        \includegraphics[width=\wid]{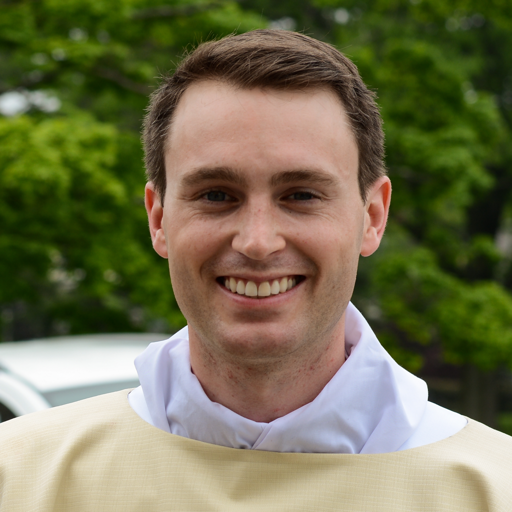} & 
        \hspace{\mrg}
        \includegraphics[width=\wid]{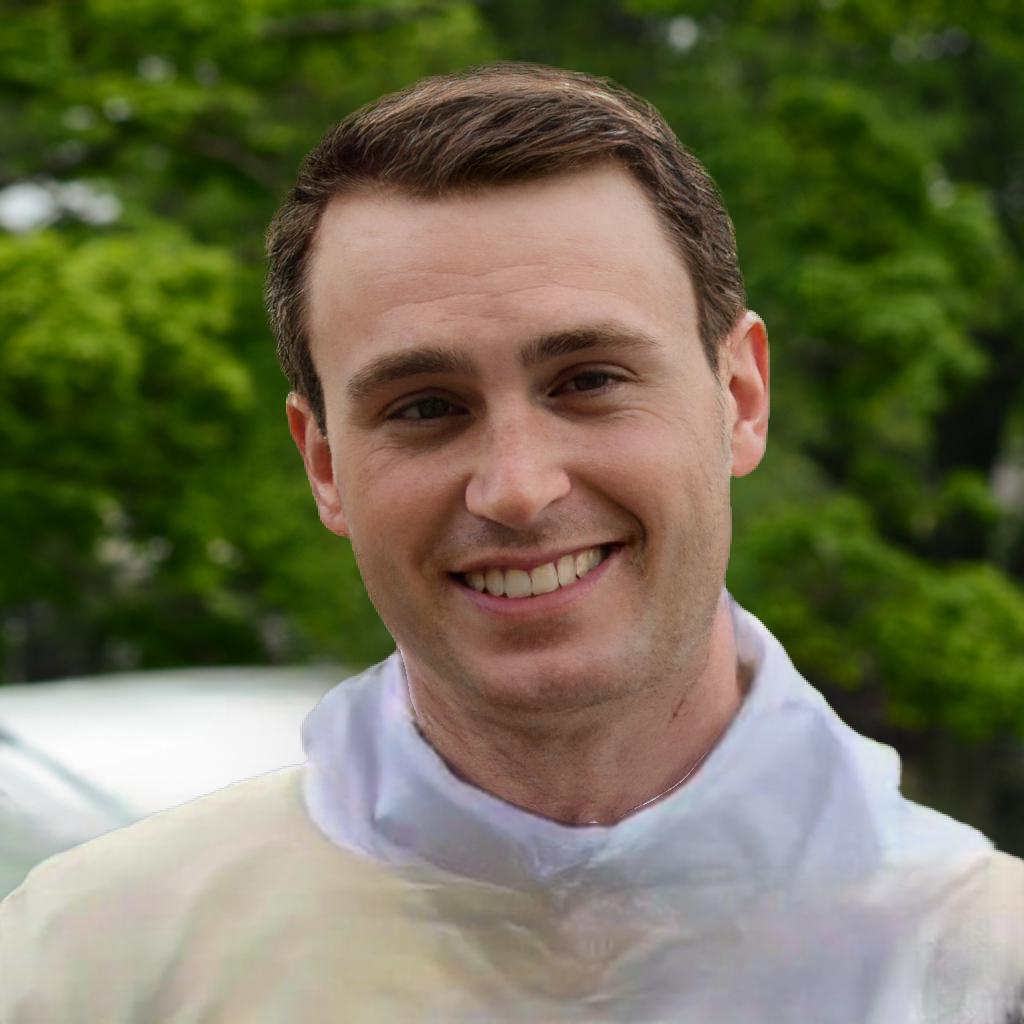} & 
        \hspace{\mrg}
        \includegraphics[width=\wid]{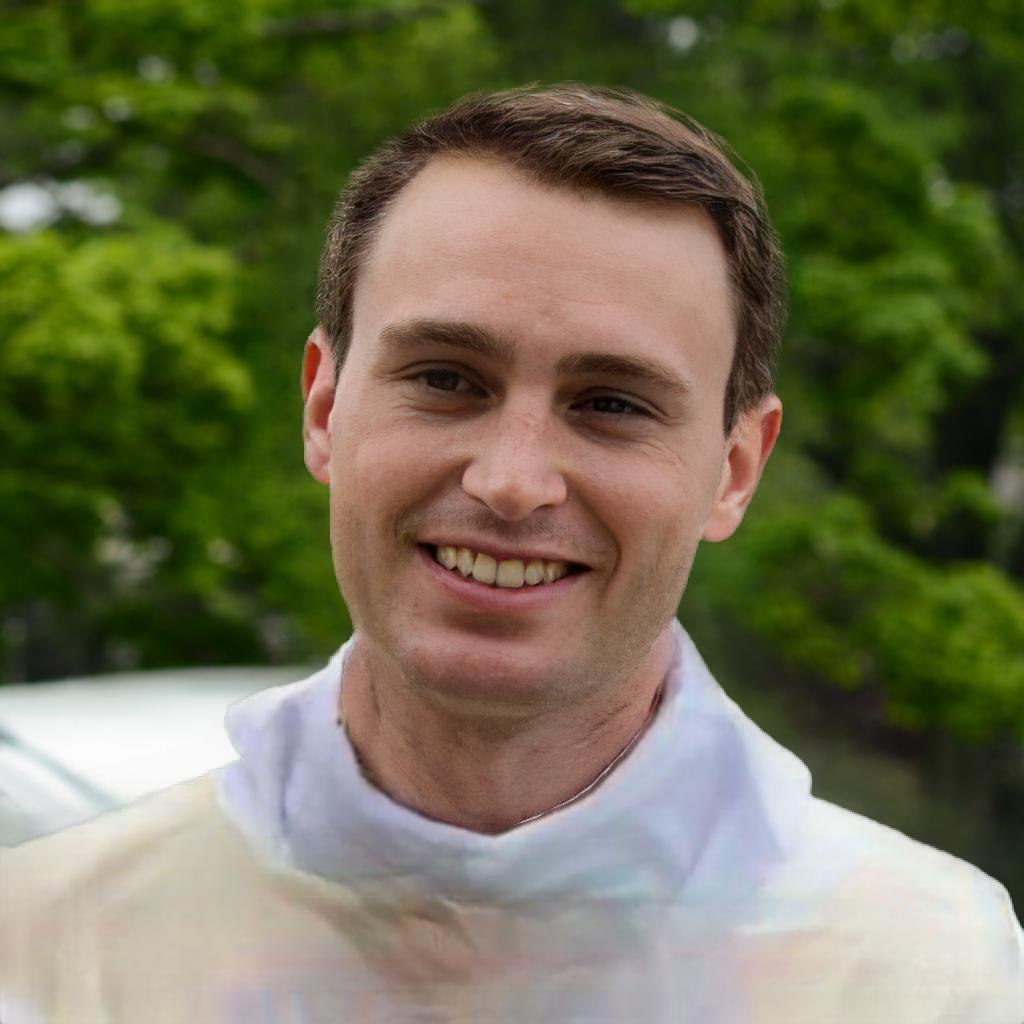} & 
        \hspace{\mrg}
        \includegraphics[width=\wid]{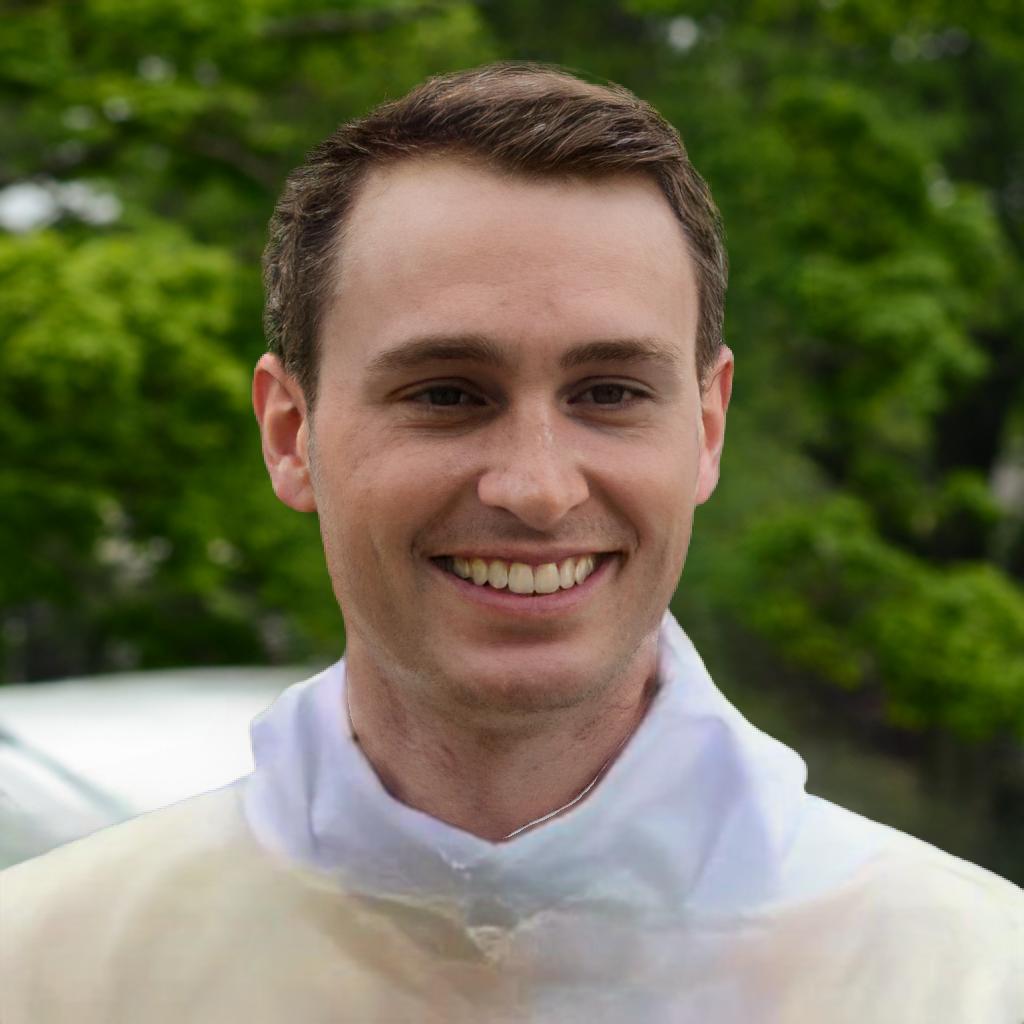} & 
        \hspace{\mrg}
        \includegraphics[width=\wid]{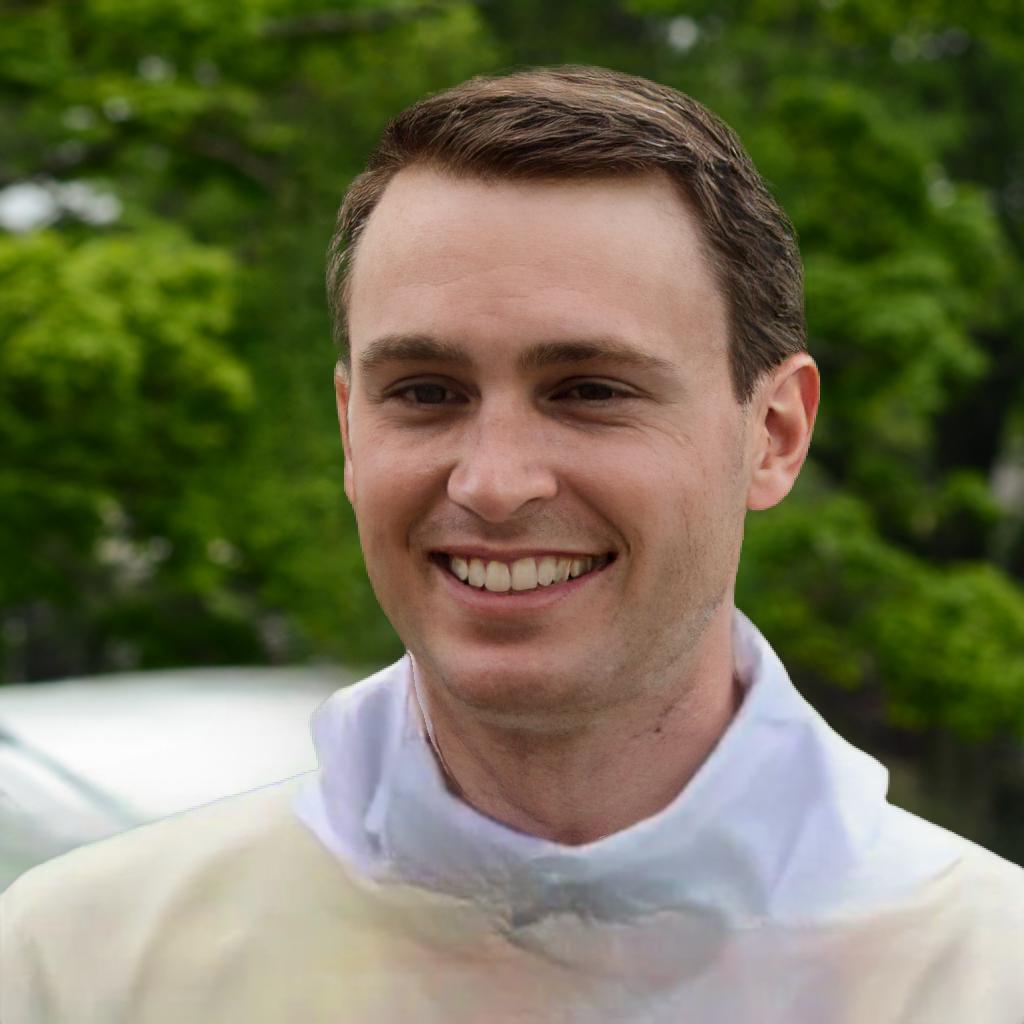} & 
        \hspace{\mrg}
        \includegraphics[width=\wid]{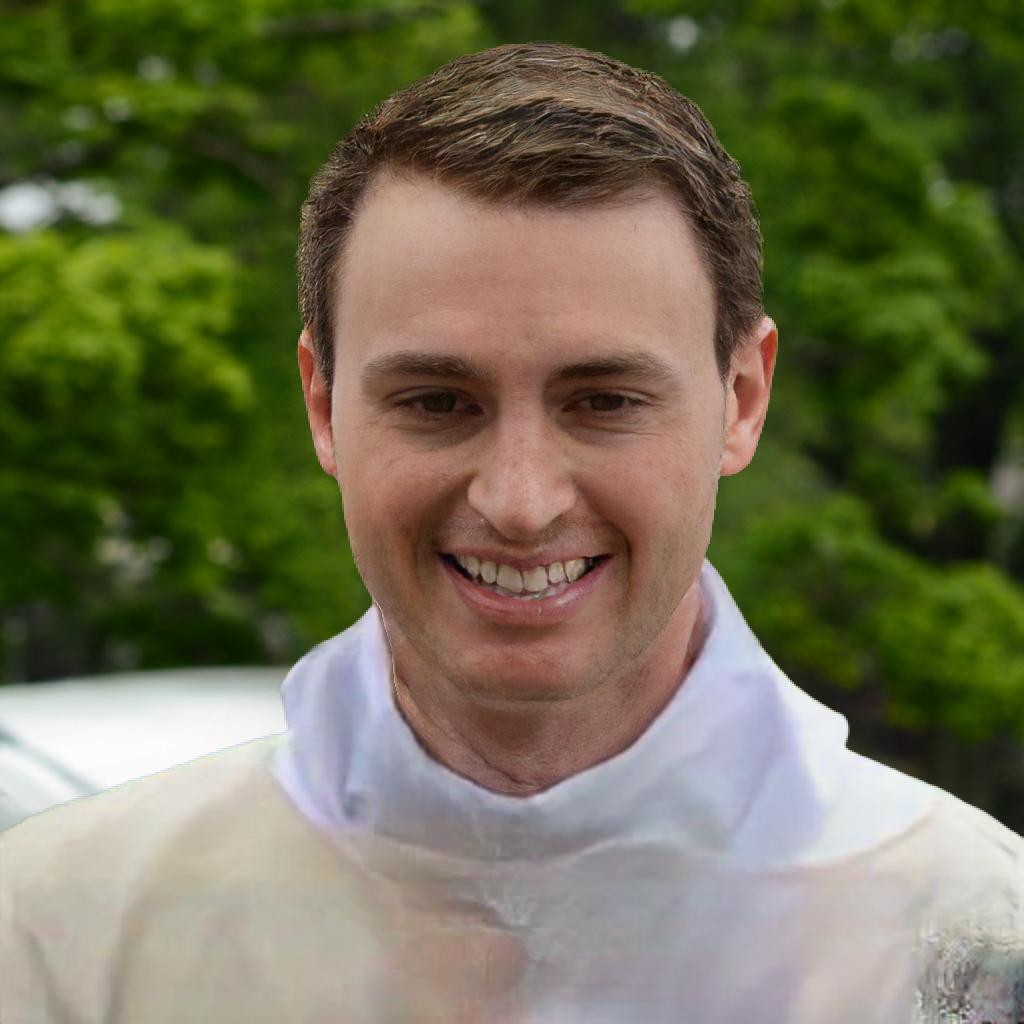} & 
        \hspace{\mrg}
        \includegraphics[width=\wid]{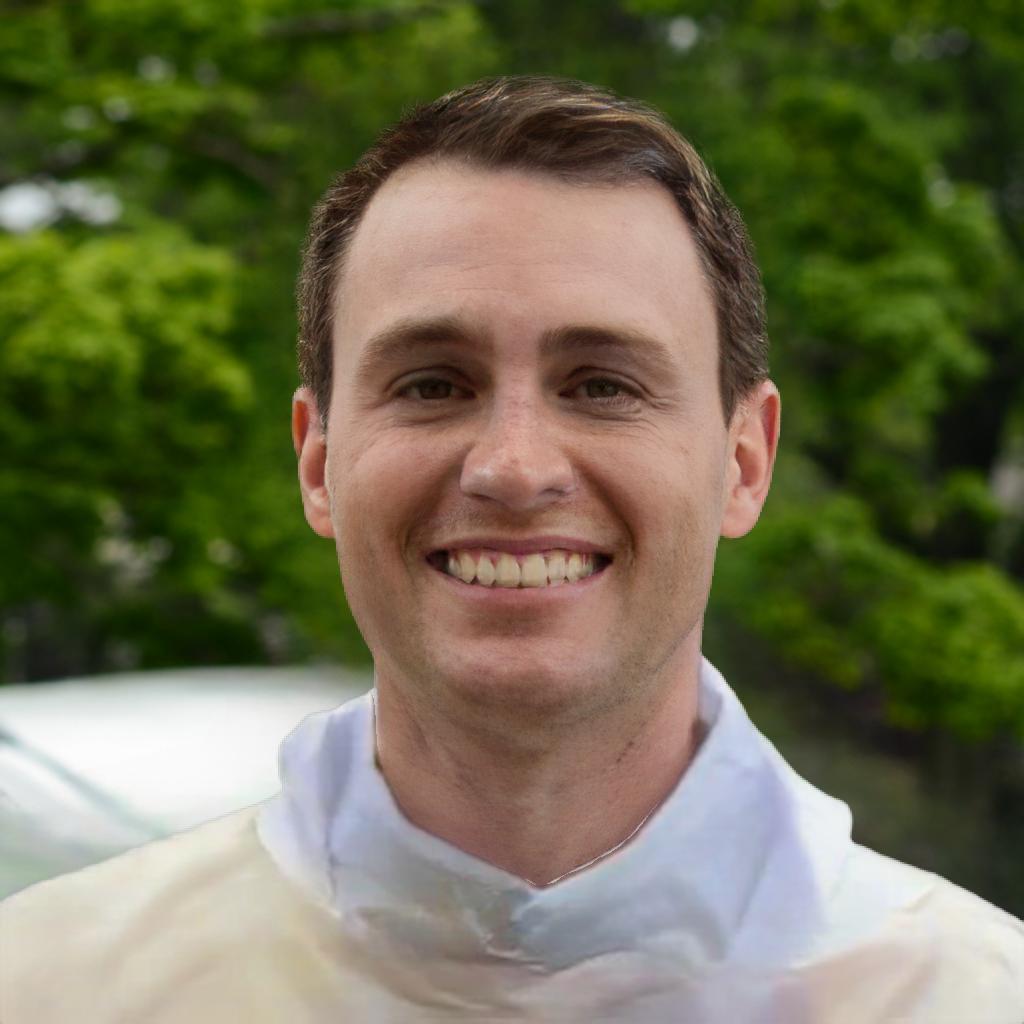}
        \\
        \textbf{Source} & 
        \hspace{\mrg} 
        \textbf{Roll +15\textdegree} & 
        \hspace{\mrg} 
        \textbf{Roll -15\textdegree} & 
        \hspace{\mrg} 
        \textbf{Roll +15\textdegree} & 
        \hspace{\mrg} 
        \textbf{Roll -15\textdegree} & 
        \hspace{\mrg} 
        \textbf{Pitch +15\textdegree} & 
        \hspace{\mrg} 
        \textbf{Pitch -15\textdegree}
    \end{tabular}
    \vspace{-0.4cm}
    \caption{Result of the explicit head rotation.}
    \label{fig:turns}
\end{figure*}

\subsection{Gaze loss}
To get more natural facial appearance, we put into operation  specialized gaze loss based on gaze and blink estimation models. Our model was trained to distill a state-of-the-art gaze detection system (RT-GENE) ~\cite{Fischer2018RTGENERE} and blink estimation model (RT-BENE). ~\cite{Cortacero_2019_ICCV}. We distilled two systems in one model with two heads with the common backbone, one to predict gaze direction and another one to predict blink. First, we infer both models on 60k random frames from our dataset. We did this in order to extract the maximum information from the images of the eyes. As a backbone for our model we used VGG-16 that takes one image of the eye (either left or right) and predict latent vector with size 256, next we sum both vector to get bound representation of eyes. We also derive features from 2D keypoints, for this we use a simple network consists of 3 liner layers with ReLU activations that produce latent vector with size 64. Next, we utilize 2 separate heads, both contain only 2 liners layers with ReLU activations. For the gaze prediction head we use as an input  sum of eye vectors concatenated with keypoint vector and for blink prediction just sum of eye vectors. 

We train this model for 60 epoches with batch size equal to 64. We use Adam optimizer with initial learning rate equal to 0.8e-3, betas=(0.9, 0.999), eps=1e-08, weight decay=0 and one cycle learning rate schedule with steps per epoch equal number of batches in epoch and \text{pct start}=0.1. We use MAE and MSE losses with $w_\text{MAE} = 15$ and $w_\text{MSE} = 10$, we treat predictions from RT-GENE and RT-BENE as ground truth.

\subsection{Explicit control of the pose}
Our system allows some explicit control of a human pose on an output image. First, we can either preserve scale of the source image, that could be utilized in video conference, or use scale and translation ($s \text{\small\&} t$) from the driving image to fully mimic the driver (Figure~\ref{fig:scale}). Despite the fact that we didn't pay any attention to disentangle expression and head rotation, we found that we can both make formalization (Figure~\ref{fig:front}) and apply head rotation from frontal pose on moderate angels, we found that it works at least for 15\textdegree angles (Figure~\ref{fig:turns}).


\section{Additional results}
\begin{figure*}[!ht]
    \centering    
    \setlength{\wid}{0.19\textwidth}
    \setlength{\mrg}{-0.3cm}
    \begin{tabular}{ccccc}
        \includegraphics[width=\wid]{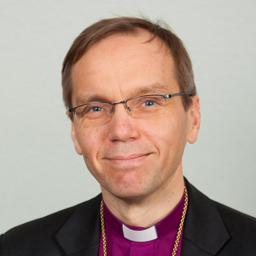} & 
        \hspace{\mrg}
        \includegraphics[width=\wid]{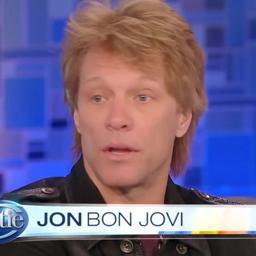} & 
        \hspace{\mrg}
        \includegraphics[width=\wid]{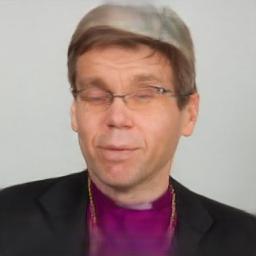} & 
        \hspace{\mrg}
        \includegraphics[width=\wid]{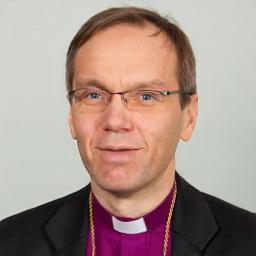} & 
        \hspace{\mrg}
        \includegraphics[width=\wid]{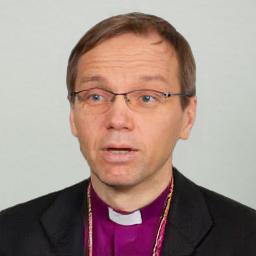} 
        \\ %
        \includegraphics[width=\wid]{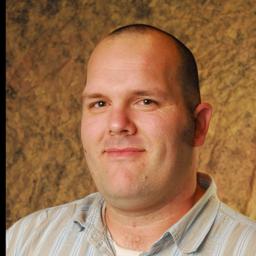} & 
        \hspace{\mrg}
        \includegraphics[width=\wid]{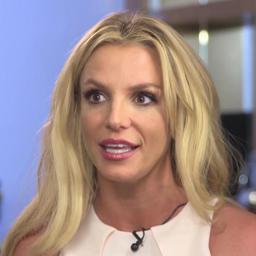} & 
        \hspace{\mrg}
        \includegraphics[width=\wid]{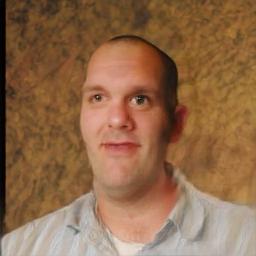} & 
        \hspace{\mrg}
        \includegraphics[width=\wid]{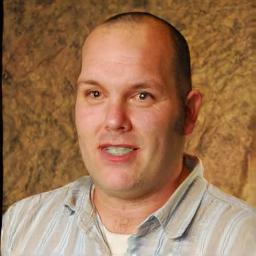} & 
        \hspace{\mrg}
        \includegraphics[width=\wid]{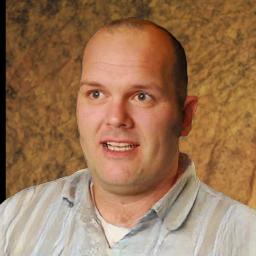} 
        \\ %
        \includegraphics[width=\wid]{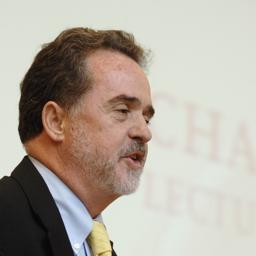} & 
        \hspace{\mrg}
        \includegraphics[width=\wid]{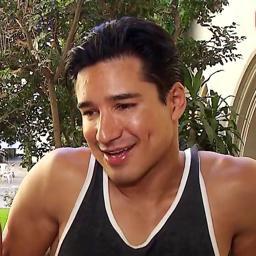} & 
        \hspace{\mrg}
        \includegraphics[width=\wid]{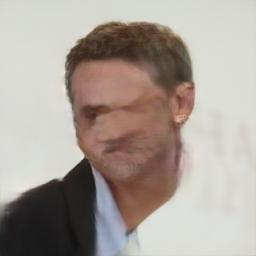} & 
        \hspace{\mrg}
        \includegraphics[width=\wid]{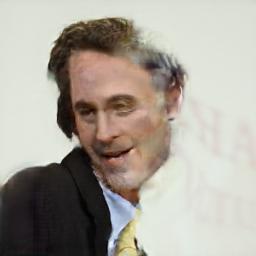} & 
        \hspace{\mrg}
        \includegraphics[width=\wid]{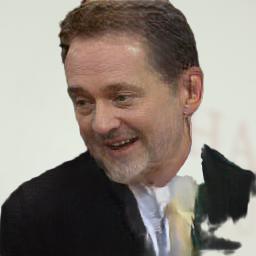} 
        \\ %
        \includegraphics[width=\wid]{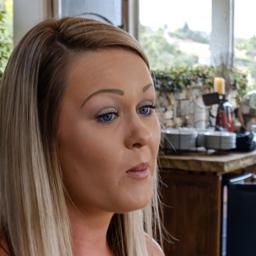} & 
        \hspace{\mrg}
        \includegraphics[width=\wid]{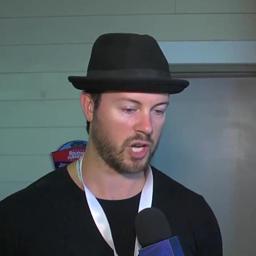} & 
        \hspace{\mrg}
        \includegraphics[width=\wid]{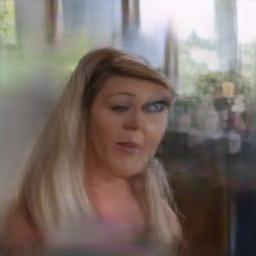} & 
        \hspace{\mrg}
        \includegraphics[width=\wid]{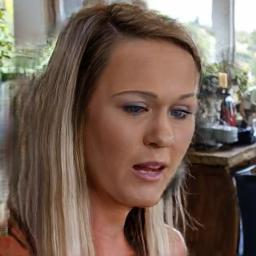} & 
        \hspace{\mrg}
        \includegraphics[width=\wid]{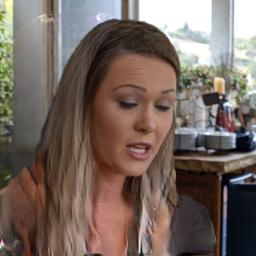} 
        \\
        \includegraphics[width=\wid]{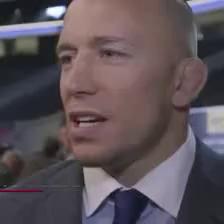} & 
        \hspace{\mrg}
        \includegraphics[width=\wid]{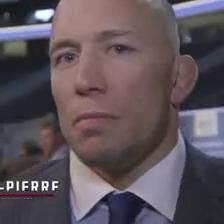} & 
        \hspace{\mrg}
        \includegraphics[width=\wid]{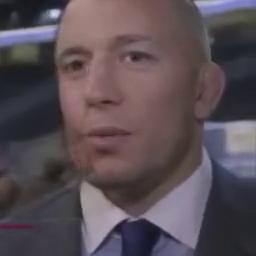} & 
        \hspace{\mrg}
        \includegraphics[width=\wid]{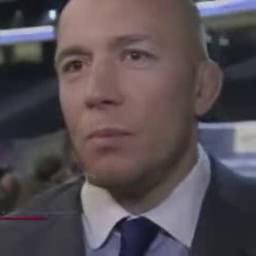} &
        \hspace{\mrg}
        \includegraphics[width=\wid]{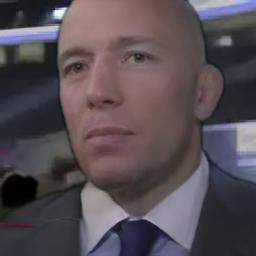} 
        \\
        \includegraphics[width=\wid]{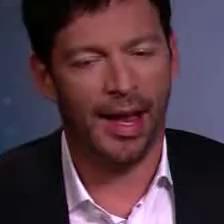} & 
        \hspace{\mrg}
        \includegraphics[width=\wid]{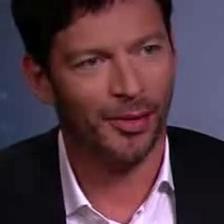} & 
        \hspace{\mrg}
        \includegraphics[width=\wid]{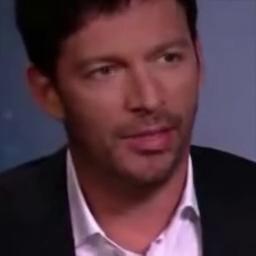} & 
        \hspace{\mrg}
        \includegraphics[width=\wid]{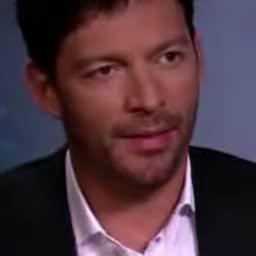} &
        \hspace{\mrg}
        \includegraphics[width=\wid]{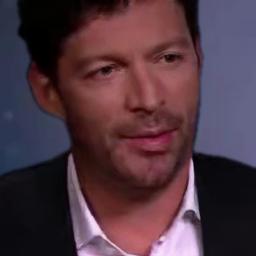} 
        \\
        \textbf{Source} & 
        \hspace{\mrg} 
        \textbf{Driver} & 
        \hspace{\mrg} 
        \textbf{FOMM}~\cite{Siarohin2019FirstOM} & 
        \hspace{\mrg} 
        \textbf{HeadGAN}~\cite{Doukas2021HeadGANON} & 
        \hspace{\mrg} 
        \textbf{Ours}
    \end{tabular}
    \vspace{-0.4cm}
    \caption{A qualitative comparison of head avatar systems in cross-reenactment scenario (top four rows) and self-reenactment scenario (bottom two rows) at $256 \times 256$ resolution.}
    \label{fig:256px_abl}
\end{figure*}

\begin{figure*}
    \centering    
    \setlength{\wid}{0.24\textwidth}
    \setlength{\mrg}{-0.3cm}
    \begin{tabular}{cccc}
        \includegraphics[width=\wid]{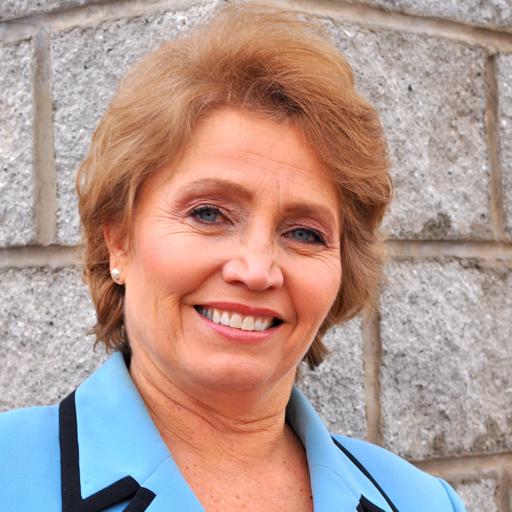} &
        \hspace{\mrg}
        \includegraphics[width=\wid]{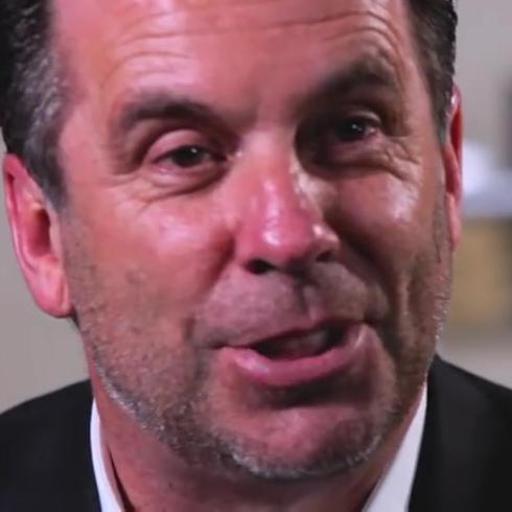} &
        \hspace{\mrg}
        \includegraphics[width=\wid]{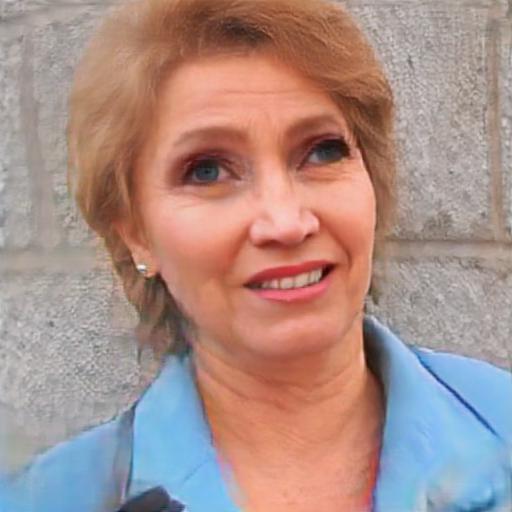} & 
        \hspace{\mrg}
        \includegraphics[width=\wid]{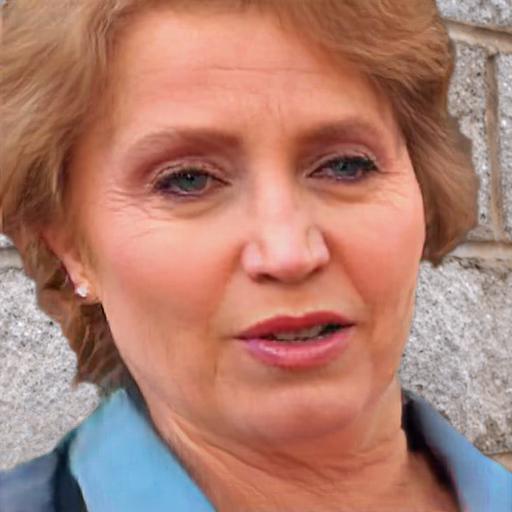}
        \\ %
        \includegraphics[width=\wid]{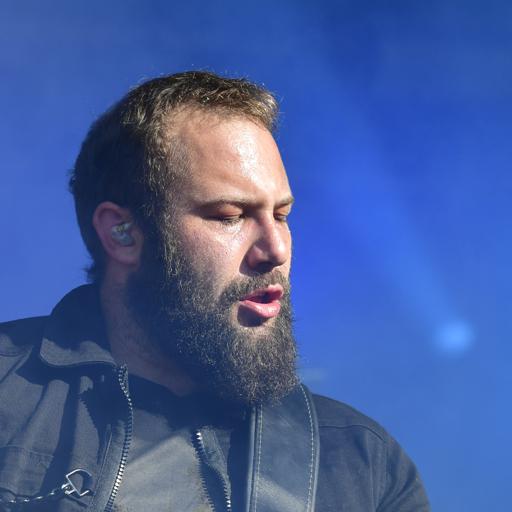} &
        \hspace{\mrg}
        \includegraphics[width=\wid]{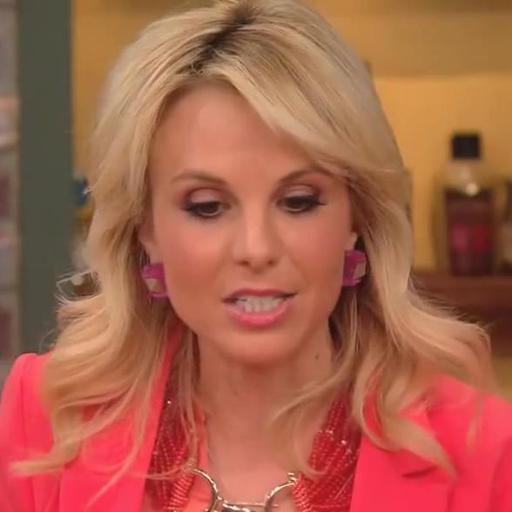} &
        \hspace{\mrg}
        \includegraphics[width=\wid]{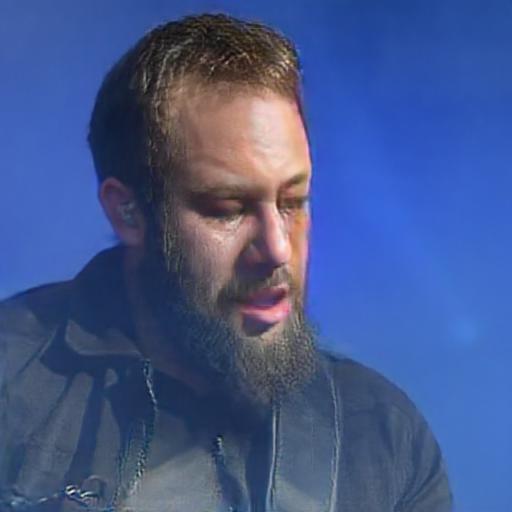} & 
        \hspace{\mrg}
        \includegraphics[width=\wid]{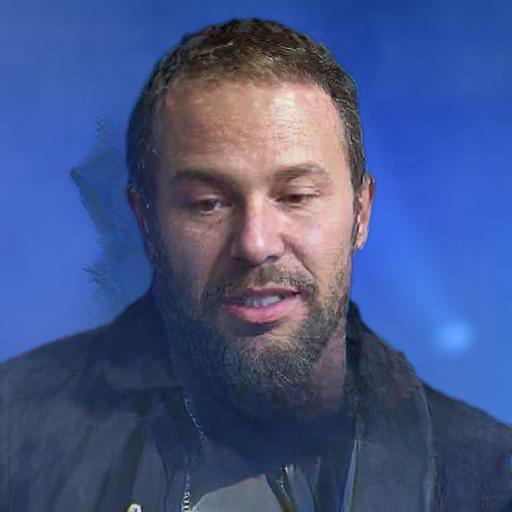}
        \\ %
        \includegraphics[width=\wid]{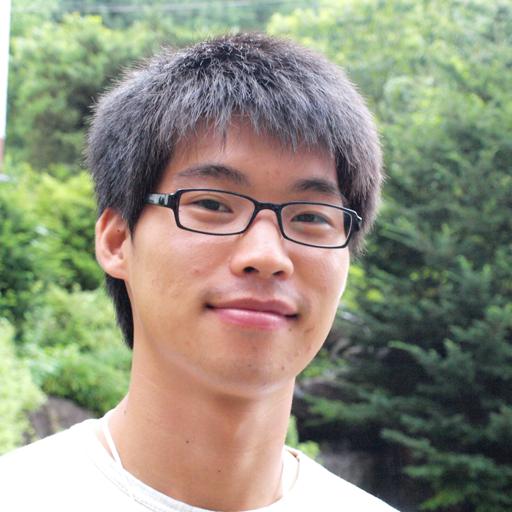} &
        \hspace{\mrg}
        \includegraphics[width=\wid]{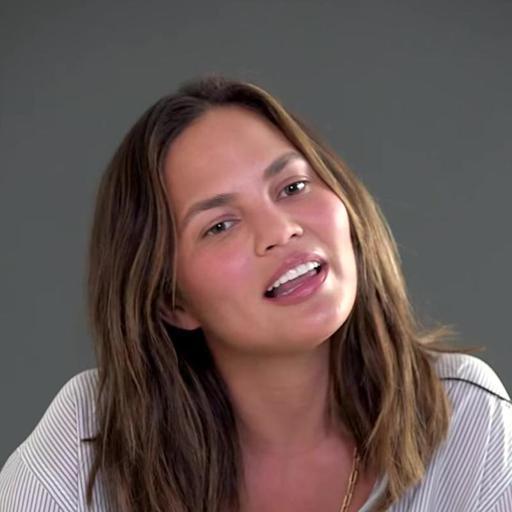} &
        \hspace{\mrg}
        \includegraphics[width=\wid]{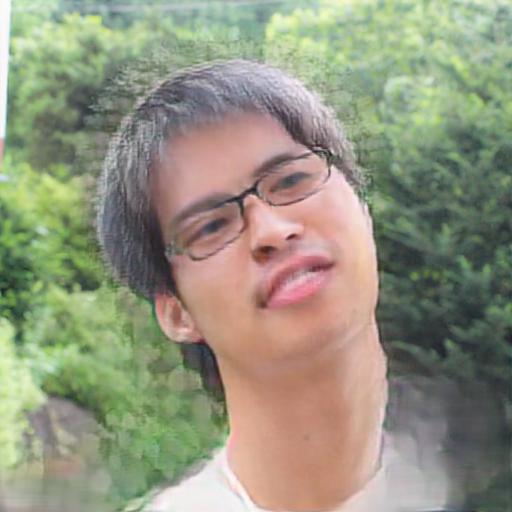} & 
        \hspace{\mrg}
        \includegraphics[width=\wid]{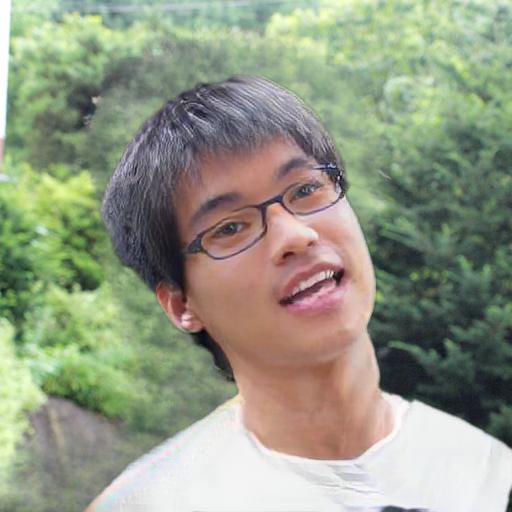}
        \\
        \includegraphics[width=\wid]{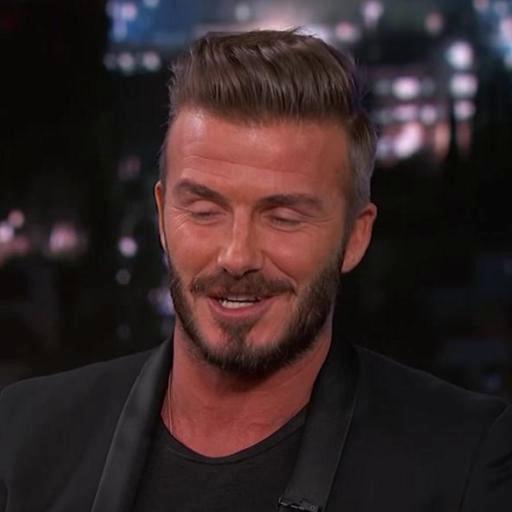} & 
        \hspace{\mrg}
        \includegraphics[width=\wid]{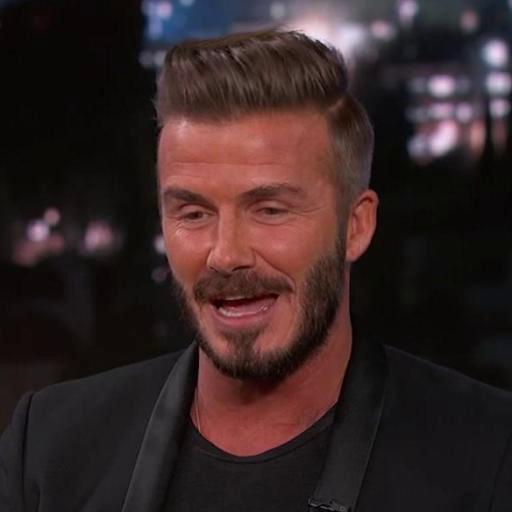} & 
        \hspace{\mrg}
        \includegraphics[width=\wid]{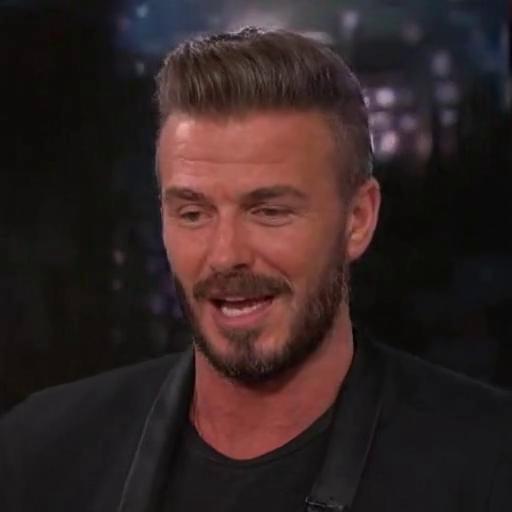} & 
        \hspace{\mrg}
        \includegraphics[width=\wid]{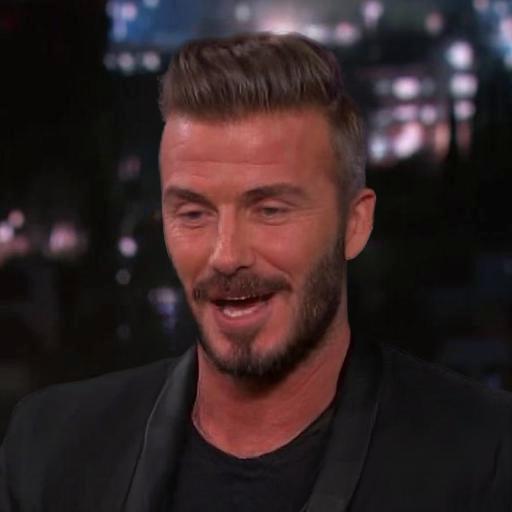}
        
        \\
        \includegraphics[width=\wid]{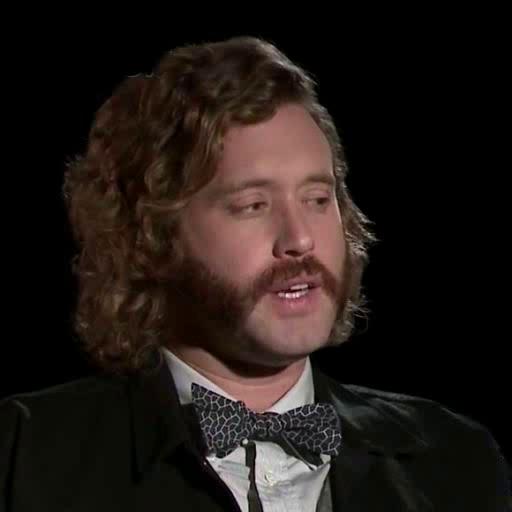} & 
        \hspace{\mrg}
        \includegraphics[width=\wid]{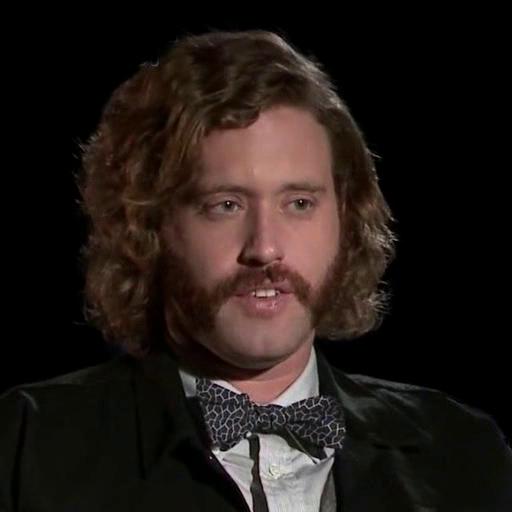} & 
        \hspace{\mrg}
        \includegraphics[width=\wid]{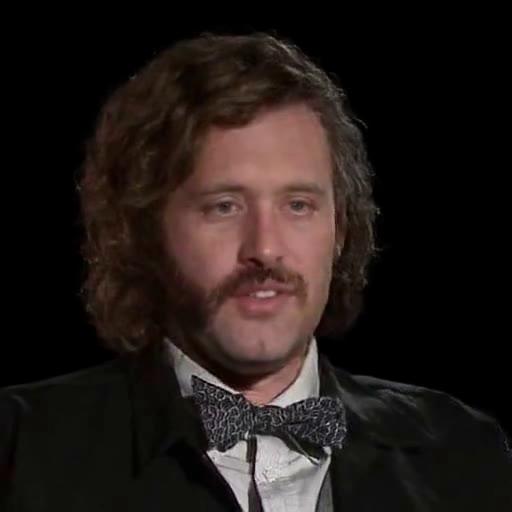} & 
        \hspace{\mrg}
        \includegraphics[width=\wid]{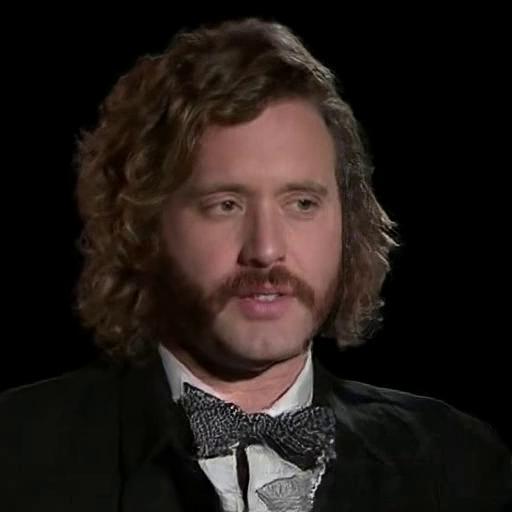}
        \\ 

        \textbf{Source} & 
        \hspace{\mrg} 
        \textbf{Driver} & 
        \hspace{\mrg} 
        \textbf{Face-V2V}~\cite{Wang2021OneShotFN} & 
        \hspace{\mrg} 
        \textbf{Ours} 
    \end{tabular}
    \vspace{-0.4cm}
    \caption{A qualitative comparison of head avatar systems in cross-reenactment scenario (top three rows) and self-reenactment scenario (bottom two rows) at 512px resolution.}
    \label{fig:512px_abl}
\end{figure*}

\begin{figure*}
    \centering    
    \setlength{\wid}{0.19\textwidth}
    \setlength{\mrg}{-0.3cm}
    \setlength{\mrgv}{0cm}
    \begin{tabular}{ccccc}
        \includegraphics[width=\wid]{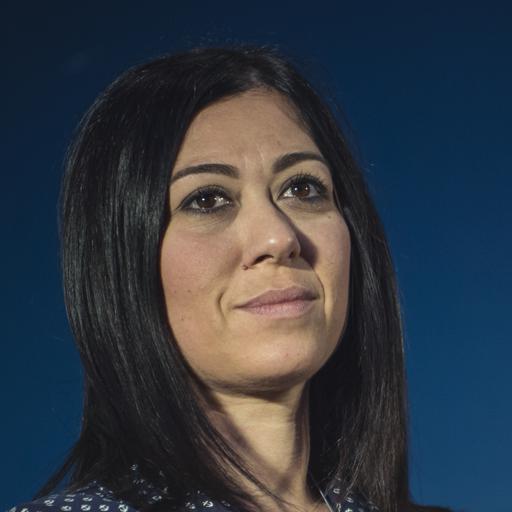} & 
        \hspace{\mrg}
        \vspace{\mrgv}
        \includegraphics[width=\wid]{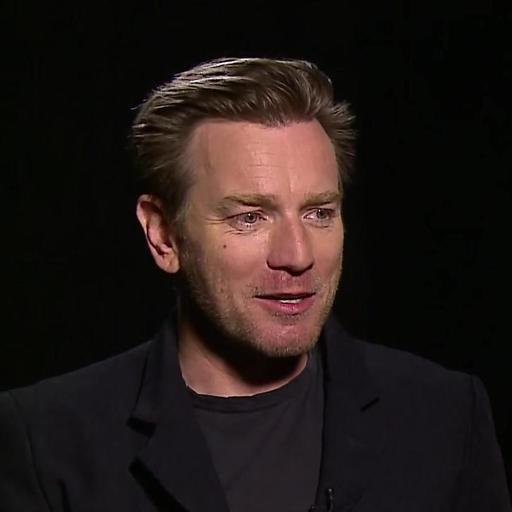} & 
        \hspace{\mrg}
        \includegraphics[width=\wid]{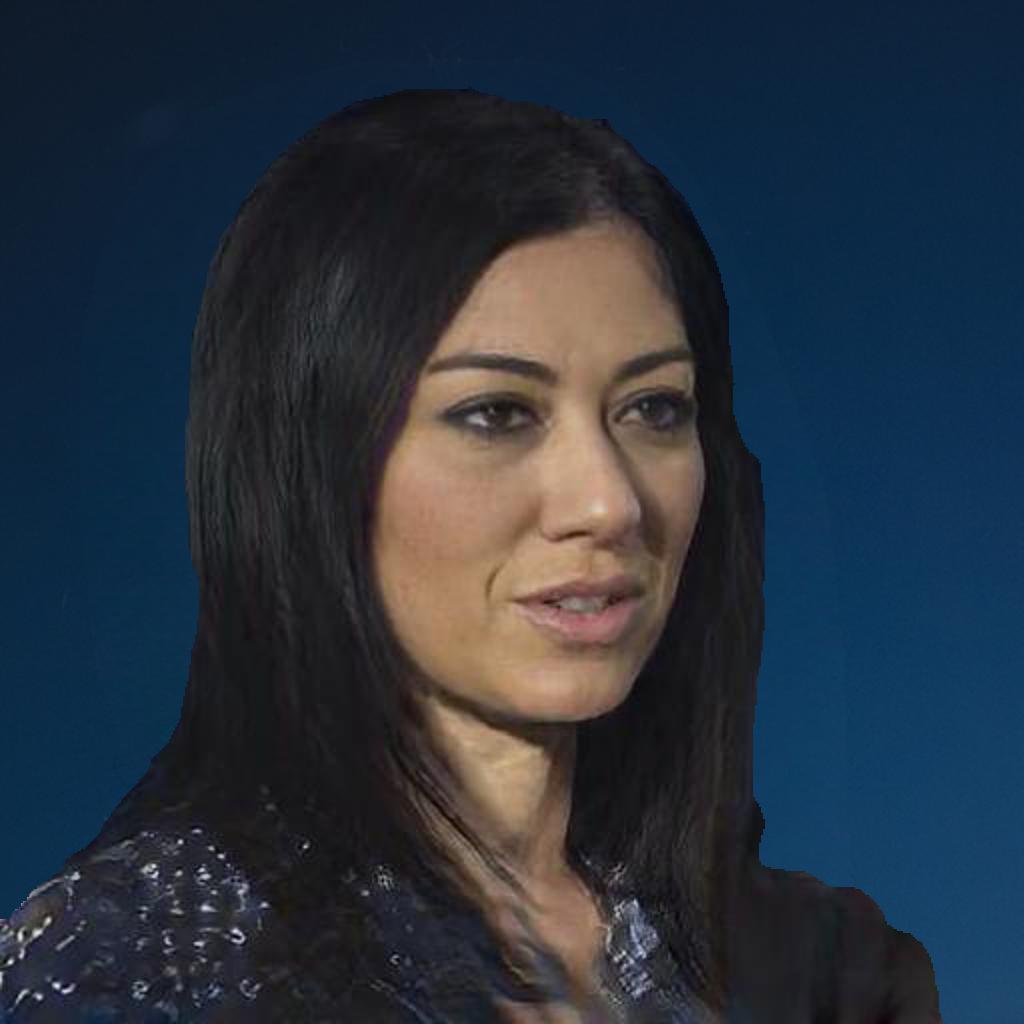} & 
        \hspace{\mrg}
        \includegraphics[width=\wid]{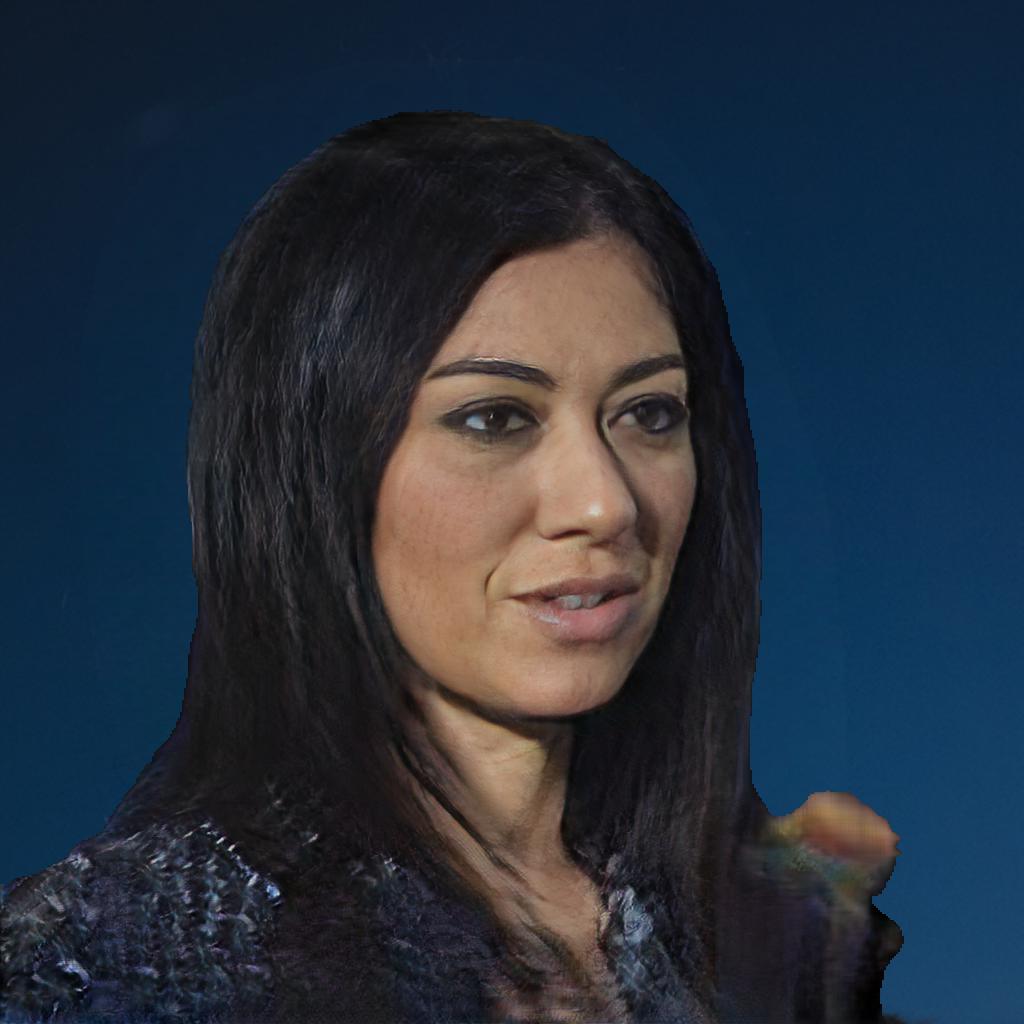} & 
        \hspace{\mrg}
        \includegraphics[width=\wid]{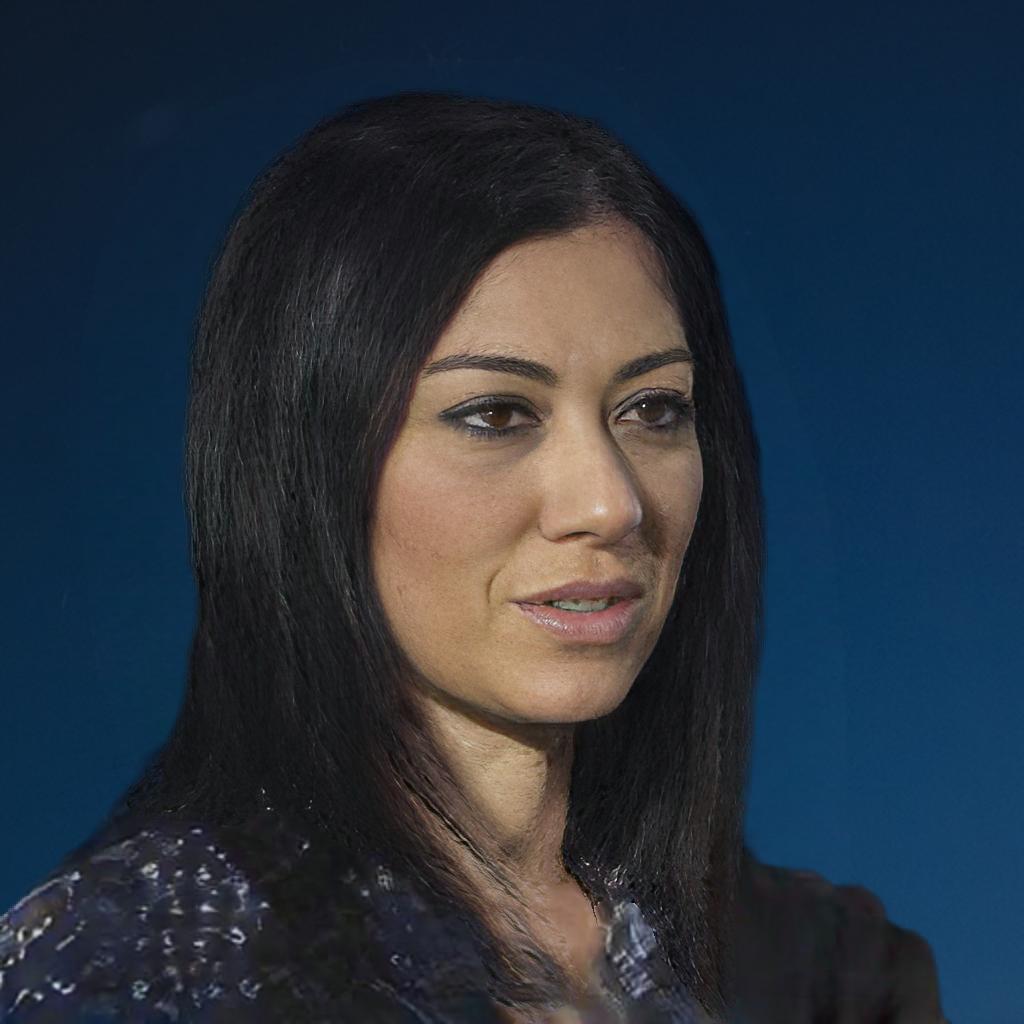} 
        \\ %
        \includegraphics[width=\wid]{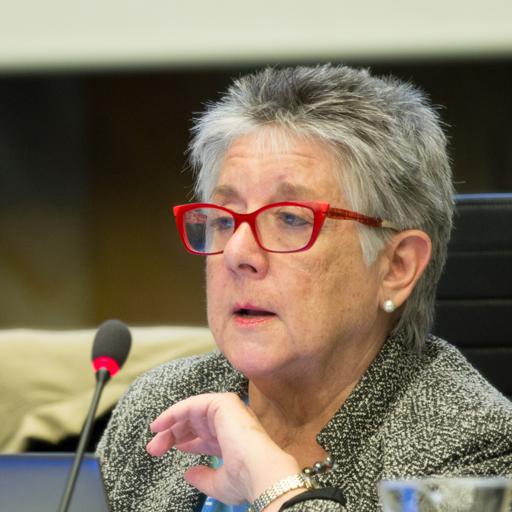} & 
        \hspace{\mrg}
        \vspace{\mrgv}
        \includegraphics[width=\wid]{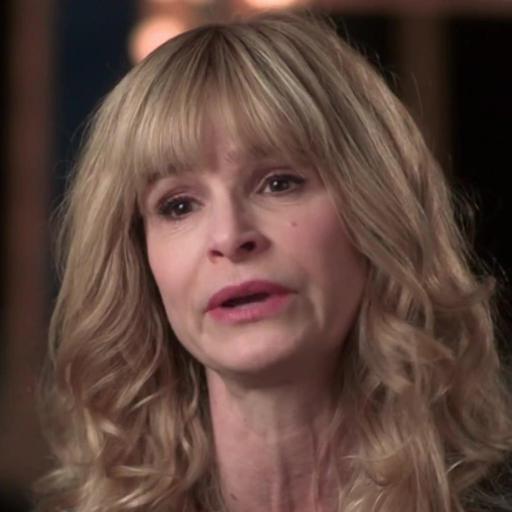} & 
        \hspace{\mrg}
        \includegraphics[width=\wid]{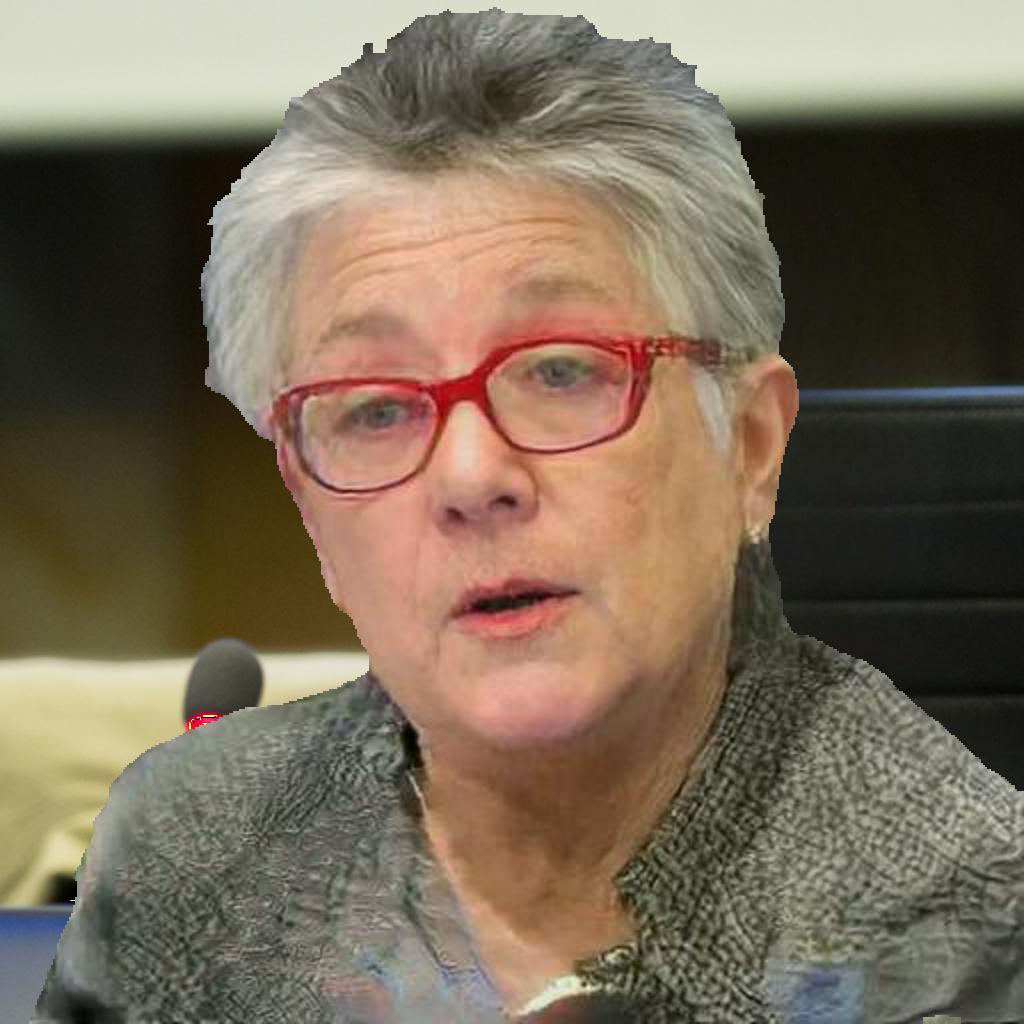} & 
        \hspace{\mrg}
        \includegraphics[width=\wid]{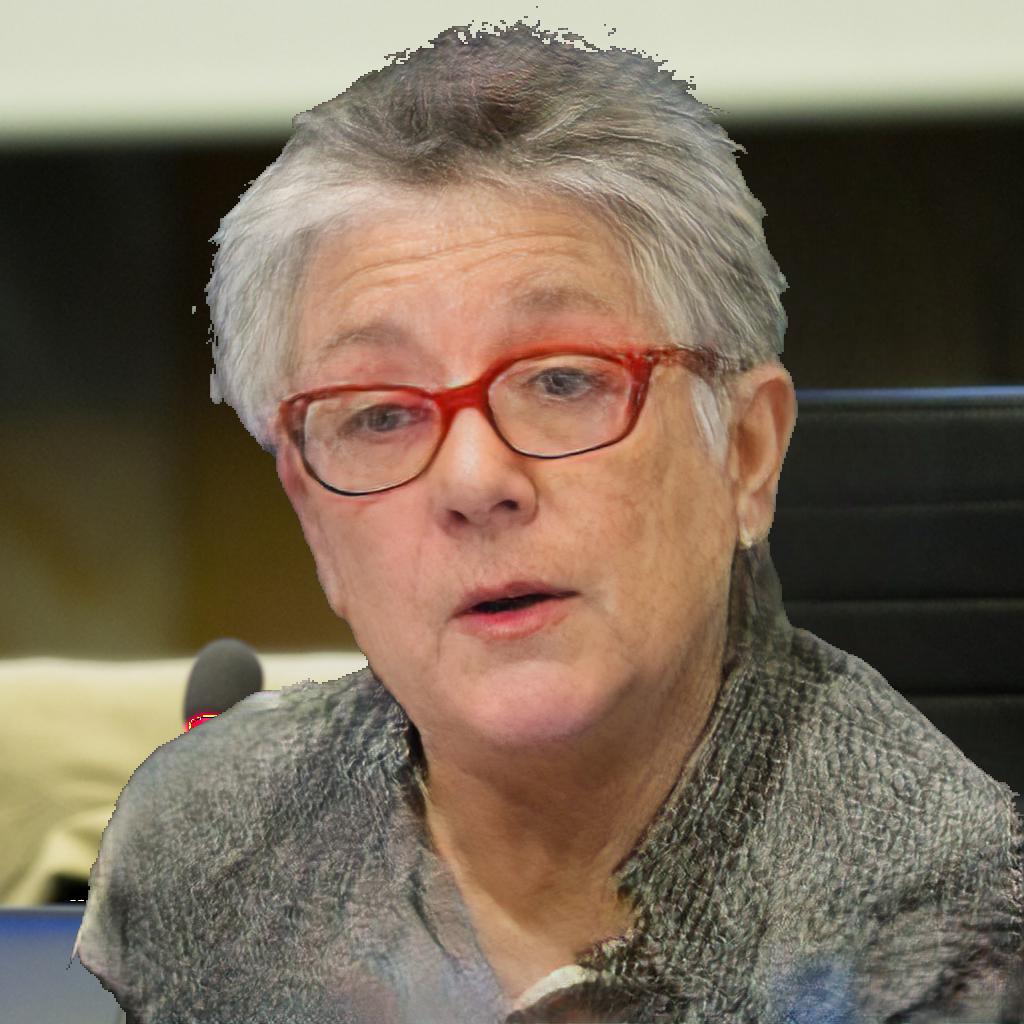} & 
        \hspace{\mrg}
        \includegraphics[width=\wid]{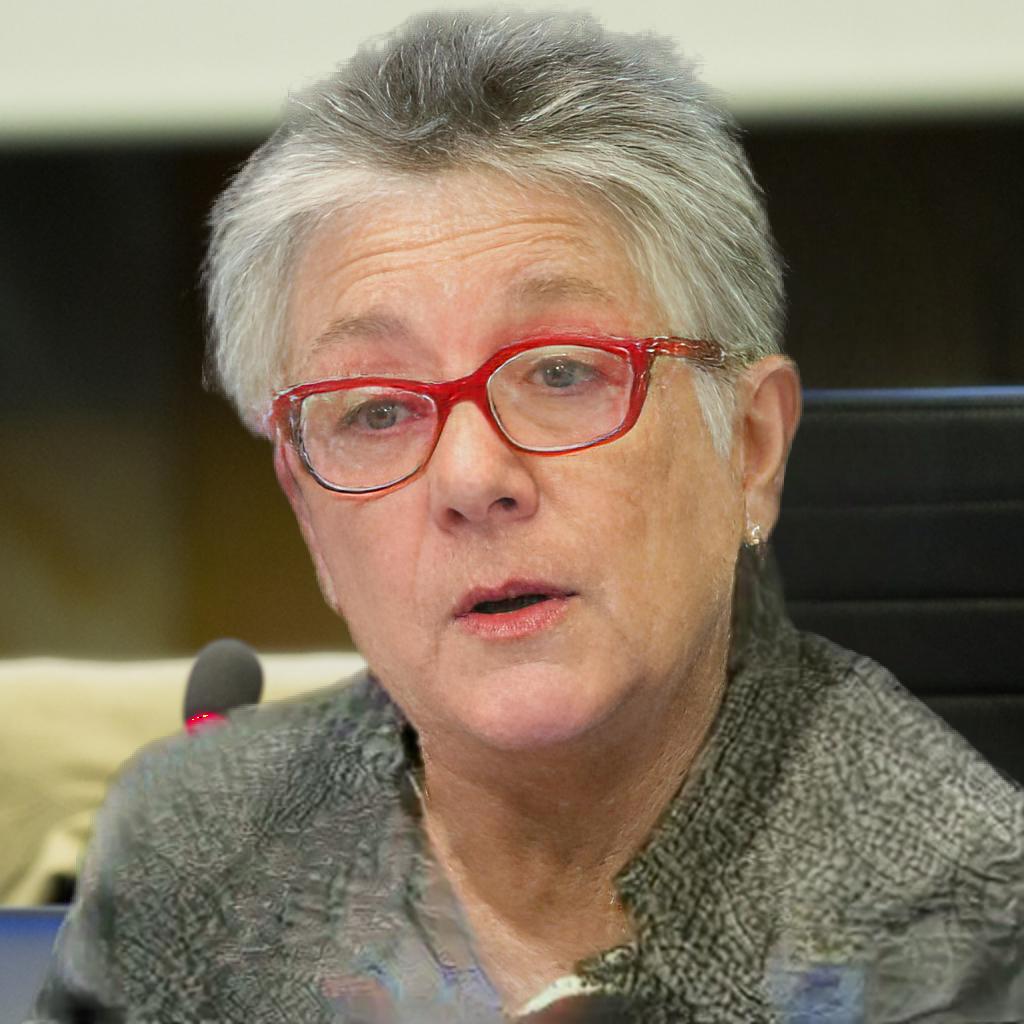} 
        \\ %
        \includegraphics[width=\wid]{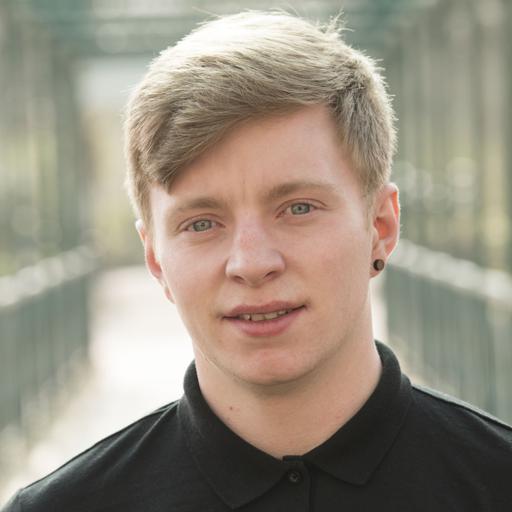} & 
        \hspace{\mrg}
        \vspace{\mrgv}
        \includegraphics[width=\wid]{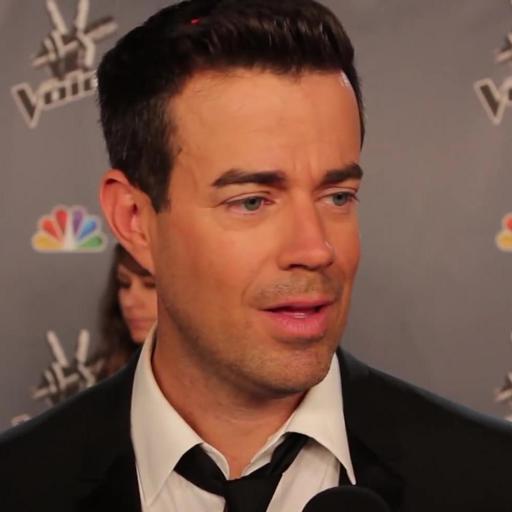} & 
        \hspace{\mrg}
        \includegraphics[width=\wid]{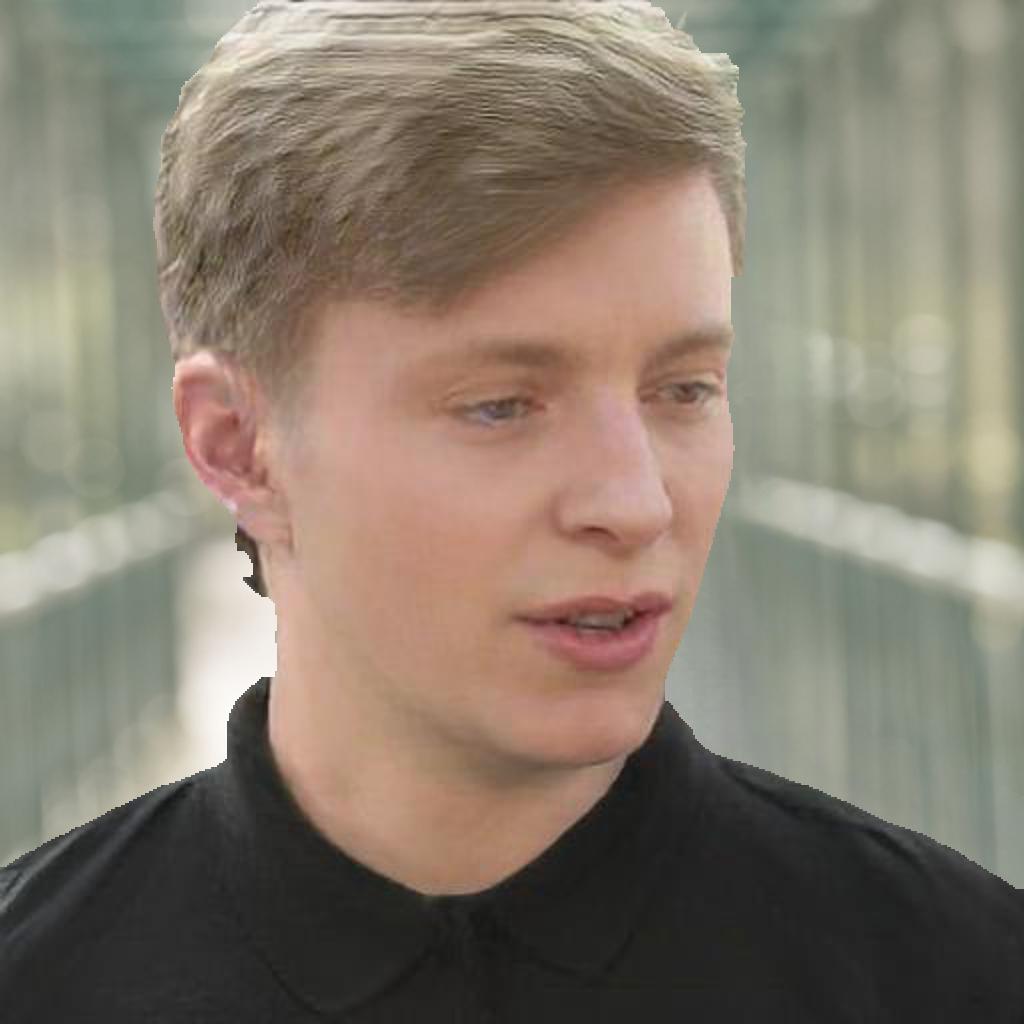} & 
        \hspace{\mrg}
        \includegraphics[width=\wid]{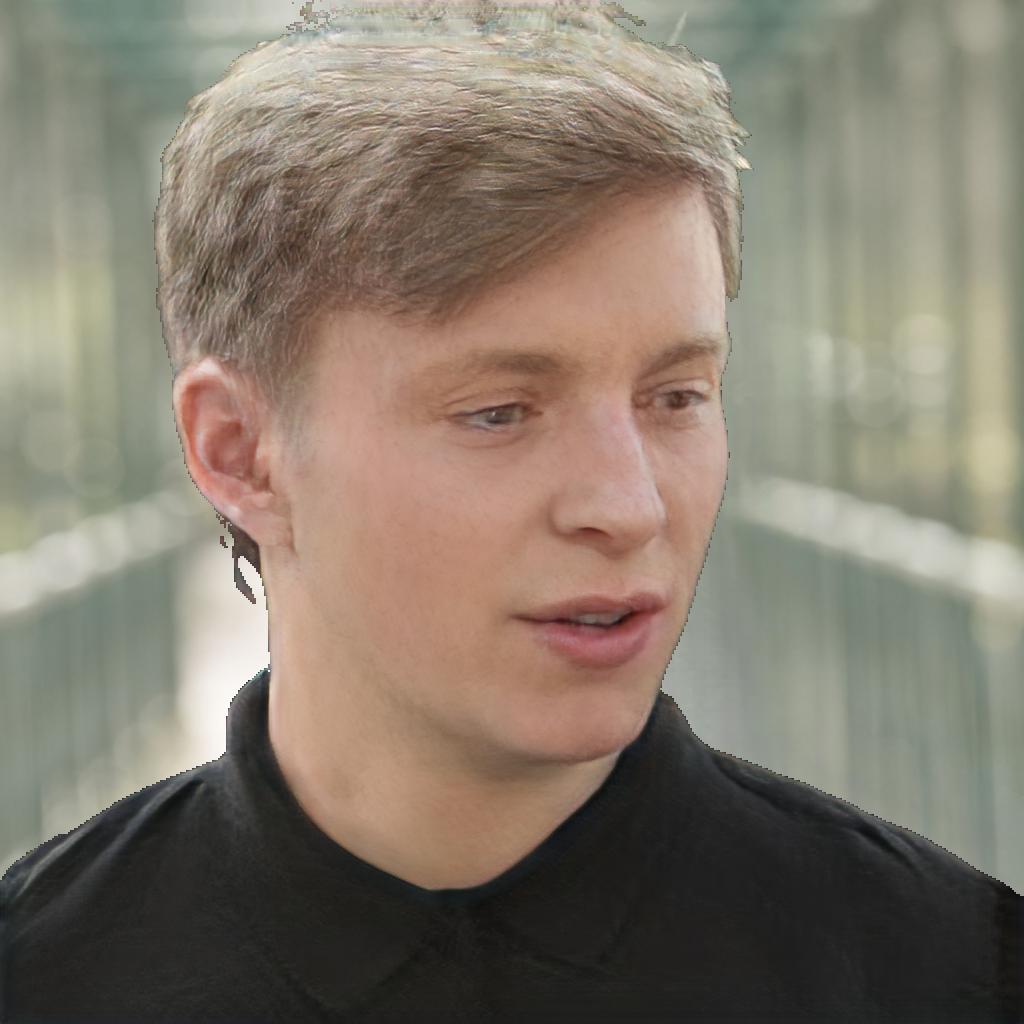} & 
        \hspace{\mrg}
        \includegraphics[width=\wid]{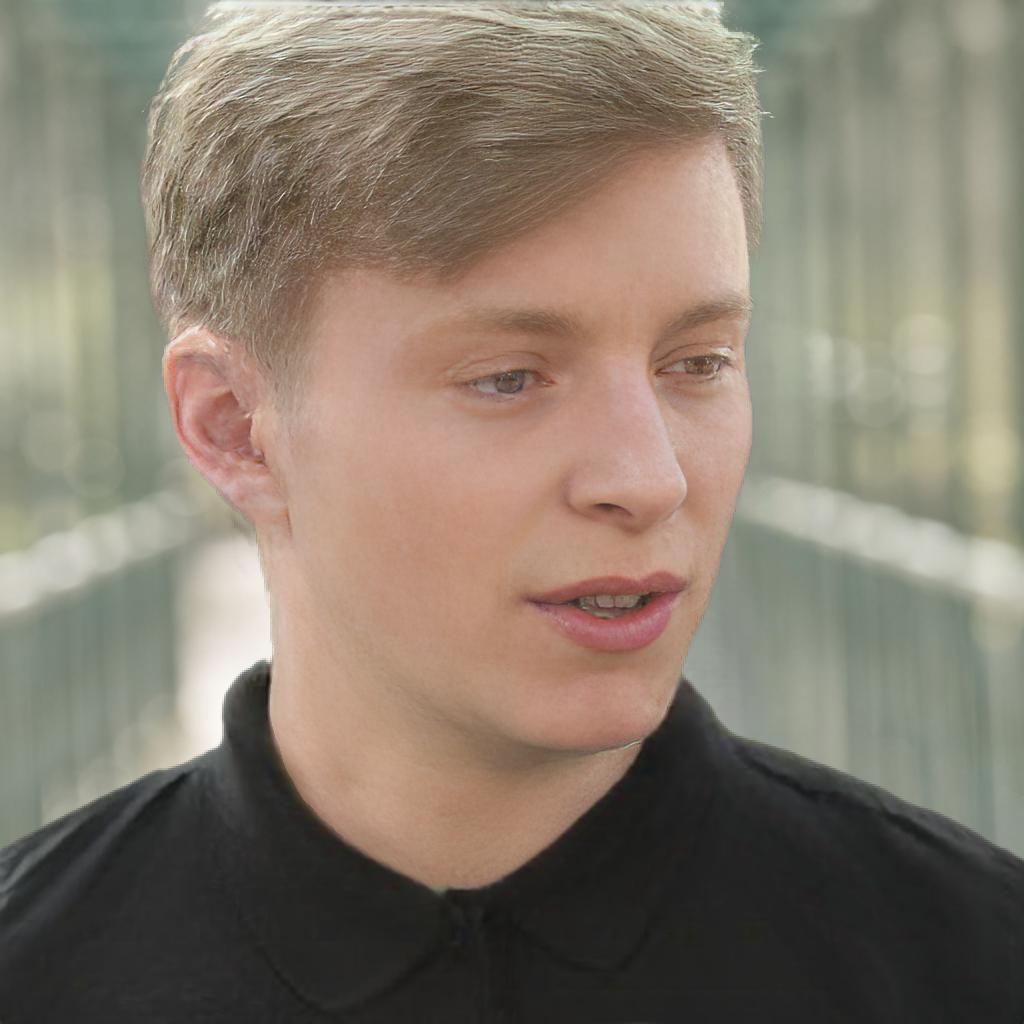} 
        \\ %
        \includegraphics[width=\wid]{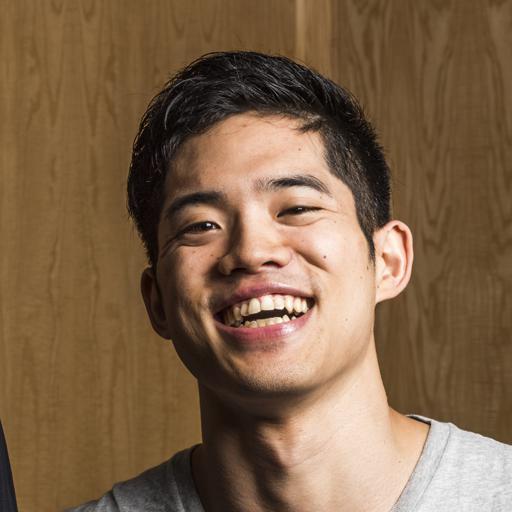} & 
        \hspace{\mrg}
        \vspace{\mrgv}
        \includegraphics[width=\wid]{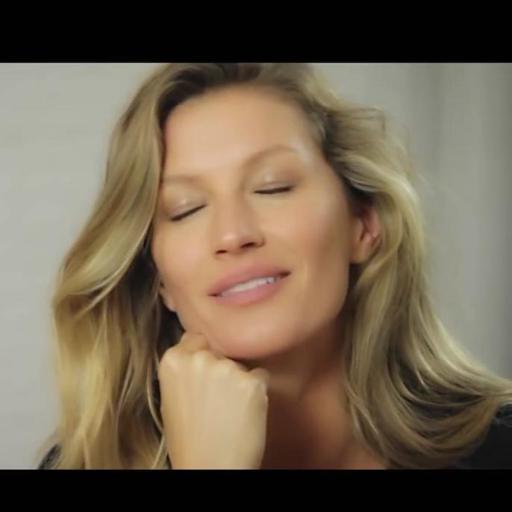} & 
        \hspace{\mrg}
        \includegraphics[width=\wid]{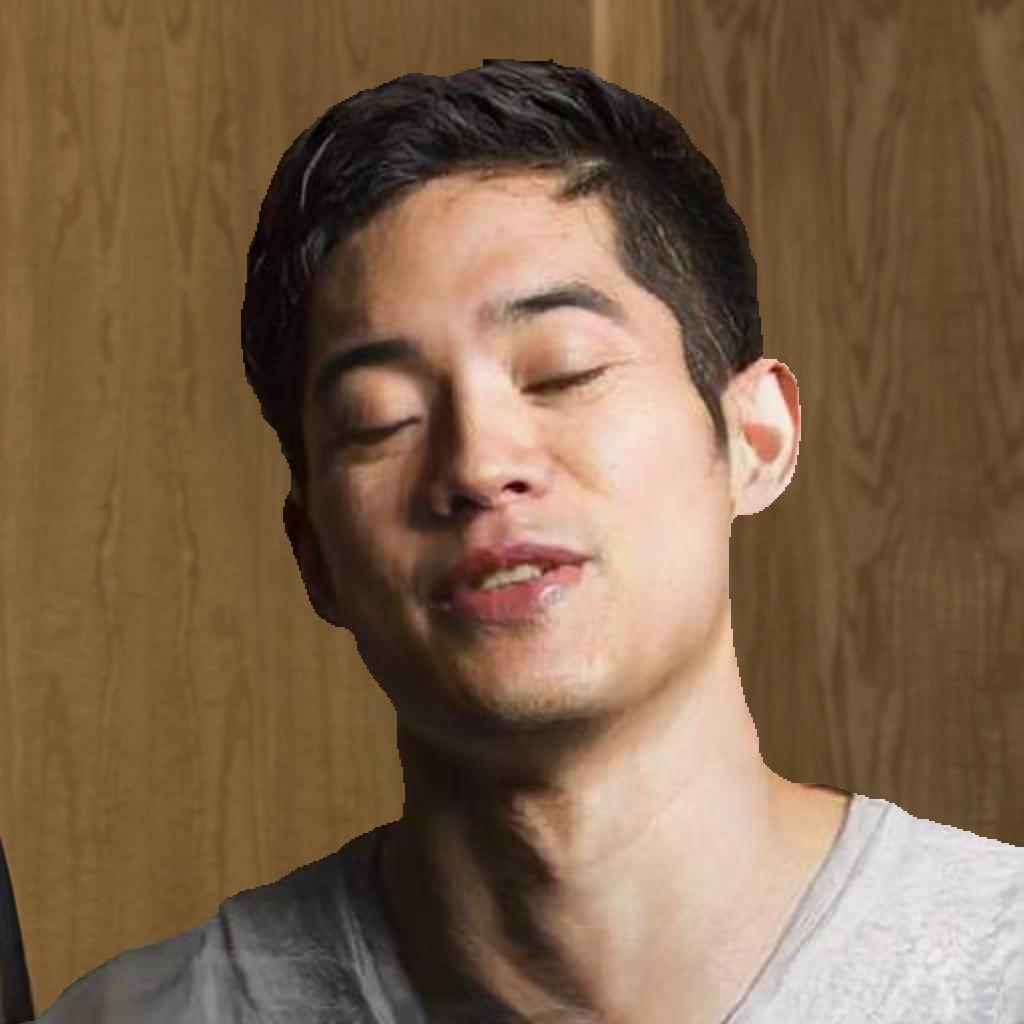} & 
        \hspace{\mrg}
        \includegraphics[width=\wid]{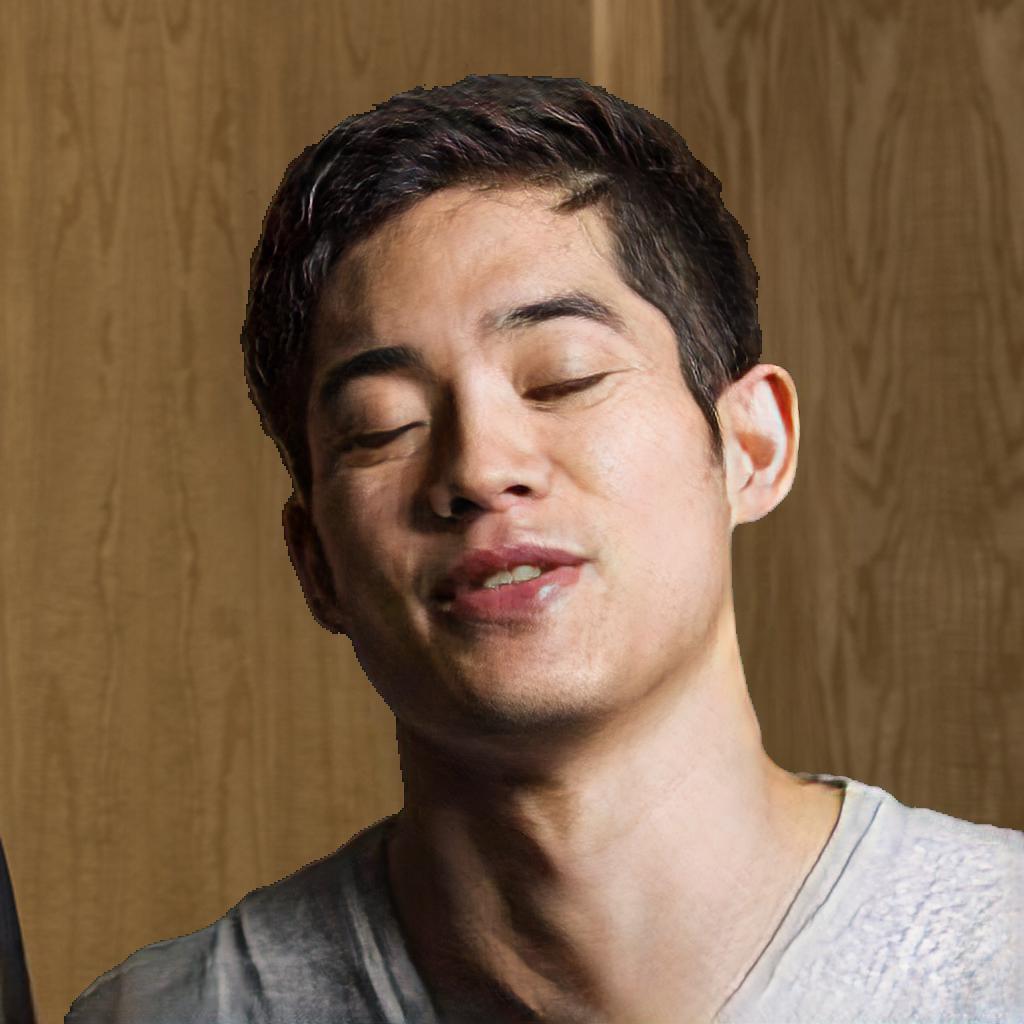} & 
        \hspace{\mrg}
        \includegraphics[width=\wid]{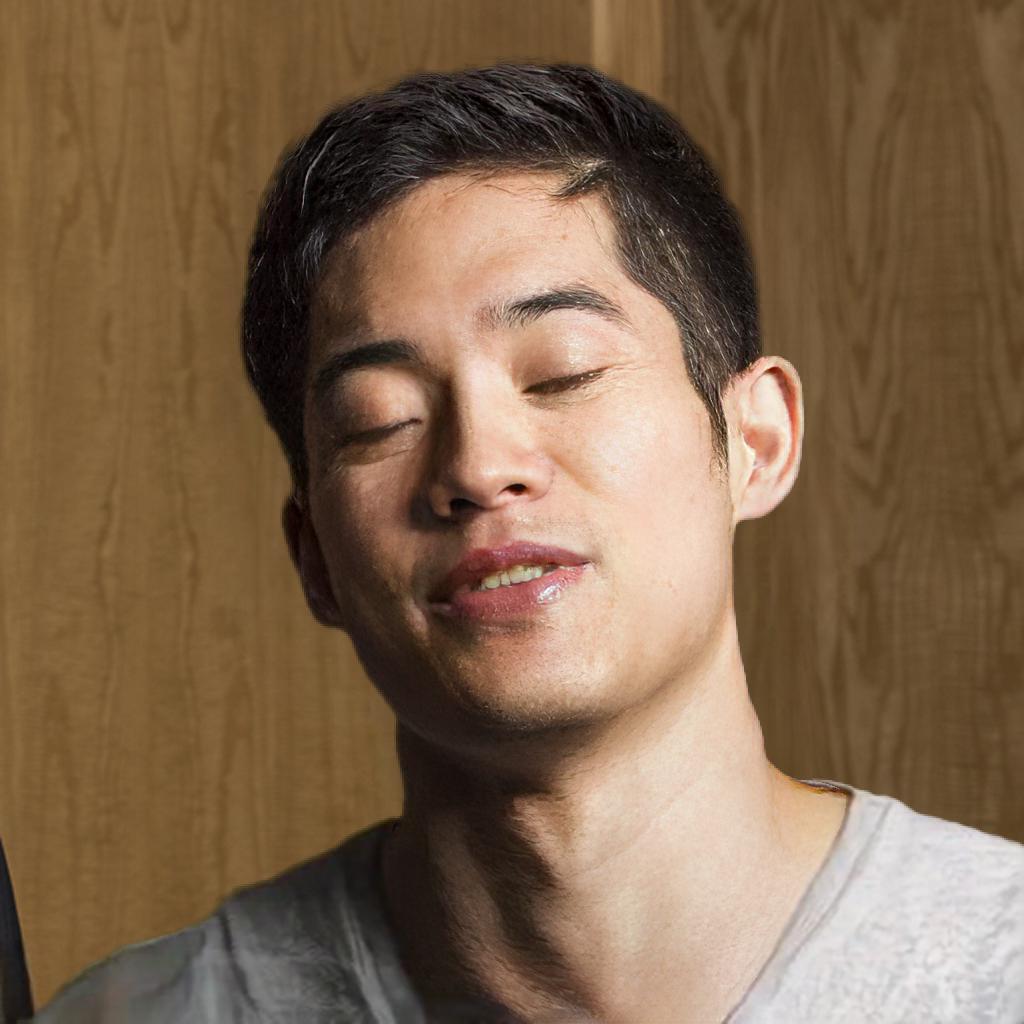} 
        \\ %
        \includegraphics[width=\wid]{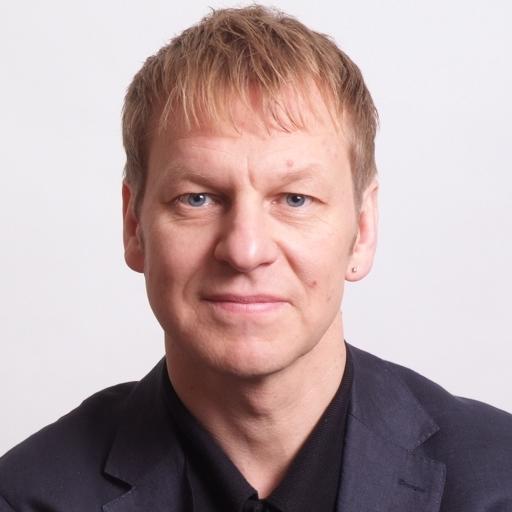} & 
        \hspace{\mrg}
        \vspace{\mrgv}
        \includegraphics[width=\wid]{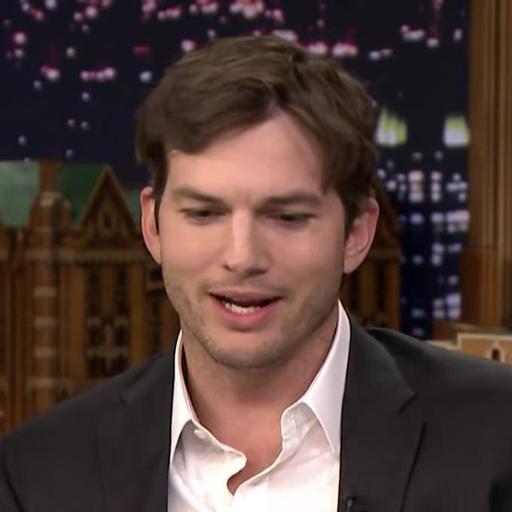} & 
        \hspace{\mrg}
        \includegraphics[width=\wid]{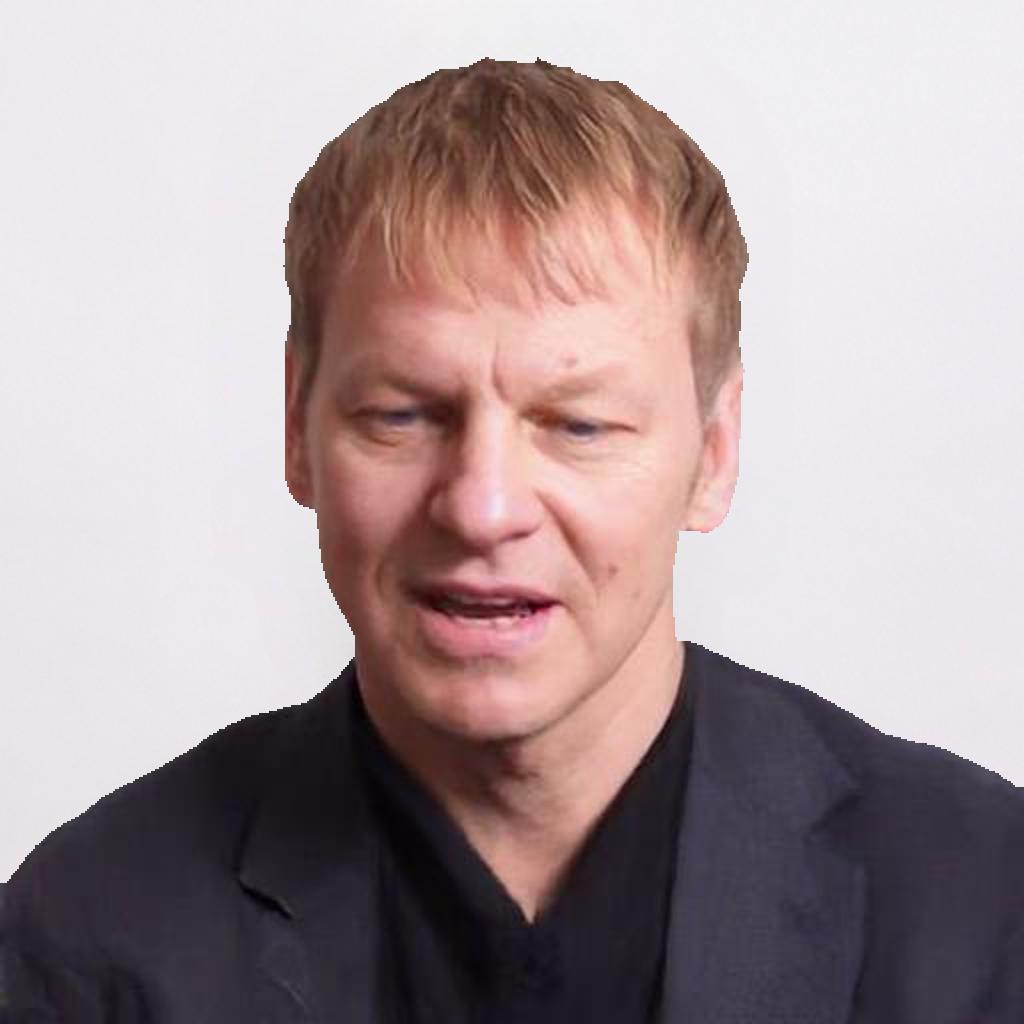} & 
        \hspace{\mrg}
        \includegraphics[width=\wid]{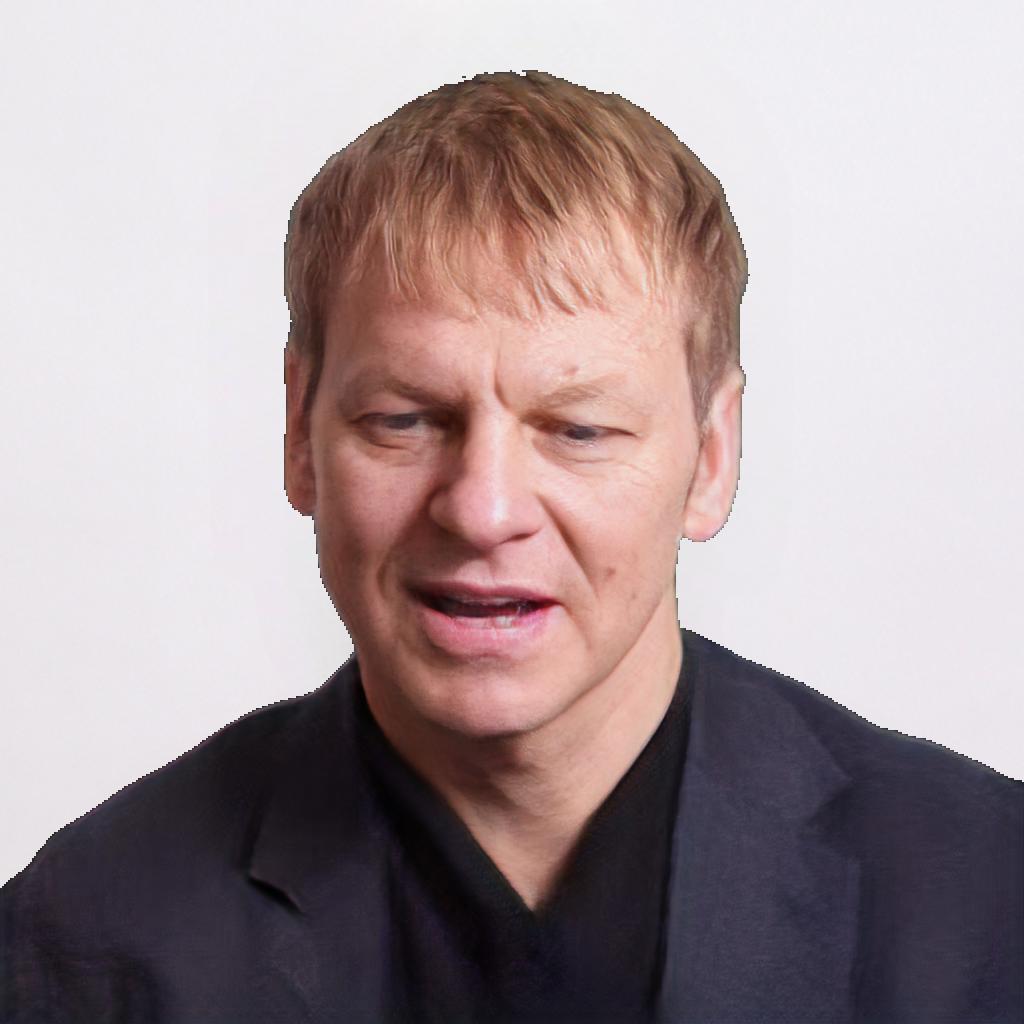} & 
        \hspace{\mrg}
        \includegraphics[width=\wid]{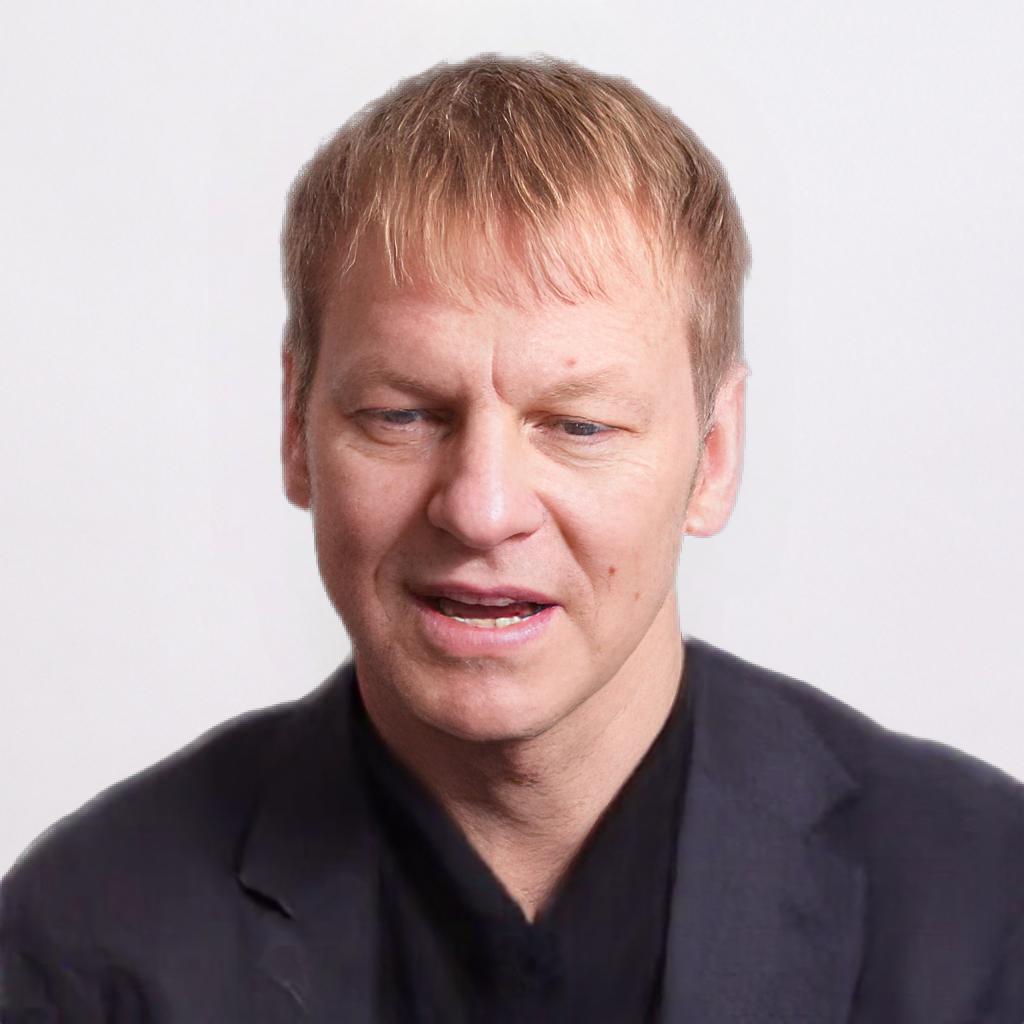} 
        \\ %
        \includegraphics[width=\wid]{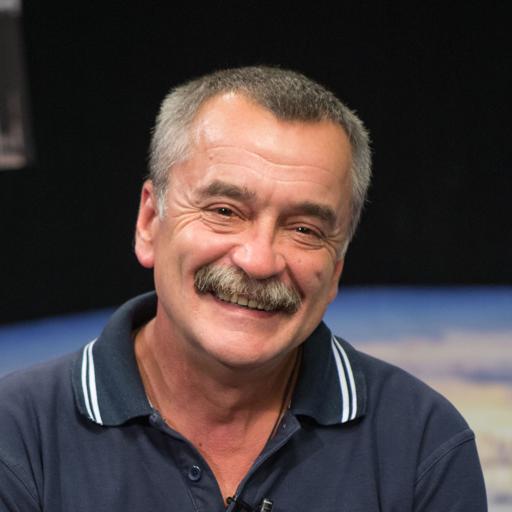} & 
        \hspace{\mrg}
        \vspace{\mrgv}
        \includegraphics[width=\wid]{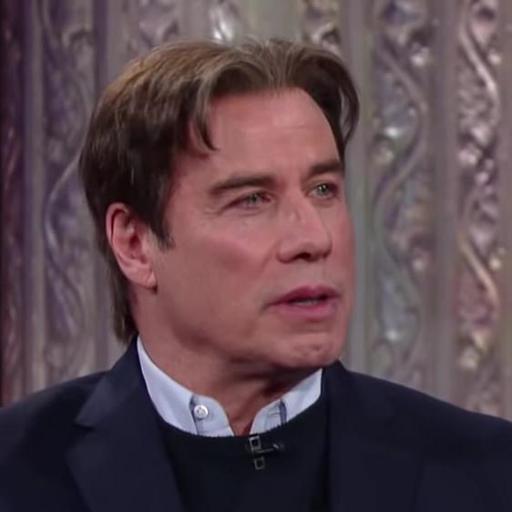} & 
        \hspace{\mrg}
        \includegraphics[width=\wid]{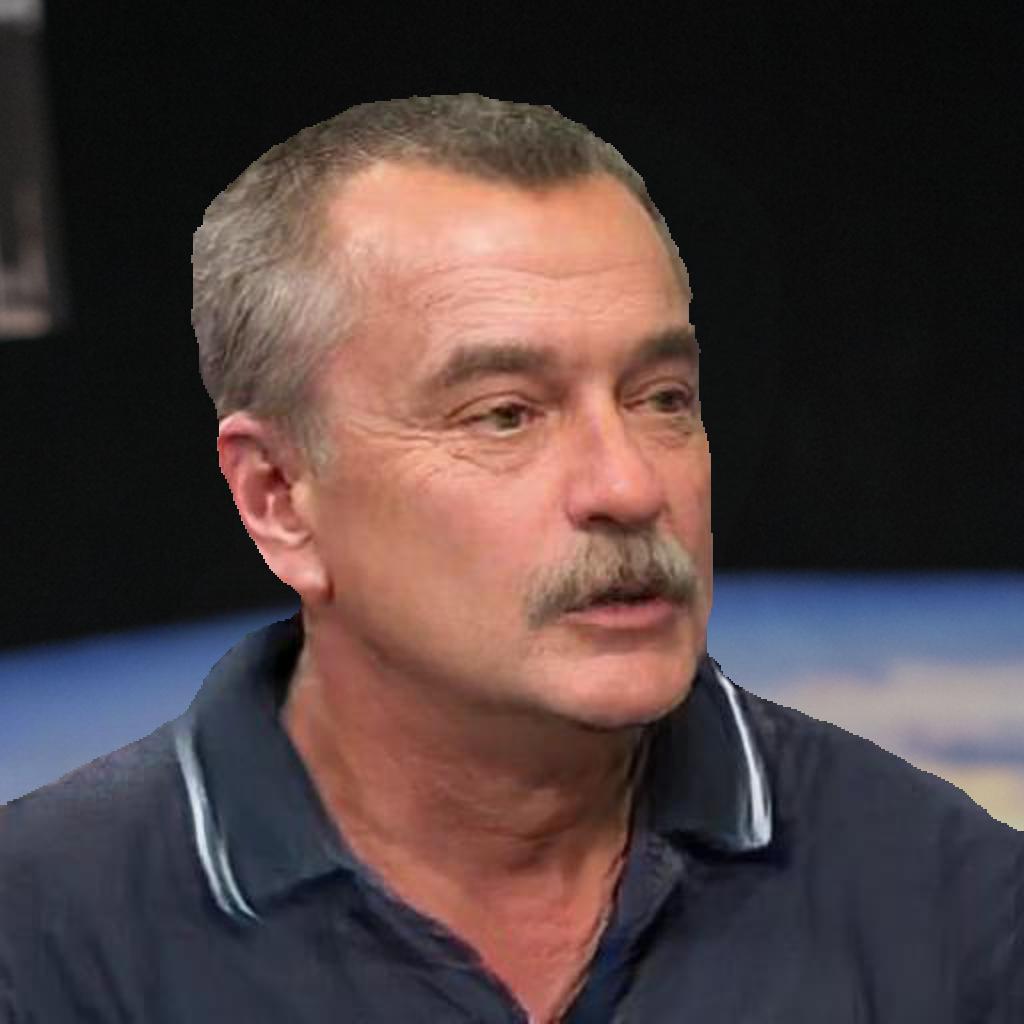} & 
        \hspace{\mrg}
        \includegraphics[width=\wid]{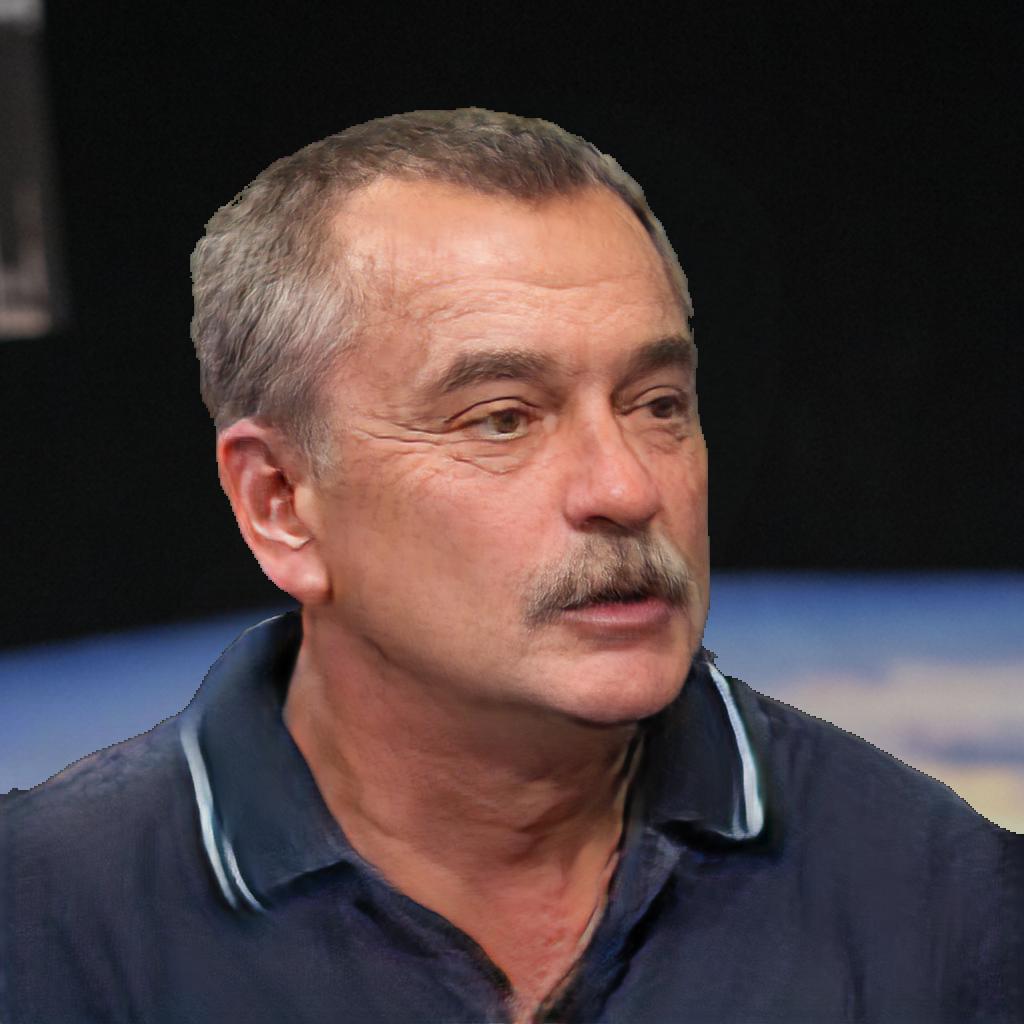} & 
        \hspace{\mrg}
        \includegraphics[width=\wid]{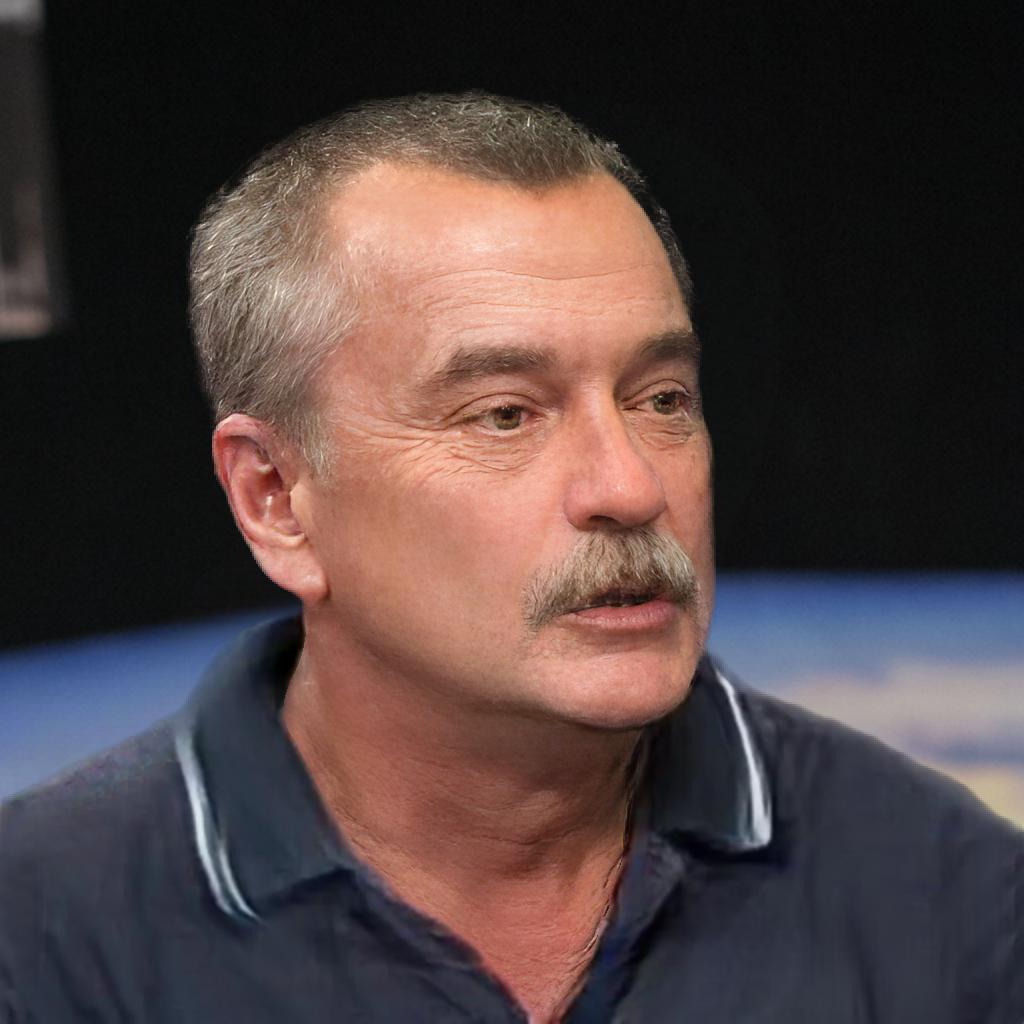} 
        \\ %
        \textbf{Source} & 
        \hspace{\mrg} 
        \textbf{Driver} & 
        \hspace{\mrg} 
        \textbf{Ours (base w/ bicubic)} & 
        \hspace{\mrg} 
        \textbf{HiFaceGAN}~\cite{Yang2020HiFaceGANFR} & 
        \hspace{\mrg} 
        \textbf{Ours (HR)}
    \end{tabular}
    \vspace{-0.4cm}
    \caption{A qualitative comparison of different super-resolution methods applied to the output of our base model. While performing better than a baseline bicubic upsampling, we can see that the state-of-the-art super-resolution method (HiFaceGAN) cannot achieve the same level of high-frequency details fidelity as our approach. Digital zoom-in is recommended.}
    \label{fig:1024px_abl}
\end{figure*}

We demonstrate the comparison of our method  for both cross- and self-reenactment in Figure~\ref{fig:256px_abl} for $256 \times 256$ resolution and in Figure~\ref{fig:512px_abl} for $512 \times 512$. Also, we show qualitative comparison in cross-reenactment scenario for  $1024 \times 1024$ resolution in Figure~\ref{fig:1024px_abl}.

Moreover, we attach a few demonstration videos for one megapixel resolution and video comparison for cross-reenactment $256 \times 256$ and self-reenactment $512 \times 512$ scenarios. We strongly encourage reader to check this video.

One of the interesting points is that the model learns meaningful features in volume tensor, that encodes the geometry of the give source to $\v_s$ with the shape $96 \times 16 \times 64 \times 64$ and attach the video of animation this volumetric tensor in supplementary files.
\begin{figure*}
    \centering    
    \setlength{\wid}{0.24\textwidth}
    \setlength{\mrg}{-0.3cm}
    \begin{tabular}{cccc}
        \includegraphics[width=\wid]{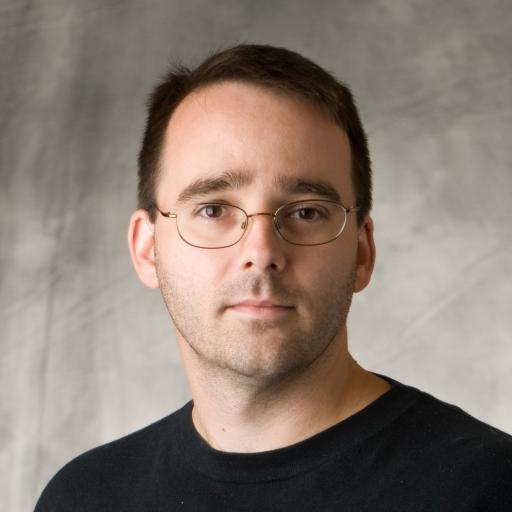} &
        \hspace{\mrg}
        \includegraphics[width=\wid]{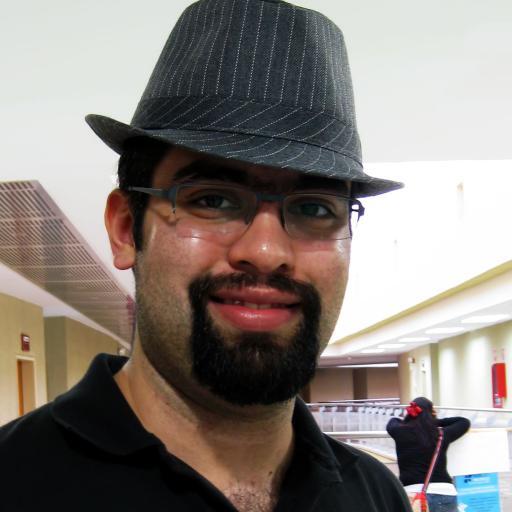} &
        \hspace{\mrg}
        \includegraphics[width=\wid]{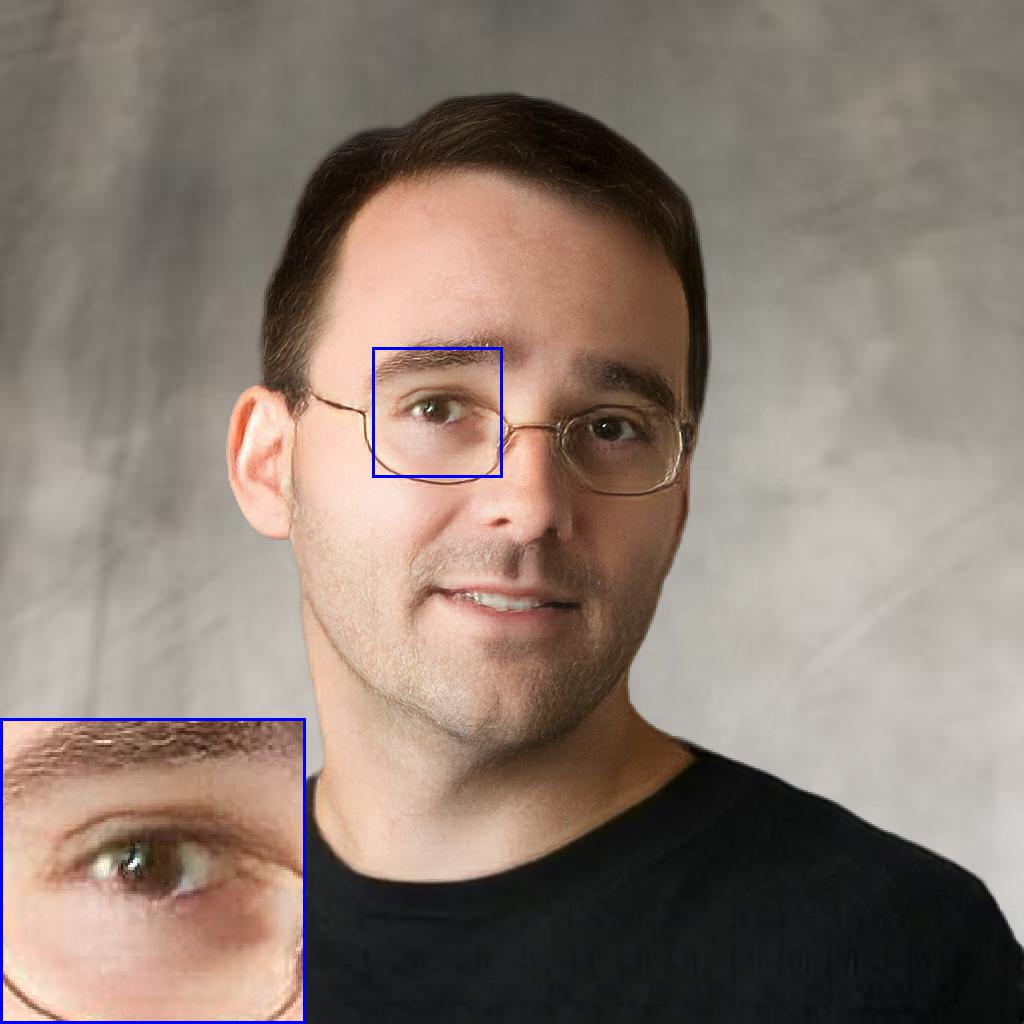} & 
        \hspace{\mrg}
        \includegraphics[width=\wid]{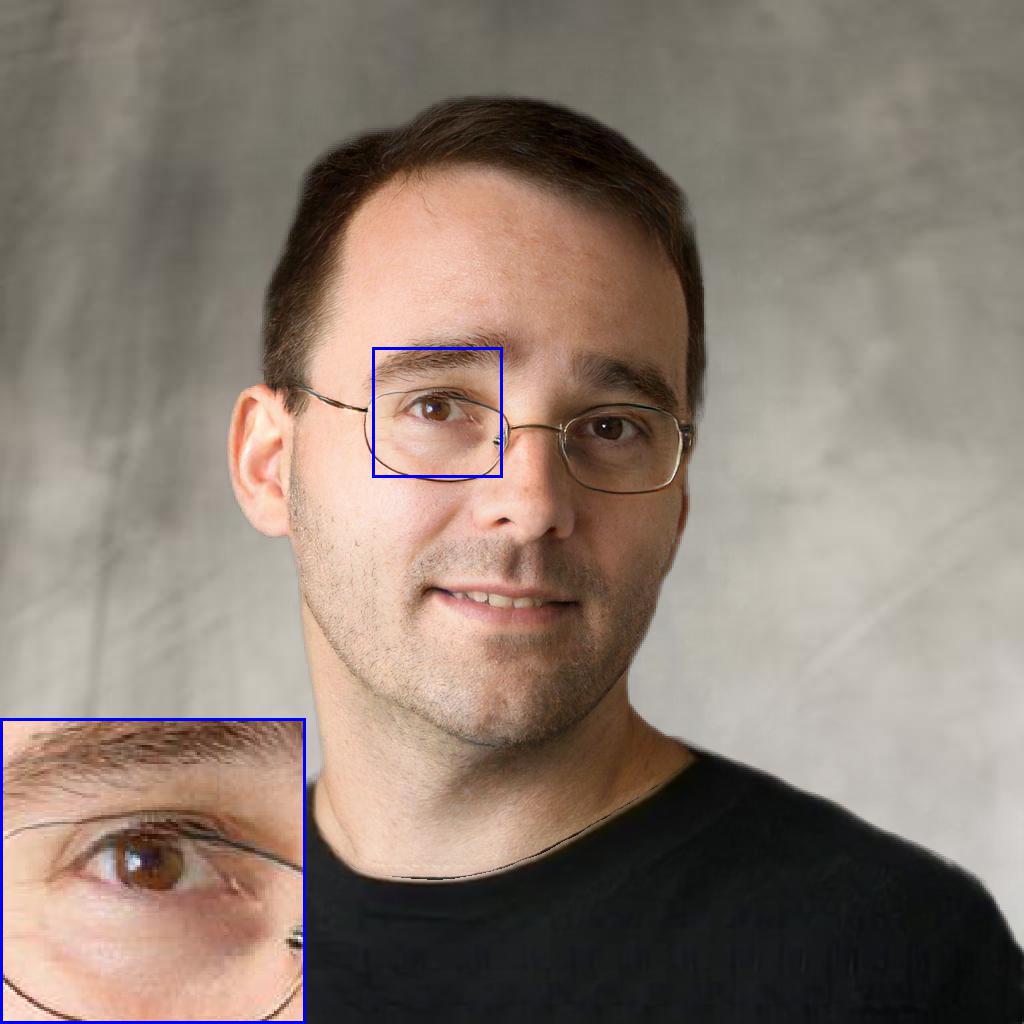}
        \\ %
        \includegraphics[width=\wid]{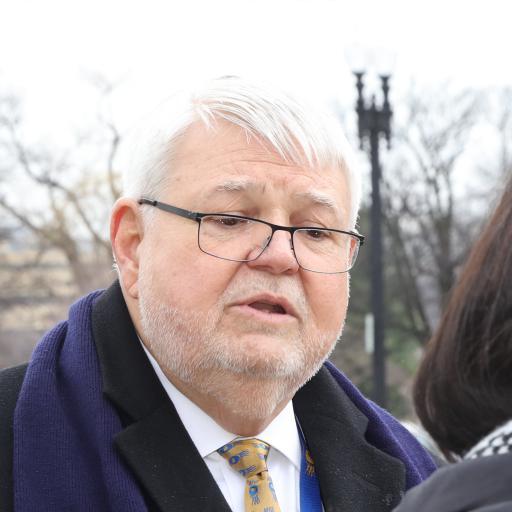} &
        \hspace{\mrg}
        \includegraphics[width=\wid]{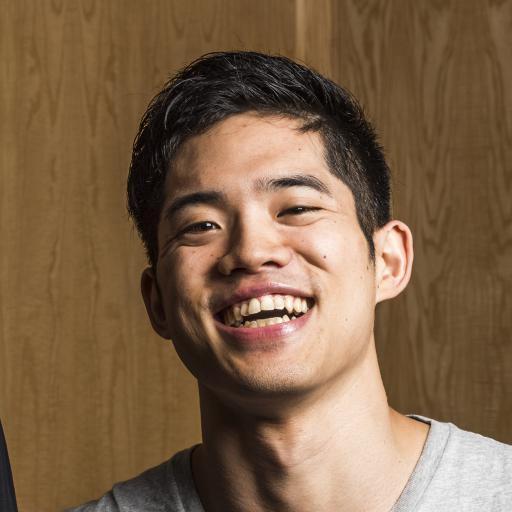} &
        \hspace{\mrg}
        \includegraphics[width=\wid]{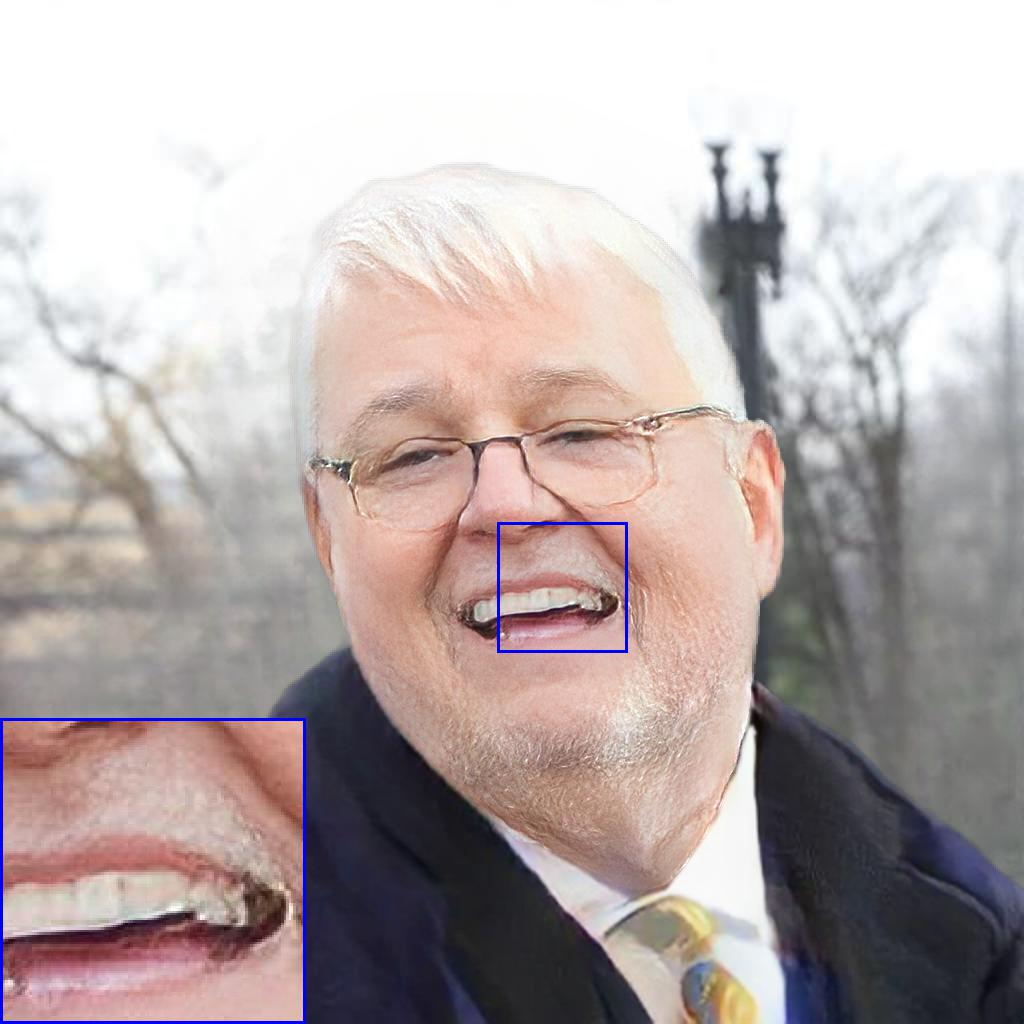} & 
        \hspace{\mrg}
        \includegraphics[width=\wid]{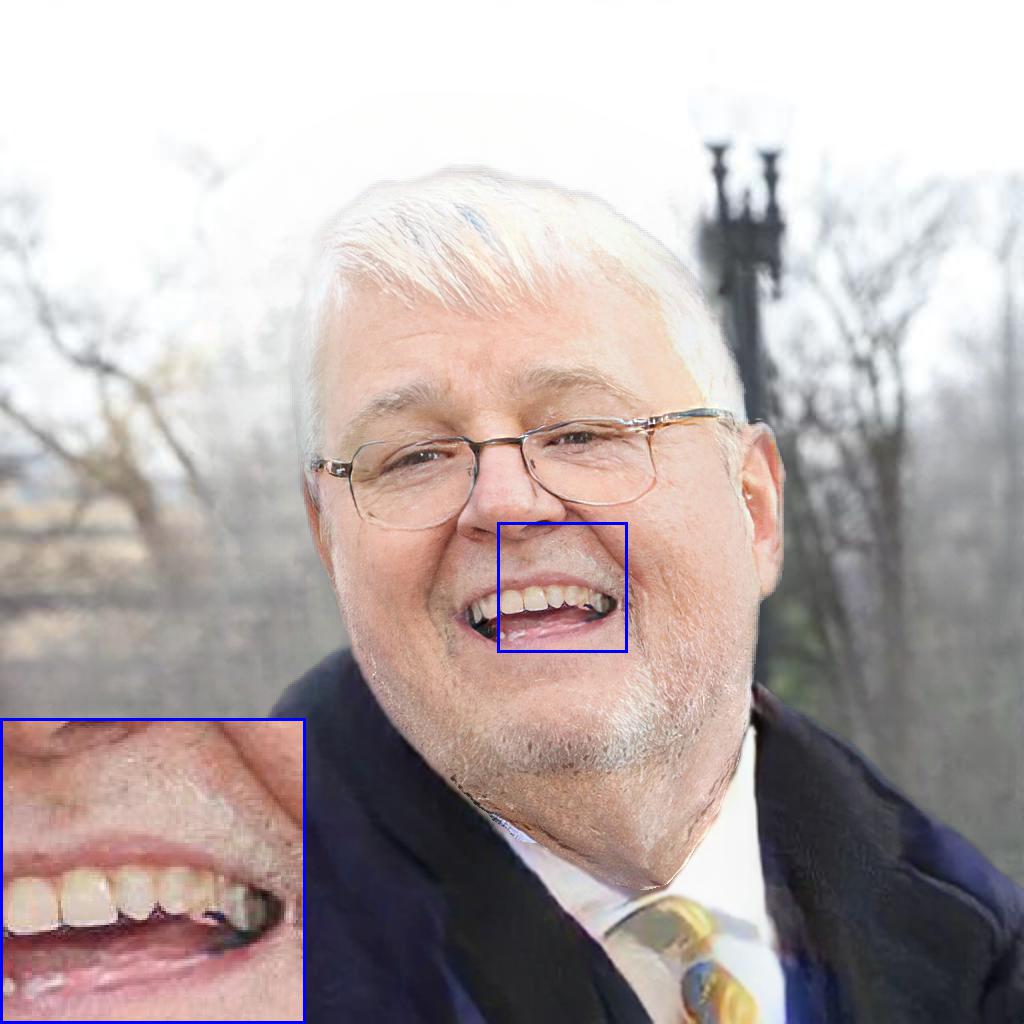}

        \\ 

        \textbf{Source} & 
        \hspace{\mrg} 
        \textbf{Driver} & 
        \hspace{\mrg} 
        \textbf{One stage, fine-tuned} & 
        \hspace{\mrg} 
        \textbf{Final} 
    \end{tabular}
    \vspace{-0.4cm}
    \caption{A qualitative comparison of one stage training with fine-tuned decoder and our final two stage training. Pay special attention to the area around the eyes, glasses, teeth, hair and skin, where the difference between the two approaches is most noticeable. }    \label{fig:ft_vs_2_stage}
\end{figure*}


\end{document}